%% file: acl_latex.tex
\pdfoutput=1

\documentclass[11pt]{article}

\usepackage[preprint]{acl}

\usepackage{times}
\usepackage{latexsym}

\usepackage[T1]{fontenc}

\usepackage[utf8]{inputenc}

\usepackage{microtype}

\usepackage{inconsolata}

\usepackage{graphicx}
\input{commands.tex}

\title{\texttt{JuStRank}: Benchmarking LLM Judges for System Ranking}

\author{Ariel Gera, Odellia Boni, Yotam Perlitz, \\
\textbf{Roy Bar-Haim, Lilach Eden and Asaf Yehudai}\\
 IBM Research}

\begin{document}
\maketitle
\begin{abstract}
Given the rapid progress of generative AI, there is a pressing need to systematically compare and choose between the numerous models and configurations available. The scale and versatility of such evaluations make the use of LLM-based judges a compelling solution for this challenge. Crucially, this approach requires first to validate the quality of the LLM judge itself. Previous work has focused on \emph{instance-based} assessment of LLM judges, where a judge is evaluated over a set of responses, or response pairs, while being agnostic to their source systems. We argue that this setting overlooks critical factors affecting system-level ranking, such as a judge's positive or negative bias towards certain systems. To address this gap, we conduct the first large-scale study of LLM judges as \emph{system rankers}. System scores are generated by aggregating judgment scores over multiple system outputs, and the judge's quality is assessed by comparing the resulting system ranking to a human-based ranking.
Beyond overall judge assessment, our analysis provides a fine-grained characterization of judge behavior, including their \emph{decisiveness} and \emph{bias}.


\end{abstract}

\input{main}

\bibliography{custom}

\appendix
\input{appendix}

\end{document}

%% file: commands.tex
\usepackage[export]{adjustbox}

\usepackage{subfig}

\usepackage{booktabs}
\usepackage{hyperref}
\usepackage{longtable}
\usepackage{tcolorbox}

\usepackage{bbm}
\usepackage{stmaryrd}
\usepackage{amsmath,amssymb}

\usepackage{amsthm}

\usepackage{refcount}

\newcommand\benchmark[0]{\texttt{JuStRank}}
\newcommand\benchmarklong[0]{Judges for System Ranking}

\newcommand{\simsmall}[0]{{\raise.17ex\hbox{$\scriptstyle\sim$}}}

\newcommand\biasstd{$\delta$}

%% file: main.tex
\section{Introduction}

The evaluation of Large Language Models (LLMs) is rapidly adopting the LLM-as-a-judge paradigm \cite{zheng2023llmaaj}, where automatic evaluations with LLMs complement the use of human annotators, or even replace them altogether.
LLM-based judges are increasingly relied upon to conclude which models exhibit superior performance, whether novel training and inference approaches are beneficial, and ultimately which LLM configurations offer a better value proposition to users.

Since relying on an inaccurate judge will likely result in sub-optimal decisions, this trend lends an urgency to evaluating the performance of the LLM judges themselves. 
Indeed, recent works attempt to benchmark judging capabilities, compiling leaderboards of judge performance \cite{lambert2024rewardbench, tan2024judgebench} as well as analyzing their sensitivities and biases \cite{wang2023large,wei2024systematic,bavaresco2024llms,feuer2024style,liu2024rm,xu2024pride,ye2024justice}.

\begin{figure}[t]
    \centering
    \includegraphics[width=\columnwidth]{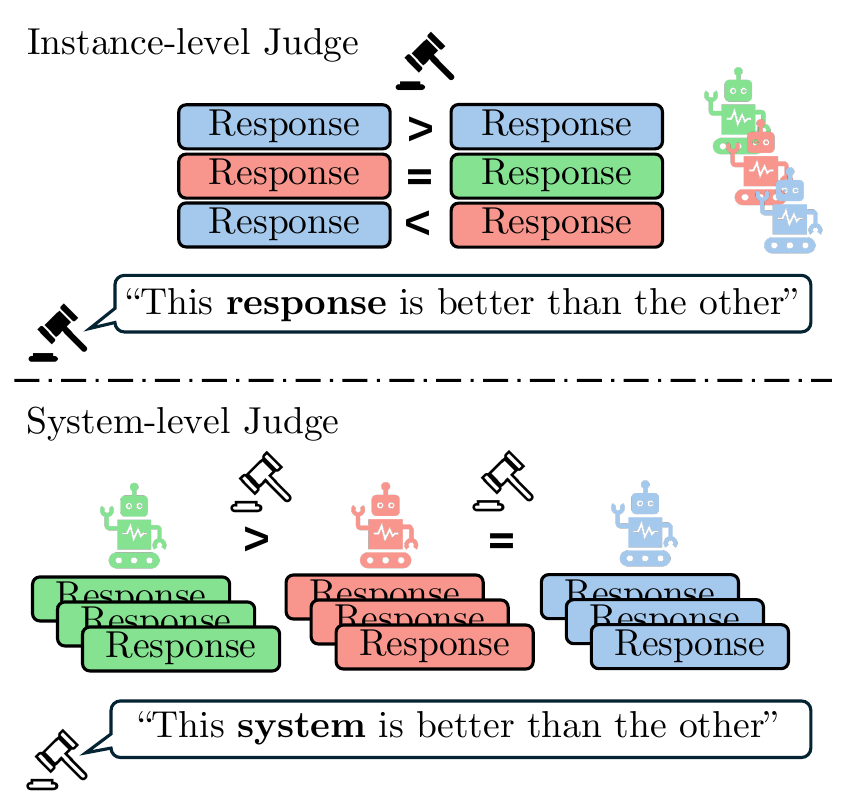} 
    \caption{\textbf{Instance and system level judges make different calls:} An \textit{instance-level} judge (top) is used to make decisions about the quality of individual responses (which may be produced by different systems). A \textit{system-level judge} (bottom) is used to make decisions about the overall quality of systems. For clarity, in this illustration, we focus on pairwise decisions.
    }
    \label{fig:system_level}

\end{figure}

\begin{figure*}[t]
    \centering    \includegraphics[width=0.95\linewidth]{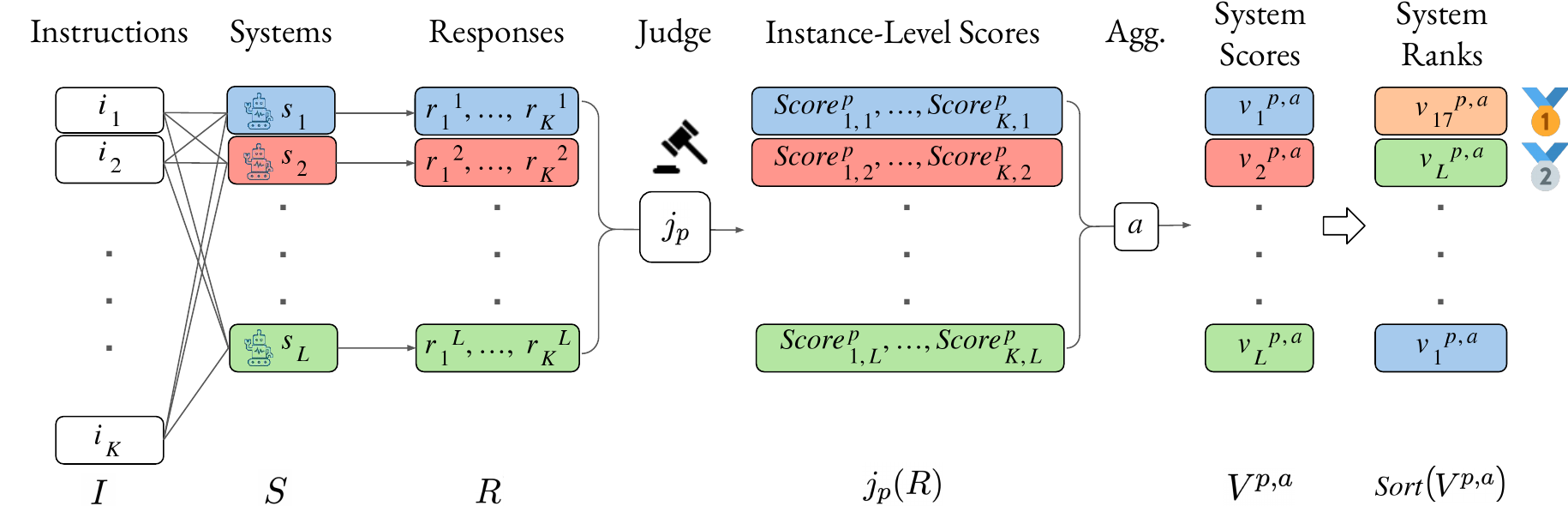} 
    \caption{\textbf{System-level judge pipeline.} Schematic of our 
    data generation pipeline for judge system rankings.}
    \label{fig:schema}
\end{figure*}

These works all focus on the \textit{instance-level performance} of judges. A ``good'' instance-level judge is expected to make a correct judgment about each response, regardless of the system generating it. For example, given a specific pair of responses, the judge may be asked to determine which one is better (Figure~\ref{fig:system_level}, top).
This approach is very much in line with prevailing paradigms for model alignment (e.g., RLHF, DPO; \citealp{lee2024rlaifvsrlhfscaling}) and synthetic data generation \cite{yehudai2024achieving}; these often rely on LLM judges and reward models for making instance-level pairwise decisions on the quality of individual responses.

Although judges are \textbf{evaluated} based on their instance-level performance, very commonly they are actually \textbf{used} for making system-level decisions; namely, to compare and rank different models or different configurations (Figure~\ref{fig:system_level}, bottom). 
Crucially, even very good instance-level capabilities do not guarantee accurate model ranking; and at the same time, mediocre performance on instances could still yield a very accurate overall ranking (\citealp{dorner2024limits}, \S\ref{sec:system}). 
Thus, the \textit{system-level performance} of judges -- that is, to what degree they can correctly decide between candidate systems, and produce accurate model performance rankings -- remains largely an open question. 
Furthermore, system-level evaluations can unveil an entire range of under-explored judge qualities, such as being biased towards certain models or making un-calibrated model preference judgments.

In this work we aim to address this gap, and characterize the system-level evaluation capabilities and behaviors of LLM-based judges. 
To this end, 
we introduce a novel judge benchmark -- \benchmark{} \textit{(\benchmarklong{})}. \benchmark{} compares judges by their ability to correctly rank models, based on agreement with a ground-truth model ranking.
\benchmark{} encompasses a collection of $48$ state-of-the-art judges, including both general-purpose LLMs and reward models. Our large-scale benchmark and analysis allow us to explore the performance and behavior of judges as system rankers.

Our contributions are as follows:


1. We introduce \benchmark{}, the first large-scale benchmark of judges for ranking target systems.


2. We quantify the tendency of a judge to exhibit \textit{system bias}, where some models are judged ``unfairly'' (\S\ref{ssec:bias}). 

3. We reveal an emergent quality of a system-level judge, its \textit{decisiveness} factor; decisive judges consistently amplify the gap between strong and weak target systems (\S\ref{ssec:exaggeration}).

4. To facilitate further research into judge behavior, we release our data\footnote{\href{https://huggingface.co/datasets/ibm-research/justrank_judge_scores}{\benchmark{} Judge Scores data}}, comprising $1.5$M judgment scores given by LLMs and reward models. 

\section{The Gap in Judge Benchmarking}
\label{sec:system}

In this section, we outline why existing estimations of judge performance are insufficient to decide which judge is best at choosing between target systems 
(Figure \ref{fig:system_level}, bottom).
    
At present, users looking for a judge for ranking models, will likely choose it according to the available instance-level judge benchmarks.
Yet, from a theoretical standpoint instance-level judge performance does not directly correspond to system-level judge performance~\cite{dorner2024limits}. 

More specifically, instance-level judge evaluations focus on \textit{how many} errors the judge makes, and do not address the \textit{distribution} of these errors across systems. 
 
For system-level judge evaluation, however, the error distribution plays a key role, as judge errors may distribute unevenly across systems, impacting their induced ranking~\cite{dorner2024limits,von2024measure}.
For example, a judge may exhibit an unjustifiable preference (positive bias) for responses from a particular system $A$. 
Thus, this judge will tend to give $A$ an incorrect ranking, even if it makes very few mistakes on responses from other systems (i.e., has an overall high instance-level accuracy). 
Hence, a more uniform distribution of errors -- reflecting less biased judgment -- is a desirable quality for system-level judges, and one that may lead to a more accurate ranking. 
 
Drawing on this observation, our goal here is to construct a system-level benchmark for judges.
As a benchmark tailored for system-level evaluation, it will enable reliably estimating a judge's ability to rank systems; moreover, our ranking-oriented analysis can shed light on judge behaviors and biases, as they occur in real-world data.

\section{Task Formulation}
In this work we study the use of LLM-based judges for determining the relative quality of systems\footnote{Henceforth, we will use the term \textit{System} to refer to a target model or pipeline that performs a task, and \textit{Judge} for one that is asked to score (or compare) the quality of such systems. Generative LLMs can act as both systems and judges.}, over a given set of user instructions (prompts).

Formally, we begin with a set of $L$ systems $\mathbf{S} = \{s_l\}_{l=1}^{L}$, and $K$ user instructions $\mathbf{I} = \{i_k\}_{k=1}^{K}$. 
Each system produces a response for each such user instruction, denoted as
$R = \{r_k^l\}_{k,l=1,1}^{k, l=K, L}$, 
such that $s_l(i_k) = r_k^l$ 
(see Figure~\ref{fig:schema}). 

Judges $\mathbf{J} = \{j_p\}_{p=1}^{P}$ 
map a pair of instruction $i_k$, and system response $r_k^l$
to a scalar score that estimates the quality of the response. 
Each judge has a specific \textit{realization} for performing this score mapping\footnote{
We note that some realizations, such as the comparative realization in \S\ref{ssec:realizations}, may incorporate a separate set of responses to perform the judgment.}, of the form: $j_p(i_k, r_k^l) = Score_{k, l}^p$.
Once a judge $j_p$ scores all 
$K\times L$ responses,
we can define a scores matrix $j_p(R) \in \mathbb{R}^{K\times L}$ where $j_p(R)_{k,l} = Score_{k,l}^p$.

In order to quantify system-level quality, we must apply an \textit{aggregation method}, $a \in A = \{a\colon \mathbb{R}^{K\times L} \longrightarrow \mathbb{R}^{L} \}$. The aggregation method $a$ maps a scores matrix $j_p(R)$ to a system-level vector $V^{p,a} \in \mathbb{R}^{L}$ where each entry, $V^{p,a}_l$, is a single overall quality score for system $s_l$ by judge $j_p$. In turn, ordering the system scores in $V^{p,a}$ induces a ranking over the systems set $\mathbf{S}$.

We test the performance of judge $j_p$ as a ranker by checking the correlation between the ranking induced by $V^{p,a}$ and a golden ranking for $\mathbf{S}$.

\input{leaderboard}

\section{Experimental setup}
To explore judge performance and behavior, we 
utilize responses from multiple systems (\S\ref{ssec:arena_hard}) and run reward model judges (\S\ref{ssec:reward}) and LLM judges (\S\ref{ssec:realizations}) over these responses. 
To obtain system rankings, we experiment with different aggregation methods (\S\ref{ssec:aggregations}) over the judge scores. 
Finally, the resulting rankings are compared against a gold system ranking, 
taken
from a separate dataset (\S\ref{ssec:chatbot}).

\subsection{System Responses Data}
\label{ssec:arena_hard}
We utilize the \href{https://huggingface.co/datasets/lmarena-ai/arena-hard-auto/tree/main/data/arena-hard-v0.1/model_answer}{Arena Hard v0.1} dataset \cite{li2024crowdsourced} for a diverse set of instructions and system responses. The dataset uses a curated set of $K=500$ challenging instructions, $I$. As of September $2024$, it includes responses from $L=63$ systems, $S$, totaling about $32$K pairs of instructions and their associated system responses, $R$.

\subsection{Generating Judgments}
For every judge realization, $j_p$, we generate a judgment scores matrix, $j_p(R)$, over $R$. In total, we examine $48$ judge realizations, yielding a total of $1.5$M individual judge scores ($63$ systems $\times~500$ instances $\times~48$ judge realizations). 

\subsubsection{Reward Models}
\label{ssec:reward}

We run multiple reward models over $R$. While their exact architectures vary, reward models generally produce a scalar quality score for a given pair of an instruction and a system response.

We utilize the following reward models: ArmoRM-Llama3-8B-v0.1 \cite{wang-etal-2024-interpretable}, Eurus-RM-7b \cite{yuan2024advancing}, InternLM2-7b-reward, InternLM2-20b-reward \cite{cai2024internlm2}, Skywork-Reward-Llama-3.1-8B-v0.2 \cite{liu2024skywork}, Llama-3-OffsetBias-RM-8B \cite{park-etal-2024-offsetbias}, GRM-Llama3.2-3B-ft \cite{yang2024regularizing}, URM-LLaMa-3.1-8B \cite{lou2024uncertainty}.

\subsubsection{LLM Judge Realizations}
\label{ssec:realizations}
Unlike dedicated reward models that produce a single score, generative LLMs can be prompted to judge in multiple ways. Thus, for every LLM we examine several judge realizations.

\paragraph{Absolute judgment - Numeric score \textit{(Numeric)}} The LLM judge is given an instruction and system response, and is asked to provide a quality score for the response between $0$ and $100$.

\paragraph{Absolute judgment - Textual score \mbox{(\textit{Likert})}} The judge 
provides a quality score of the response on a Likert \cite{likert1932technique} scale with $5$ labels: \textit{[Very Bad, Bad, Mediocre, Good, Very Good]}. We then convert the textual judgments to scores in $[1-5]$.

\paragraph{Absolute judgment - Token probablities \mbox{(\textit{TokenProbs})}} The task is framed to the judge as a yes/no question: \textit{Is this a good response?}. We then extract the top log-probabilities for the first generated token, and specifically look at the probabilities for the tokens \textit{yes} or \textit{no}. The judgment score $[0.0-1.0]$ is the sum of probabilities for \textit{yes} divided by the sum of probabilities for \textit{yes} and \textit{no}.

\paragraph{Comparative judgment - Anchor model \mbox{(\textit{Anchor})}} Here the judgment task is comparative, i.e., the judge is asked to state a preference between two responses rather than an absolute quality judgment of a given response. Conducting paired comparisons between a system and all other systems is unfeasible; thus, we follow \citet{li2024crowdsourced} and use the responses of \textit{GPT-4-0314} as \textit{anchors} to which the responses of other systems are compared. Given an anchor response and a system response, we ask the judge which one it prefers. The output is then converted to scores in $[-2,+2]$ (where $0$ indicates a tie, and $+1$ / $+2$ indicate slight/strong preference for the system response over the anchor response, respectively).

In total, we collect judgments from $10$ LLMs and $4$ realizations, yielding $40$ LLM judges. Prompts for all realizations are provided in Appendix~\ref{app:prompts}.

We use the following generative LLM judges: Llama-3.1-405B-Instruct~\cite{dubey2024llama}, Llama-3.1-70B-Instruct, Llama-3.1-8B-Instruct, Llama-3-70B-Instruct, \href{https://mistral.ai/news/mixtral-8x22b/}{Mixtral-8x22B-Instruct-v0.1}, Mixtral-8x7B-Instruct-v0.1~\cite{jiang2024mixtral}, \href{https://mistral.ai/news/mistral-large-2407}{Mistral-Large-Instruct-2407}, \href{https://qwenlm.github.io/blog/qwen2.5/}{Qwen2.5-72B-Instruct}, \href{https://openai.com/index/hello-gpt-4o/}{GPT-4o} and \href{https://openai.com/index/gpt-4o-mini-advancing-cost-efficient-intelligence/}{GPT-4o-mini}.

\subsection{Aggregations}
\label{ssec:aggregations}

Given the raw judgment scores of each judge, $j_p(R)$, there are multiple ways to construct a \textit{ranking} of the $63$ target systems. We calculate rankings using \textbf{Win-rate} aggregation, \textbf{Mean} aggregation, \textbf{Median} aggregation, and \textbf{BT} (Bradley-Terry) aggregation. Details are provided in Appendix~\ref{app:agg}.

\begin{figure}[hb]
    \centering
    \includegraphics[width=\columnwidth]{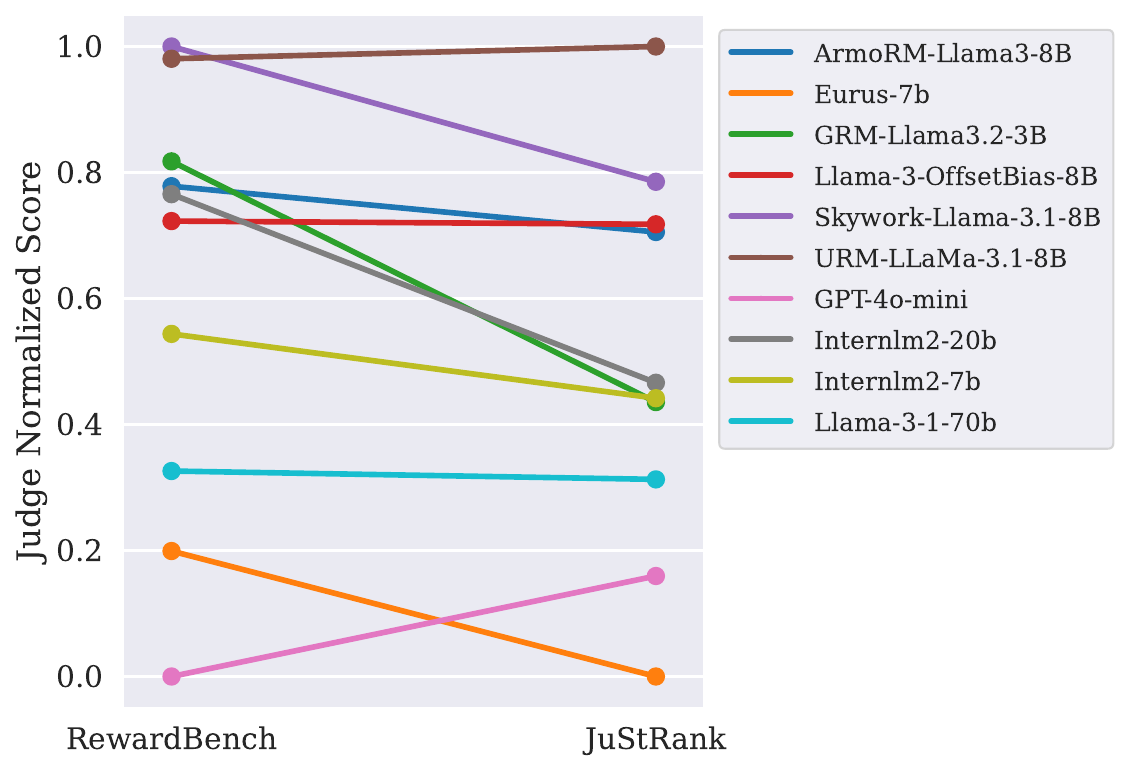} 
    \caption{\textbf{Comparison to RewardBench}. The plot depicts the relative performance of judges present in both \benchmark{} and RewardBench~\cite{lambert2024rewardbench}. For comparison, we perform Min-Max normalization over the judge performance scores (\textit{accuracy} for RewardBench, \textit{Kendall's Tau} for our results). Results shown are for the BT aggregation method; the LLM judges use the \textit{Anchor} realization, which is closest to the setting in RewardBench. Plots for the different RewardBench subsets are shown in Appendix Figure~\ref{fig:rewardbench_subsets}.}
    \label{fig:rewardbench}
\end{figure}

\begin{figure*}[t]
    \centering
    \includegraphics[width=0.7\linewidth]{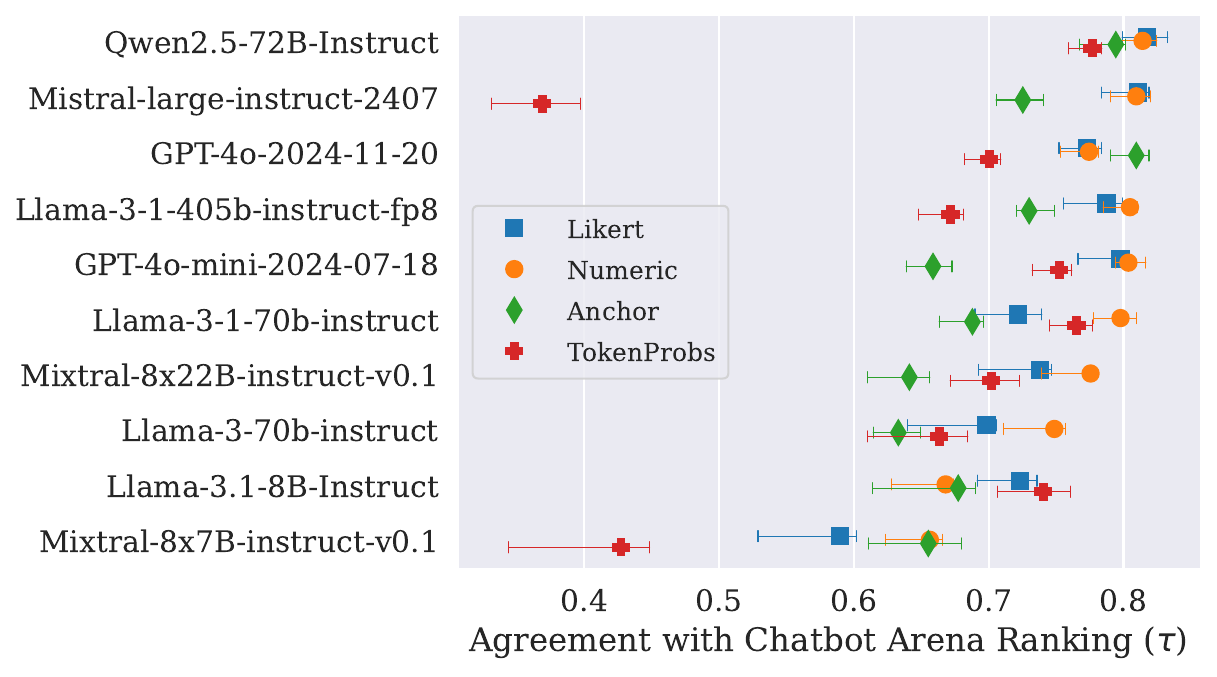} 
    \caption{\textbf{LLM judge realizations}. Kendall's Tau correlations ($\pm95\%$ bootstrapping CI) between the system rankings produced by various LLM judge realizations (\S\ref{ssec:realizations}) and the gold system ranking from Chatbot Arena.
    The plot depicts results for the BT aggregation method; for the full results, refer to App. Table~\ref{tab:leaderboard_full}.}
    \label{fig:llms}
\end{figure*}

\subsection{Gold Ranking - Chatbot Arena Battles} \label{ssec:chatbot}
Human preference data from Chatbot Arena~\cite{zheng2023llmaaj} serve as our ground-truth reference for the relative quality of systems. 
Chatbot Arena relies on human-annotated ``battles'' between system responses to produce a system ranking.
We use the \textit{English Hard Prompts} subset\footnote{\href{https://lmsys.org/blog/2024-05-17-category-hard/}{Chatbot Arena Hard Prompts}} of their data.
We chose this subset as its distribution of user instructions has been shown~\cite{li2024crowdsourced} to match that of our system response data (\S\ref{ssec:arena_hard}).
We extract the data and ranking following the official code (see Appendix~\ref{app:chatbot}).

Given a system ranking produced by a judge, we quantify judge performance via the correlation between its ranking and the reference ranking from Chatbot Arena.
Simply put, we assume that a ranking given by a good automated judge would have a high agreement with the ranking compiled from human judgments.

\section{\benchmark{} - Judge Performance Results}
\label{sec:benchmark}
Table~\ref{tab:leaderboard} depicts the $10$ top-performing judges on \benchmark{}, based on their ranking agreement ($\tau$) with the ground-truth human ranking from Chatbot Arena. For each judge model, the best-performing realization and aggregation method is shown. 

As seen in the table, there are both LLMs and reward models that reach decent agreement with the gold ranking. Moreover, several $8$B-parameter reward models are on par with much larger LLMs on the task of system ranking. Thus, we see that reward models, which are explicitly trained to make instance-level decisions between pairs of responses, can excel at the system-level ranking task as well.

Note that an identical correlation score with the ground-truth ranking does not indicate that the judges produce the \textit{same} ranking; rather, each judge has a different pattern of agreement with the ground-truth. Correlations among the judges themselves are shown in App. Fig.~\ref{fig:judge_corrs}.

\paragraph{Comparison to Instance-Level Performance} In Figure~\ref{fig:rewardbench} we compare our system-level judge leaderboard to the instance-level benchmark \href{https://huggingface.co/spaces/allenai/reward-bench}{RewardBench}~\cite{lambert2024rewardbench}. 
The results demonstrate that better instance-level judges are not always better system rankers, highlighting the discrepancy between the two tasks. Thus, \benchmark{} offers a novel perspective on judge ability.
However, there may be additional factors at play as well.
For LLM judges, we use a slightly different realization from the comparative prompts used for RewardBench.
Moreover, since creators of reward models aim to do well on RewardBench, it is possible that some newer reward models are slightly overfitted to this test distribution.

\input{win_rate_examples}

\subsection{Effects of LLM Realizations}

Figure~\ref{fig:llms} depicts the performance of the LLM judge models by their realization~(\S\ref{ssec:realizations}). 
The plot demonstrates that the choice of realization has a considerable effect on the system ranking quality; this appears to be nearly as important as the identity of the LLM used. We confirm this finding using statistical variance analysis (Appendix \ref{app:stats}).

Many works recommend asking LLMs for comparative rather than absolute judgments \cite{zheng2023llmaaj}. However, in our experiments the comparative realization (\textit{Anchor}) exhibits lower performance, with the notable exception of GPT-4o.
The best realizations overall were \textit{Numeric} and \textit{Likert}, where the judge is asked to provide a verbalized quality score. This is in line with findings from \citet{tian-etal-2023-just}, who report better calibration with verbalized LLM confidence scores. The higher performance for both \textit{Numeric} and \textit{Likert} realizations -- compared to \textit{Anchor} and \textit{TokenProbs} -- is statistically significant (App.~\ref{app:stats}).

We also note that each realization induces a characteristic distribution of judge scores, $D^p$, such that $Score_{k, l}^{p} \sim{D^p}$.
Notably, the LLM judges tend to produce particular score values more often than others. Refer to Appendix~\ref{app:distributions} for more details.

\section{Judge Behavior}
Next, we explore more fine-grained judge behaviors, beyond the bottom-line system rankings. 

To that end, we focus on the judgment task of pairwise system preference, as this is the foundation of system ranking tasks. As in \S\ref{sec:benchmark}, our aim is to gain an understanding of judge performance and characteristics, by comparing judge behavior on pairwise system preference to ground-truth data.

\paragraph{Pairwise Win-Rates}
For every judge $j_p$, and for every pair of systems ($s_a$, $s_b$), the win-rate of $s_a$, denoted by $WR^p(s_a, s_b)$, is the number of instances where it received a higher score than $s_b$, divided by the number of non-tied instances (cf. Appendix~\ref{app:winrate}).
Thus, we calculate the pairwise win-rate for each system pair according to each judge. Note that the win-rates are calculated on the scores matrix $j_p(R)$, i.e., before applying an aggregation method.

\paragraph{Gold Win-Rates}
Similarly, we extract gold pairwise win-rates, $WR^g$, from Chatbot Arena (App.~\ref{app:chatbot}). $59$ systems appear both in our response data (\S\ref{ssec:arena_hard})
and in Chatbot Arena; in total, we have both judge and gold data for $968$ head-to-head comparisons between pairs of systems.

\input{beta_fig}

\subsection{Some Judges are Particularly Decisive}
\label{ssec:exaggeration}

Figure~\ref{fig:pred_wr_examples} depicts the relationship between predicted win-rates and gold win-rates for several judges.
The quadrants in the figure indicate whether the judge's pairwise preference decision is aligned with the gold preference. 
As can be expected, the judge predictions in Figure~\ref{fig:pred_wr_examples} are often centered around the ground-truth win-rates determined by humans. But strikingly, some judges exhibit unique prediction patterns, yielding win-rates that are consistently closer to the extremes ($0.0$ / $1.0$) compared to the human data. For instance, for pairs with a ground-truth 
win-rate of $\simsmall0.8$, 
the predicted win-rate in the judgments of Llama-405B (Fig.~\ref{fig:pred_wr_examples}, right) tends to exceed $0.9$. Put simply, when faced with a response from a strong system, the judge is very likely to prefer it over the response of a less capable system, even where human judges are less decisive.

This sigmoidal win-rate prediction pattern resembles behaviors previously described for classifier calibration~\cite{silva2023classifier}, where classifiers may exhibit ``overconfidence'' in their predicted probabilities.\footnote{Note, however, that the behavior in our case does not reflect judge probability scores, but rather the empirical ratio of instances where the responses $\{r_k^l\}_{l=1}^{l=L}$ of a system $k$ are preferred over those of another system.}
Thus, following \citet{kull2017beta}, we quantify judges' decisive (overconfident) behavior by fitting the cumulative beta distribution function to the win-rate prediction plots. This enables describing judge prediction behavior in terms of a single fit value $\alpha=\beta$, where $\alpha \in [0, \infty]$, a value of $\alpha=1$ represents no over- or under-decisiveness, and larger values represent more decisive behavior (refer to Appendix~\ref{app:beta} for details). Figure~\ref{fig:beta_example} and App. Fig.~\ref{fig:all_betas} depict the beta curve fit for win-rates of various judges.

Figure~\ref{fig:beta_by_judge} compares judge realizations in terms of their decisiveness behavior. We see that LLM judges are usually more decisive when directly asked to provide a quality score, and in particular a textual one (\textit{Likert}); in contrast, the realization that relies on token probabilities (\textit{TokenProbs}) does not give rise to such a pattern, and can even result in judge ``indecision'' (i.e., $\alpha < 1$). 

This pattern can be explained from two directions. First, the human judgments (\S\ref{ssec:chatbot}) were collected from multiple individuals, who likely have differing preferences; this may introduce some noise that could lead to \textit{less extreme win-rates in the gold data}. The other factor is the judges, who may rely on certain heuristics to identify responses from strong systems \cite{feuer2024style}, leading to \textit{more extreme win-rates in the judge data}. While the variance between judges (Fig.~\ref{fig:beta_by_judge}) supports the latter, we cannot determine this conclusively.

In practical terms, extreme win-rates can be beneficial to users, as they increase the likelihood of a correct system preference decision given a smaller set of responses (see~\citealp{ashury-tahan-etal-2024-label}).

\subsection{Bias Towards Specific Systems}
\label{ssec:bias}

A major concern when using judges for system preference is \textit{judge bias} -- a judge may treat a specific system ``unfairly'', by consistently judging its responses too favorably or too harshly (see \citealp{von-daniken-etal-2024-favi}).

We define the bias $B_{s_a}^p$ of judge $j_p$ towards system $s_a$ by the expectation over the differences between the predicted and gold win-rates,
over all systems that $s_a$ interacts with. Formally, $B_{s_a}^p=\mathbb{E}_{s_{b}\in S} (WR^p(s_a, s_b) - WR^g(s_a, s_b))$.\footnote{Our formulation of bias aims to reflects the practical impact of the judge bias on system preference. This is in contrast to the Favi-Score metric proposed by~\citet{von-daniken-etal-2024-favi}, which is decoupled from the overall accuracy of preference decisions.} In other words, if according to $j_p$ the win-rates of system $s_a$ are (on average) higher than those in the human data, we will say that $j_p$ exhibits \textit{positive bias} towards it; and if they are lower than the ground-truth, $j_p$ would be said to exhibit \textit{negative bias}.

Note that the decisiveness behavior in \S\ref{ssec:exaggeration} directly entails a general bias pattern in some judges -- namely, a positive bias towards strong systems, and a negative bias towards weak ones. Thus, we calculate a \mbox{decisiveness-corrected} bias, ${B'_{s_a}}^p$, where the gold win-rate $WR^{g}$ is replaced by $WR^{g'_{p}}$, i.e., the predicted value for the gold win-rate on the beta distribution fit for judge $j_p$ (App.~\ref{app:beta}).

We observe some consistent trends of system-specific bias that are common across judges. Figure~\ref{fig:bias_top} depicts systems for which there is high bias across judges. For instance, most judges exhibit a strong positive bias towards \href{https://nexusflow.ai/blogs/athene}{Athene-70B}, to the extent that it is often ranked by them as the \#1 system. In contrast, GPT-4-0613, which is $27$th in the gold ranking, receives negative bias, resulting in a median rank of $38$ among the judges.

We also ask whether LLM judges exhibit \textit{self-bias}~\cite{xu2024pride,panickssery2024llm},
i.e., bias towards the system that uses the same underlying LLM. While we find some instances of self-bias, this is not a consistent effect across judge realizations (App. Table~\ref{tab:self_bias}).

To quantify the overall propensity of a judge for bias, we measure the standard deviation of its bias over all systems, \biasstd{} $ = \sigma_{s\in{S}}({B'}^p)$. The bias measure for each judge is presented in App.~Table~\ref{tab:leaderboard_detailed}.

\subsection{Characterizing Judge Behaviors}
We have shown that beyond their overall ranking capability (\S\ref{sec:benchmark}), judges exhibit distinct traits in their system-level judgments -- in particular, they show different levels of \textit{decisiveness} (\S\ref{ssec:exaggeration}), and overall propensities for \textit{bias} (\S\ref{ssec:bias}). Interestingly, each of these traits (cf. App. Table~\ref{tab:leaderboard_detailed}) is correlated to the ranking quality $\tau$, with $r=0.55$ for the $\alpha$ decisiveness measure, and $r=-0.56$ for the bias propensity \biasstd. At the same time, these marked traits are -- by design -- uncorrelated with each other ($r=-0.07$ between $\alpha$ and \biasstd).
Thus, our analyses reveal global system-level judge traits, 
ones that remain hidden when assessing judges from an instance-level perspective. 

\input{bias_fig}

\section{Related Work} \label{Sec:related}
Applying and assessing automatic metrics for system-level evaluation has been studied for decades, in particular for natural language generation tasks \cite{reiter-belz-2009-investigation,louis2013automatically, deutsch-etal-2022-examining}.
In the context of LLM-based judges, however, system-level evaluation is still under-explored.

Prior works on LLM-based judges
have opted for an instance-level evaluation approach, curating benchmarks of 
responses
with ground-truth quality annotations in order to evaluate judge performance. Most prominently, RewardBench \cite{lambert2024rewardbench} compares dozens of judges 
(including reward models, generative 
LLMs, and classifiers) on the task of 
deciding between pairs of outputs.
RewardBench aims to identify the most suitable judges for model alignment, e.g., for use in RLHF; in contrast, our work measures judges in terms of their ability to compare the performance of candidate systems. Another recent instance-level benchmark, JudgeBench \cite{tan2024judgebench}, focuses on curating challenging response pairs where the judge must discern subtle errors.

Multiple works are dedicated to analyzing various biases~\cite{ye2024justice} and undesirable behaviors exhibited by judges. These include positional bias~\cite{wang2023large}, verbosity bias~\cite{saito2023verbosity,chen2024odin} and self-bias~\cite{xu2024pride}, as well as sensitivity to prompts~\cite{wei2024systematic}, source datasets~\cite{bavaresco2024llms}, epistemic markers~\cite{lee2024llm} and style~\cite{feuer2024style, liu2024rm}.

Several popular benchmarks rely on LLM judges to produce leaderboards of state-of-the-art systems. 
Such benchmarks -- e.g., Arena Hard \cite{li2024crowdsourced} and AlpacaEval \cite{dubois2024length} -- do perform a system-level validation of their resulting leaderboards against other benchmark rankings (see \citealp{perlitz2024llmbenchmarksagreefixing}). However, such efforts are limited to validating the particular dataset and judge setup chosen for the benchmark (usually with
GPT-4 as the judge), rather than comparing and analyzing the performance of different judge models and implementations. \citealp{thakur2024judging} conduct a task-specific system-level evaluation of judges, over the TriviaQA \cite{joshi-etal-2017-triviaqa} dataset. Compared to their work, the present study is on a larger scale and offers novel metrics and analyses on system-level judge behaviors.

\section{Discussion}

The usage of LLM-based judges is continually expanding.
Moreover, many research papers -- proposing novel architectures, algorithms and training methods -- rely heavily on system-level evaluations using judges as evidence for the utility of their approach.
But without evaluating the judges on such system-level tasks, how can one know whether to trust such evaluations, and their conclusions?

We are the first to investigate on a large scale the performance of LLM-based judges on the system ranking task. Our resulting benchmark, \benchmark{}, will assist users and researchers in choosing the judge best suited for their needs.

Choosing a judge requires many fine-grained decisions.
A user can decide which reward model or LLM to use as the judge; opt for relative judgments or absolute scores; try various prompts; apply different aggregations to compile a ranking, etc. Furthermore, these decisions may interact in non-trivial ways (e.g., the distribution of scores a judge tends to output can dictate which aggregations will work well).  
Indeed, our findings confirm that such decisions substantially affect system-level judgments (\S\ref{sec:benchmark}), and thus are quite likely to change the model selection of an end user, or flip the conclusions of an NLP research paper.

Our system-level approach has multiple additional benefits.
First, it forces the evaluation of judges to be representative with respect to \textit{the distribution of systems that generate the responses}. In existing instance-level benchmarks this factor is not taken into account, and likely results in less accurate judge evaluations. 
Second, it affords a new perspective on what it means for a judge to be biased; 
on the one hand, we discover some decisiveness trends (\S\ref{ssec:exaggeration}) that may actually be useful for making correct preference decisions, and increasing the separability between systems; and on the other, we report some problematic biases that directly distort the judgment of particular systems (\S\ref{ssec:bias}). An important avenue for future work is to connect our findings here to the existing literature on judge biases \cite{ye2024justice}, and understand to what extent both of these behaviors stem from particular LLM style attributes \cite{feuer2024style}.



Given this vast and complex space, our work is admittedly only a first step in understanding the behavior of judges for ranking and selecting LLMs. We release our raw judgment scores data, and
encourage the community to explore these issues further: for instance, by 
training dedicated system-level judges,
exploring judge ensembles, or studying other aggregation approaches.
We believe that \benchmark{} can facilitate such research directions, as it can be easily extended to new judges without requiring additional human annotations.

Our hope is that both practitioners and researchers can benefit from \benchmark{}, by making more informed choices of judges for their needs.



\section{Conclusion}
In this work we conducted the first comprehensive evaluation of system ranking by LLM judges. We tested a wide array of judges, including reward models and different realizations of generative LLMs, over a large collection of systems.

We collected system responses over a diverse set of instructions. The judges scored each response, and we compiled a ranking by aggregating the judgments over all responses.
Then, the quality of the judge's system ranking was compared to a human ranking, producing the \benchmark{} leaderboard.

\benchmark{} allows users to pick judges that are better aligned with the goal of choosing between different models and configurations. \benchmark{} demonstrates that judge ranking abilities are 
not directly tied to LLM size or overall quality, and that some dedicated reward models are on par with leading LLM judges. 
Moreover, our analysis reveals emergent judge traits -- \emph{decisiveness} and \emph{bias} -- that are strongly correlated with their ranking ability.

\section*{Limitations}
The gold reference data -- the \textit{English Hard Prompts} subset of Chatbot Arena -- does not include user instructions or responses. Hence, we collect judgment data over Arena Hard, which contains a large set of instructions and responses. This raises some questions regarding our ability to directly compare the LLM judges and human judges. However, given that Arena Hard was designed to match the distribution of user instructions in \textit{English Hard Prompts} (see \citealp{li2024crowdsourced}), we assume that these datasets are sufficiently similar.

Our analyses of LLM judge realizations are, by necessity, limited to the specific realization prompts that we used. Several studies show that LLMs \cite{mizrahi2024state} as well as LLM judges \cite{wei2024systematic} are brittle with respect to prompt phrasing, and hence this may have had an impact on the results. In addition, there can be some variations in judge responses depending on the exact API and inference implementation used.

As in multiple other works, here we treat human preference as a single concept. In practice, however, preference is inherently subjective, and is composed of numerous dimensions (e.g., helpfulness, safety, style, coherence etc.). For instance, one individual may prefer succinct model responses while another would prefer more detailed answers. Thus there is no single ``human preference'', but rather a collection of preference decisions that depend on the annotation guidelines, cultural context, and human idiosyncrasies \cite{conitzer2024social,kirk2024prism}.

Note that following \citet{peyrard-etal-2021-better}, as well as Chatbot Arena \cite{chiang2024chatbot}, we generally regard the ground-truth quality of a system in terms of the Bradley-Terry model; simply put, a better system is a system that ``wins'' more often. Thus, in this work we do not directly consider the quality difference in system responses per instance, i.e., beyond counting wins/losses. Still, some of the aggregation methods we use (e.g., mean) implicitly reflect other perspectives on system quality.

All of our analyses are performed on heterogeneous datasets of user instructions to LLMs. Thus, while we study judges through the lens of general-purpose LLM usage, we cannot draw conclusions on judge behavior that is task-specific (or in specialized domains), nor on performance in languages other than English \cite{Gureja2024MRewardBenchER}. The issue of task, domain, and language-specific judge behavior is thus an important avenue for future work.

%% file: leaderboard.tex
\begin{table*}[ht] 

\begin{center}
\begin{tabular}{lllc}
\toprule
Judge Model & Realization & Aggregation & Agreement ($\tau$) \\
& & & with Gold Ranking \\
\midrule
\midrule
Qwen2.5-72B-Instruct & Likert & Win-Rate & .83 \\
URM-LLaMa-3.1-8B & Reward & Mean & .82 \\
GPT-4o-2024-11-20 & Anchor & Mean & .82 \\
Llama-3-1-405b-instruct-fp8 & Numeric & Mean & .81 \\
Mistral-large-instruct-2407 & Likert & BT & .81 \\
GPT-4o-mini-2024-07-18 & Numeric & Win-Rate & .81 \\
ArmoRM-Llama3-8B-v0.1 & Reward & Mean & .80 \\
Llama-3-1-70b-instruct & Numeric & Win-Rate & .80 \\
Skywork-Llama-3.1-8B-v0.2 & Reward & Mean & .79 \\
Llama-3.1-8B-Instruct & TokenProbs & Mean & .78 \\
\bottomrule
\end{tabular}
\end{center}
\caption{\textbf{Top 10 judges by ranking performance}. Judges are sorted by the Kendall's Tau correlation between their overall system ranking and the gold ranking from Chatbot Arena (\S\ref{ssec:chatbot}). For every judge model, only the best-performing realization and aggregation method is shown. For the full results, refer to Appendix Table~\ref{tab:leaderboard_full}.}  \label{tab:leaderboard}
\end{table*}

%% file: win_rate_examples.tex
\begin{figure*}[t]
\setlength{\belowcaptionskip}{-6pt}

\centering
\subfloat{\includegraphics[width=0.3\textwidth]{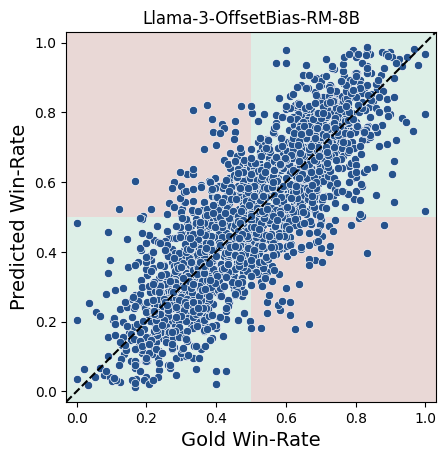}}
\subfloat{\includegraphics[width=0.3\textwidth]{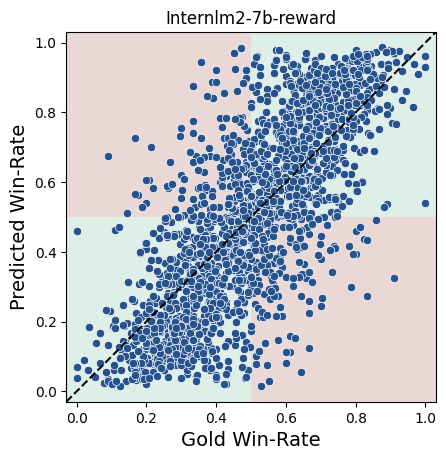}}
\subfloat{\includegraphics[width=0.3\textwidth]{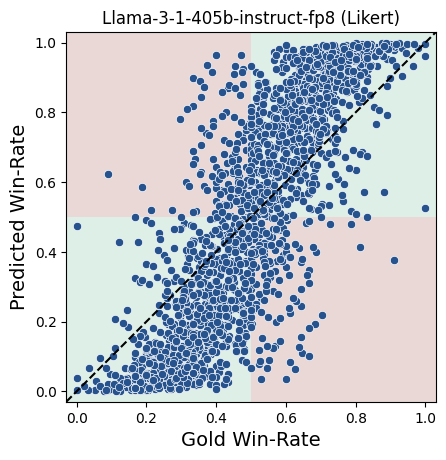}}

\caption{\textbf{Predicted pairwise win-rates}. Each point represents a win-rate between a pair of systems $WR(s_a, s_b)$ (App.~\ref{app:winrate}). The x-axis denotes the gold win-rate from Chatbot Arena, and the y-axis denotes the predicted win-rate as derived from the judge scores. The diagonal marks an exact match between the predicted and gold win-rate; the quadrants signify whether the predicted winning system is the same (green) or different (red) from the gold winning system for this pair. Note that every pair is represented twice (e.g., $WR(s_a, s_b)=0.2$, $WR(s_b, s_a)=0.8$).}
\label{fig:pred_wr_examples}

\end{figure*}

%% file: beta_fig.tex
\begin{figure*}
\centering
\setlength{\belowcaptionskip}{-2pt}


\subfloat{\label{fig:beta_example}}
\subfloat{\label{fig:beta_by_judge}}

\includegraphics[width=0.95\linewidth]{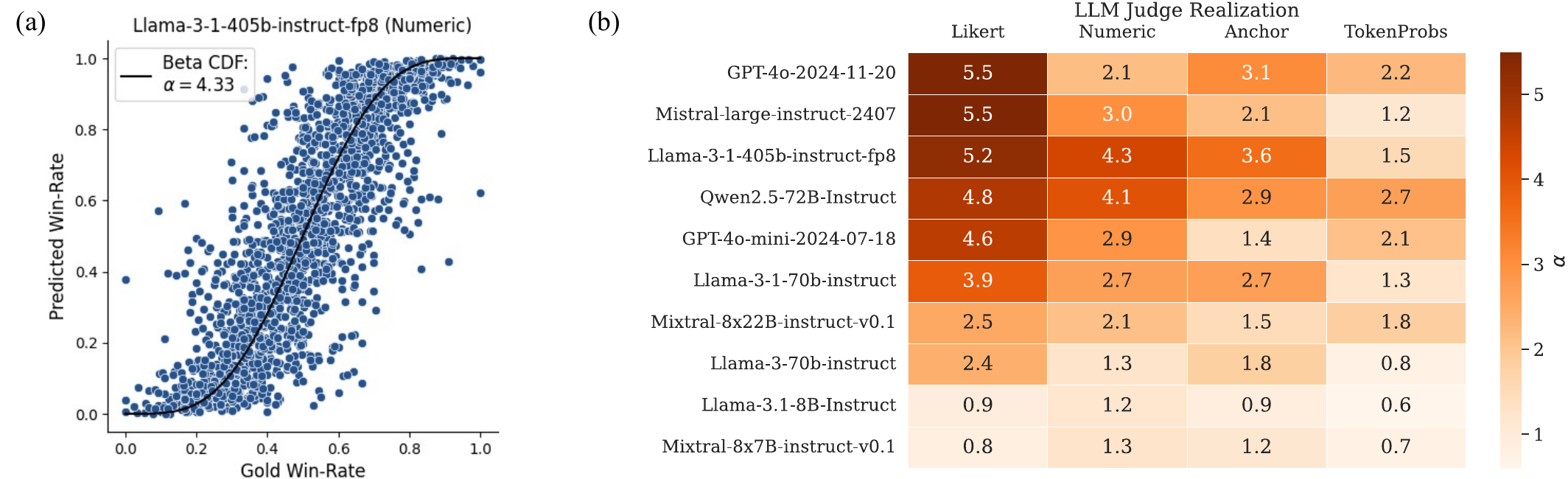}

\caption{\textbf{Beta distribution fit of pairwise win-rates}. (a): \textit{Judge beta fit example}. Each point represents the win-rate between a pair of systems, $WR(s_a, s_b)$; the curve and $\alpha$ value describe a fit to the beta distribution (App.~\ref{app:beta}). Plots for all judges are in App. Fig.~\ref{fig:all_betas}. (b): \textit{Decisiveness by judge realization}. Cell values denote the decisiveness behaviors of different LLM judge realizations, as described by the $\alpha$ value for their win-rate distribution.}
\end{figure*}

%% file: bias_fig.tex
\begin{figure}[t]
\setlength{\belowcaptionskip}{-10pt}

\includegraphics[width=1\columnwidth]{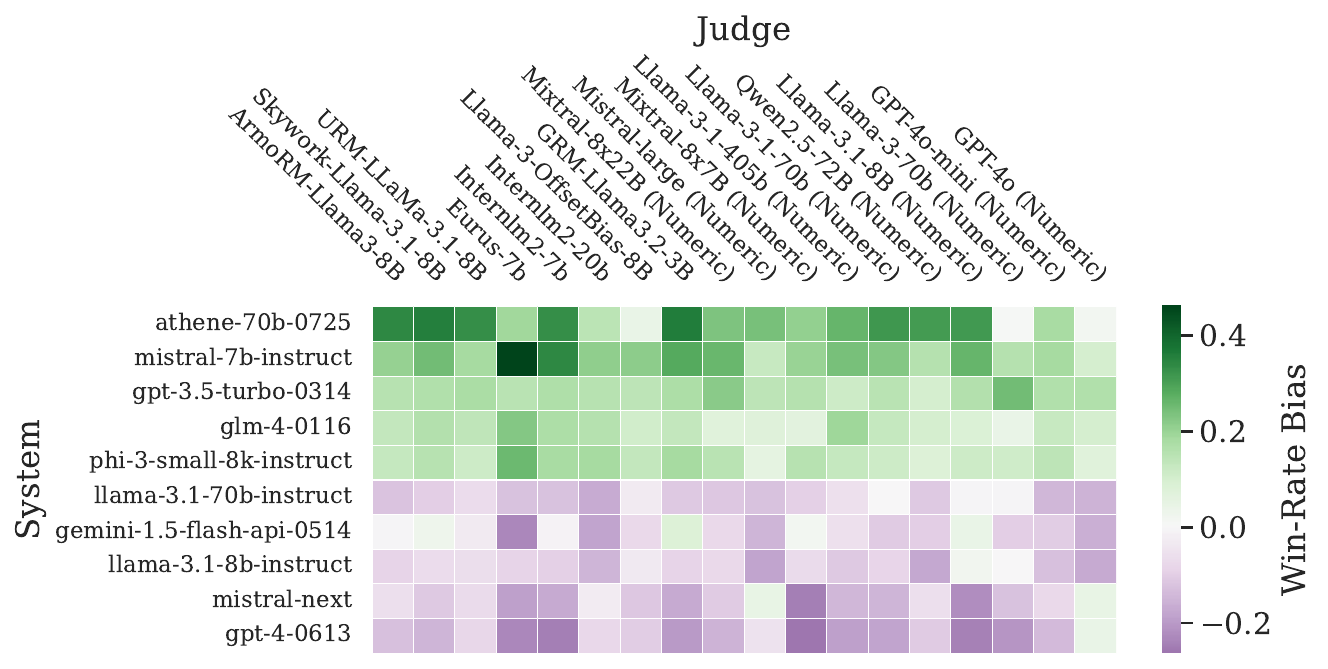} 

\caption{\textbf{System-specific judge biases}. The plot depicts win-rate biases of judges towards specific systems, with respect to the ground-truth win-rates from Chatbot Arena (after correction for the beta distribution fit of each judge). This plot portrays select systems with high bias; the full heat map, including all judge realizations and all systems, is shown in App. Fig.~\ref{fig:all_bias_corrected}.} \label{fig:bias_top}
\end{figure}

%% file: appendix.tex
\newpage



\section{Judge Score Distributions}
\label{app:distributions}

Figure~\ref{fig:score_distributions} depicts the score distributions, $D^p$, of the judges in our data.

\paragraph{Reward model distributions}
The reward models exhibit continuous score distributions. As seen in Figure~\ref{fig:score_distributions}, these distributions vary in the range of scores, as well as in the shape of the distribution. Some reward model judges have a narrow range of scores, e.g., $-0.1$ to $0.4$, whereas in others it is much wider, e.g., $-3000$ to $5000$. Similarly, some distributions are more symmetric while others have peaks at more extreme values. However, all distributions are uni-modal, with a single peak.
Moreover, we note that the continuous nature of these judgment scores also entails an absence of ties between the judged responses.

\paragraph{LLM Numeric distributions} As shown in Figure~\ref{fig:score_distributions}, even though the LLM judges are given a wide range of possible judgment scores ($[0-100]$), in practice they tend to prefer specific score values. This results in many ties when comparing responses from different systems.

\paragraph{LLM Likert distributions} Similarly to the Numeric distributions, the Likert realizations put most of their probability mass on specific scores, which leads to an even greater inclination towards ties (as here they are limited to a smaller range of scores).

\paragraph{LLM TokenProbs distributions} TokenProbs scores tend to be extreme, namely very close to either $0.0$ or $1.0$. Thus, in many cases the score gap between responses is extremely small. This can result in low judge robustness (see the error bars in Figure~\ref{fig:llms}), as well as a higher sensitivity to the choice of aggregation method.

\paragraph{LLM Anchor distributions} The distribution for Anchor judgments is mainly tied to the quality of the anchor system relative to the other systems. However, we see that it is also affected by the characteristics of the judge. For example, we see in Fig.~\ref{fig:score_distributions} that Llama-3.1-8B exhibits indecision, rating most responses as comparable to those of the anchor. In addition, for some judges, the proportion of $-1$ scores (i.e., the response is slightly worse than the anchor) or $1$ scores (the response is slightly better than the anchor) is unusually low.





\section{Aggregation Methods}
\label{app:agg}

Given the raw judgments of each judge, $j_p(R)$, there are multiple aggregation methods, $a$, that construct a \textit{ranking} over all the target systems. Here, we calculate rankings using \textbf{Win-rate} aggregation, \textbf{BT} aggregation, \textbf{Mean} aggregation, and \textbf{Median} aggregation. In the following, we provide further details on each aggregation.

\paragraph{Mean \& Median Aggregation}
These aggregation methods map a score for each system, $s_l$, by relying solely on the scores assigned to its responses by judge $j_p$. In other words, the mapping of $V_l^{p,a}$ by $a$ depends only on the column corresponding to system $s_l$ in $j_p(R)$. Accordingly, these aggregations can be viewed as an operation on the columns of the scores matrix $j_p(R)$. Specifically, for the \textbf{Mean} aggregation, $V_l^{p,a} = \frac{1}{K}\Sigma_{k=1}^K Score^p_{k,l}$. Similarly, \textbf{Median} aggregation is the median of the vector $j_p(R)_{*l}$. 

We note that for realizations with discrete score distributions (see \S\ref{app:distributions}), many systems will likely share the same median score; in this case, the Median aggregation method fails to separate the systems. Hence, Table~\ref{tab:leaderboard_full} contains only a handful of LLM judges with Median aggregation, all using the \textit{TokenProbs} realization.


\paragraph{Win-rate Aggregation}
This aggregation maps each system based on its proportion of wins over other systems, averaged over all instructions $i_k \in I$. Formally, $V_b^{p,a} = \frac{1}{K}\Sigma_{k=1}^K \frac{1}{L-1}\Sigma_{l=1, l \neq b}^L \mathbb{I}(Score^p_{k,b} > Score^p_{k,l})$, where $\mathbb{I}(\cdot)$ denotes the indicator function.

\paragraph{Bradley-Terry Aggregation}
Following \citet{chiang2024chatbot}, we use the vector of Bradley-Terry (BT) coefficients \cite{bradley1952rank} as system scores.

For calculating the BT scores we use the implementation of the Chatbot Arena official notebook\footnote{\label{notebook}\href{https://colab.research.google.com/drive/1KdwokPjirkTmpO_P1WByFNFiqxWQquwH}{Arena official notebook}}.
Whereas \citet{chiang2024chatbot} apply this method for battles between responses with a human judge, we apply it over our LLM-based judge data, i.e., each ``battle'' is a comparison between the judge scores $Score^p_{k,a}$, $Score^p_{k,b}$ for a response generated by systems $s_a$ and $s_b$. 

When there are no ties, e.g., for the reward model judges, this aggregation produces similar rankings to the win-rate aggregation.

\section{Chatbot Arena Data}
\label{app:chatbot}
The data for the Chatbot Arena LLM leaderboard (\url{https://lmarena.ai}) consists of "battles" between systems over the same instructions. In these battles, users indicate a preference (or a tie) between a pair of responses generated by different LLMs~\cite{zheng2023llmaaj,chiang2024chatbot}. 

We use their public data file from August 2024\footnote{\href{https://storage.googleapis.com/arena_external_data/public/clean_battle_20240814_public.json}{Chatbot Arena data}}, and follow the official notebook\footnotemark[\getrefnumber{notebook}]
to extract the raw data, deduplicate it, and calculate the overall system rankings. This dataset includes the human preference judgments and names of the participating systems, but not the instructions or system responses for the battles.

Here we limit the analysis to the \textit{English Hard Prompts} subset of their data\footnote{\href{https://lmsys.org/blog/2024-05-17-category-hard/}{Chatbot Arena Hard Prompts}} ($300$K battles). 
Notably, Arena Hard was specifically designed to match the distribution of user instructions in the \textit{English Hard Prompts} subset, as described by \citet{li2024crowdsourced}.
We follow their code to construct a full system ranking based on these $300$K battles, using Bradley-Terry coefficients. This yields a score for each system in their data, including $59$ systems that are also in our system responses data (\S\ref{ssec:arena_hard})

Out of this full English Hard data, we also extract a total of $113$K battles that were not judged by humans as ties, and 
that are between pairs of systems which
appear in our responses data. We then use those to calculate win-rates between pairs of systems (\S\ref{app:winrate}), yielding a total of $968$ system pairwise win-rates. Note that the Chatbot Arena data does not contain battles between every possible pairing of systems, and thus we do not have win-rates for all combinations of the $59$ systems under consideration. In addition, we limit the analysis to system pairs with at least $10$ non-tied battles.

\section{Statistical Analysis of Judge Performance}
\label{app:stats}
In \S\ref{sec:benchmark} and Table~\ref{tab:leaderboard_full} we report results of agreement with the gold ranking ($\tau$) for various judge pipelines. Each pipeline consists of a chosen judge model, a realization (\S\ref{ssec:realizations}) and an aggregation method (\S\ref{ssec:aggregations}, App.~\ref{app:agg}). 

We focus on the LLM judges and perform a three-way ANOVA (analysis of variance), with the ranking correlation $\tau$ as a dependent variable and the \textit{model}, \textit{realization} and \textit{aggregation} as factors. In addition to the variance analysis estimating the effects of these factors, we perform post-hoc pairwise comparisons to ask whether certain configurations (i.e., a specific realization/aggregation) outperform the others. We conduct all analyses using IBM SPSS Statistics v30.0.

The ANOVA shows that both the judge model and the realization have a strong influence on $\tau$, with an effect size (\textit{Partial Eta-Squared}) of $\eta^2=0.81$ for the judge model ($p<0.001$; $F=36.0$), $\eta^2=0.51$ for the realization ($p<0.001$; $F=26.6$), and $\eta^2=0.78$ for the interaction effect between model and realization ($p<0.001$; $F=10.1$). In contrast, the aggregation methods were not found to have a significant effect on $\tau$ ($\eta^2=0.02$; $p>0.5$).

We also perform Tukey's HSD \cite{tukey1949comparing} post-hoc tests to compare the means of the variables. The analysis indicates that the both the Numeric (mean $\tau = 0.75$; $\sigma_{\tau} = 0.06$) and Likert ($\tau = 0.74$; $\sigma_{\tau} = 0.07$) realizations are significantly better than the Anchor ($\tau = 0.71$; $\sigma_{\tau} = 0.07$) and TokenProbs ($\tau = 0.68$; $\sigma_{\tau} = 0.13$) realizations (all $p$ values $<=0.002$). The differences between aggregation methods are not statistically significant.

\section{Pairwise Win-Rates}
\label{app:winrate}
We denote the win-rate of system $s_a$ over system $s_b$ as $WR(s_a, s_b)^p$ where $p$ denotes the judge upon which the win-rate was calculated, and $p \in J\cup \{g\}$, where $g$ stands for human gold data. 

The win-rate of system $s_a$ over system $s_b$ according to judge $j_p$ over the set of instances $I$ is calculated as the proportion of instances where the score given by $j_p$ to the response generated by $s_a$ surpasses that of system $s_b$, where ties are excluded. Namely $WR^p(s_a, s_b) = \frac{1}{K-|T^p_{a,b}|}\Sigma_{k=1}^K \mathbb{I}( Score^p_{k,a} > Score^p_{k,b})$ Where $T^p_{a,b}=\{i_k| Score^p_{k,a} = Score^p_{k,b}\}$, and $\mathbb{I}(\cdot)$ denotes the indicator function. 
Notice that $WR^p(s_a,s_b) = 1-WR^p(s_b,s_a)$.




To quantify the agreement between the judge and gold win-rates we also define an \textit{Accuracy} metric. This measures the proportion of pairs where the judge pairwise system preference
decisions are in agreement with those of the human gold-data. In other words, we want to count the pairs that appear in the first and third quadrants in Figure~\ref{fig:pred_wr_examples}; namely, the pairs where the judge and gold win-rate are both bigger than $0.5$, or the pairs where both are lower than $0.5$, representing agreement on the winning system.
For that, we denote all the pairs of systems
we have in the gold data as $\{s_{a^m}, s_{b^m}\}_{m=1}^M$.
Now the \textit{Accuracy} is defined as follows:
\begin{align*}
    Acc_{WR}^{p} = \frac{1}{M} \Sigma_{m=1}^{M} \mathbb{I}( \mathbb{I}( WR^p(s_{a^m},s_{b^m})>0.5) \\ = \mathbb{I}( WR^g(s_{a^m},s_{b^m})>0.5) )
\end{align*}
Additionally, we define a second metric, the \textit{Mean Squared Error} over all win-rate pairs.
\begin{align*}
MSE_{WR}^{m} = \frac{1}{M} \Sigma_{m=1}^{M} (WR^g(s_{a^m},s_{b^m}) \\- WR^p(s_{a^m},s_{b^m}))^2.
\end{align*}

The $Acc_{WR}^{p}$ scores are in high agreement with the \benchmark{} judge ranking quality scores $\tau$ (Pearson correlation of $r=0.96$ for the BT aggregation, $r=0.79$ for the Mean aggregation). 
This highlights the direct link between judges' ability to rank systems and their performance on pairwise system preference. 

The $MSE_{WR}^{p}$ scores have a low correlation with the \benchmark{} judge $\tau$ scores ($r=-0.19$ for the BT aggregation, $r=-0.07$ for the Mean aggregation). This can be explained by the decisiveness effect (\S\ref{ssec:exaggeration}), where judges deviate substantially from the gold win-rate, but mostly toward the stronger system in the pair.





\section{Beta Distribution Fit}
\label{app:beta}

Following \citet{kull2017beta}, we model the relation between judge and gold win-rates using the cumulative distribution function (CDF) of the Beta distribution. We parameterize the distribution such that both shape parameters $\alpha$ and $\beta$ are equal ($\alpha = \beta$). 

The CDF of the Beta distribution, defined over the interval $[0, 1]$, for $\alpha = \beta \in [0, \infty]$ provides a wide range of function fits: a linear $y=x$ fit for $\alpha = 1$, a sigmoidal fit for larger $\alpha$ values, and approaching a step function as $\alpha \to \infty$. These attributes make it particularly suited for our data characteristics.

Given a set of data points $\{(WR^p(s_{a^m},s_{b^m}), WR^g(s_{a^m},s_{b^m})\}_{m=1}^M$, where $WR^p(s_{a^m},s_{b^m}) \in [0, 1]$ represents the judge win-rate and $WR^g(s_{a^m},s_{b^m}) \in [0, 1]$ denotes the gold win-rate between system, $s_{a^m}$ and $s_{b^m}$. We fit the Beta CDF by optimizing the shape parameter $\alpha$. The optimization objective is minimizing the sum of absolute errors (SAE) between the judge win-rate, $WR^p(s_{a^m},s_{b^m})$, and the predicted values from the Beta CDF. In order to capture the behavior across the entire range of win-rates, we weight the errors by the distance of $WR^p$ from $0.5$:


\begin{align*}
\text{SAE} = \sum_{m=1}^M 
\gamma(WR^p(s_{a^m}, s_{b^m})) \cdot \bigg| WR^p(s_{a^m}, s_{b^m}) \\
    - F_{\text{Beta}}(WR^g(s_{a^m}, s_{b^m}); \alpha) \bigg|
\end{align*}


where $F_{\text{Beta}}(x; \alpha)$ denotes the Beta CDF with shape parameters $\alpha = \beta$, and $\gamma$ is the distance of $WR^p$ from $0.5$.  

The optimization was performed using the \texttt{scipy.optimize.minimize}\footnote{\href{https://docs.scipy.org/doc/scipy/reference/generated/scipy.optimize.minimize.html}{SciPy Documentation for scipy.optimize.minimize}} function, with the parameter ($\alpha$) constrained to a reasonable range \([0.1, 10000]\).
This approach efficiently identified the best-fit parameter ($\alpha$).

The resulting Beta CDF closely captures the empirical data distribution, as validated both quantitatively, through low SAE, and qualitatively via visual inspection. Figure~\ref{fig:all_betas} depicts the fitted Beta CDF curve and the observed data points, demonstrating the effectiveness of this approach for modeling the judges' predicted win-rate distribution.

\onecolumn
\section{LLM Judge Prompts}
\label{app:prompts}
Below we list the prompts we use for each LLM judge realization (\S\ref{ssec:realizations}).

\begin{tcolorbox}[title=Numeric,float,floatplacement=h!,fontupper=\linespread{1.1}\fontfamily{lmr}\selectfont]
Here is a user input and a model response. On a scale of $0$ to $100$, to what extent is this a good response for the given input? Reply with your rating score without any preceding explanation. Input: \textit{[user instruction]}

Response: \textit{[system response]}

Rating ($0$-$100$):
\end{tcolorbox}

\begin{tcolorbox}[title=Likert,float,floatplacement=h!,fontupper=\linespread{1.1}\fontfamily{lmr}\selectfont]
Here is a user input and a model response. To what extent is this a good response for the given input? Provide a rating from one of the following choices: 'Very Bad', 'Bad', 'Mediocre', 'Good', 'Very Good'. Reply using the format of [[rating]], for example: '[[Mediocre]]'

Input: \textit{[user instruction]}

Response: \textit{[system response]}

Rating:
\end{tcolorbox}

\begin{tcolorbox}[title=TokenProbs,float,floatplacement=ht,fontupper=\linespread{1.1}\fontfamily{lmr}\selectfont]
Here is a user input and a model response. Is this a good response for the given input? Answer with only yes/no. Input: \textit{[user instruction]}

Response: \textit{[system response]}

Good response? (Yes/No):
\end{tcolorbox}

\begin{tcolorbox}[title=Anchor,float,floatplacement=h,fontupper=\linespread{1.1}\fontfamily{lmr}\selectfont]
Here is a user input and responses from two assistants, A and B. Which response is better? You must output only one of the following choices as your final verdict with a label:

\begin{enumerate}

\item Assistant A is significantly better: [[A>{}>B]]

\item Assistant A is slightly better: [[A>B]]

\item Tie, relatively the same: [[A=B]]

\item Assistant B is slightly better: [[B>A]]

\item Assistant B is significantly better: [[B>{}>A]
\end{enumerate}

Example output: "My final verdict is tie: [[A=B]]".
\\
\\
<|User Prompt|>

\textit{[user instruction]}
\\
\\
<|The Start of Assistant A's Answer|>

\textit{[system response]}

<|The End of Assistant A's Answer|>
\\
\\
<|The Start of Assistant B's Answer|>

\textit{[anchor system response]}

<|The End of Assistant B's Answer|>

Final Verdict:
\end{tcolorbox}

\input{full_leaderboard}

\begin{figure*}[h]
\subfloat{\includegraphics[width=0.5\textwidth]{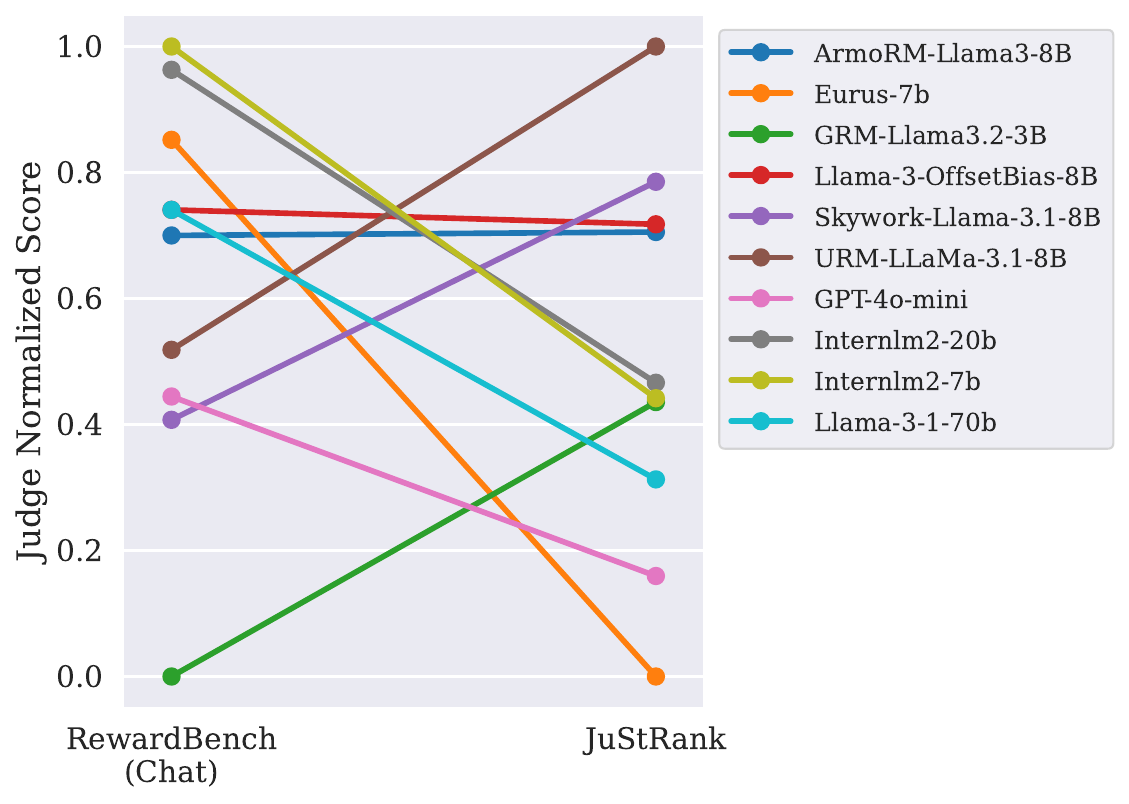}}
\subfloat{\includegraphics[width=0.5\textwidth]{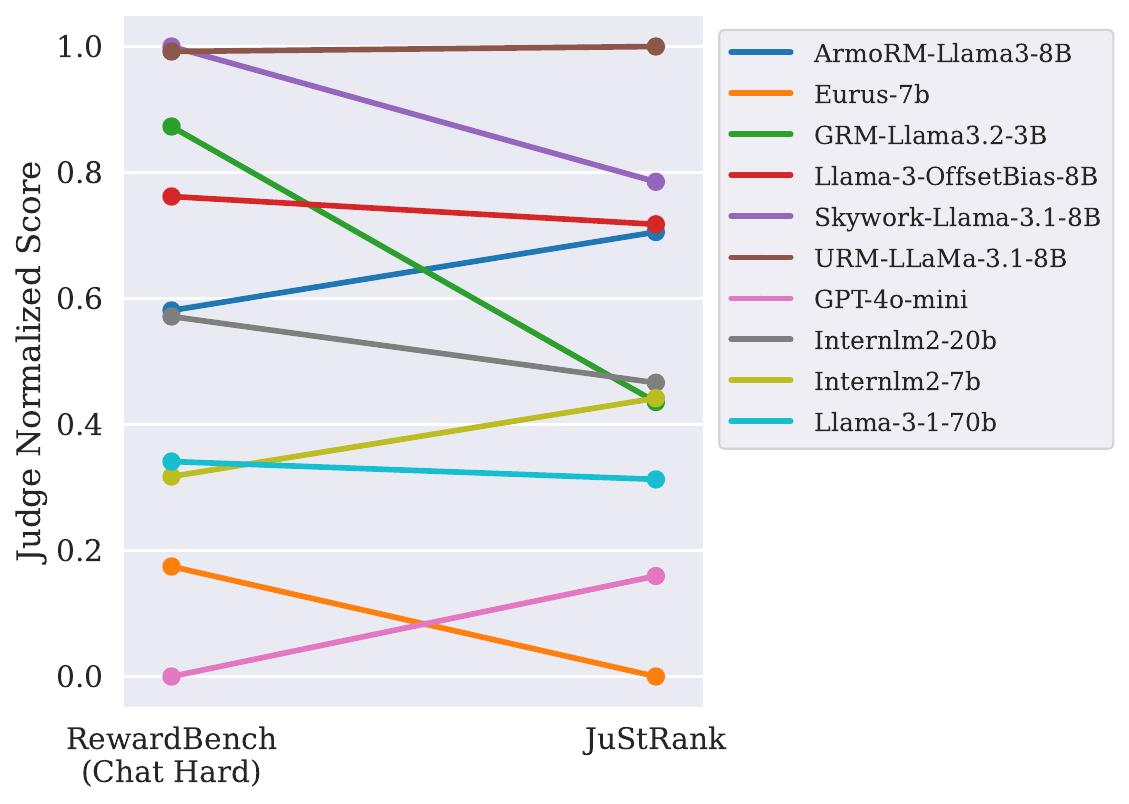}}

\subfloat{\includegraphics[width=0.5\textwidth]{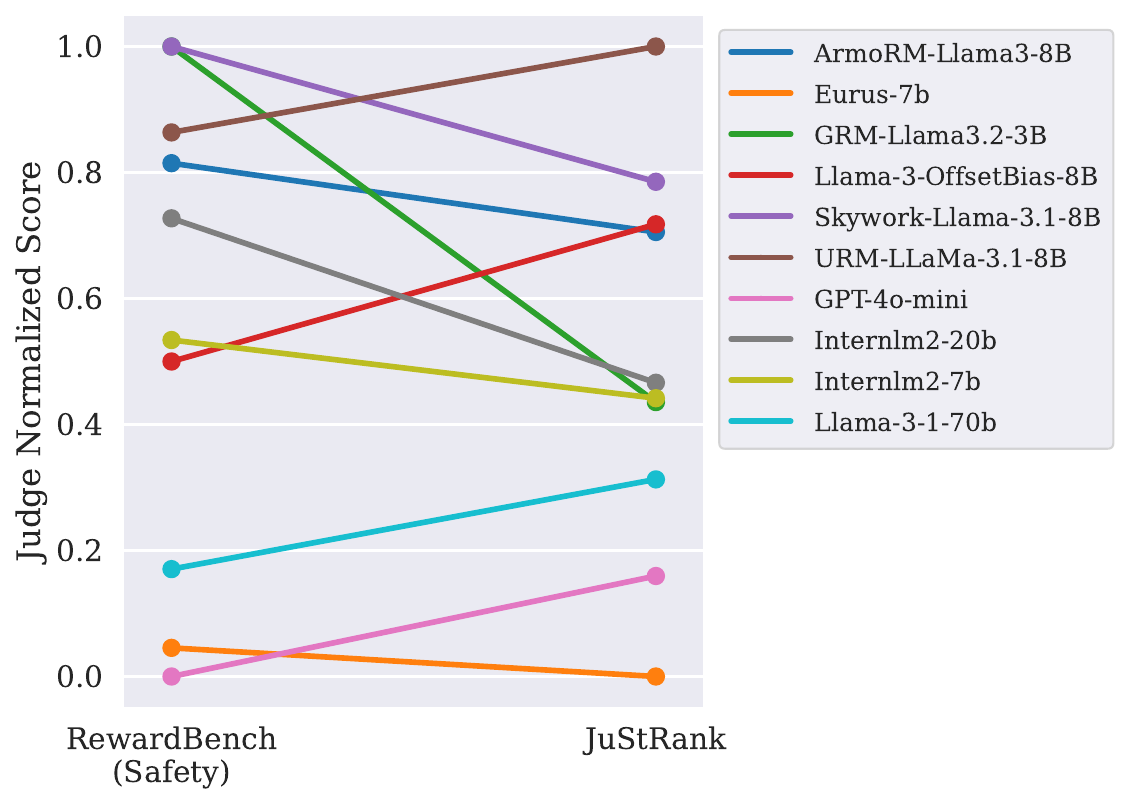}}
\subfloat{\includegraphics[width=0.5\textwidth]{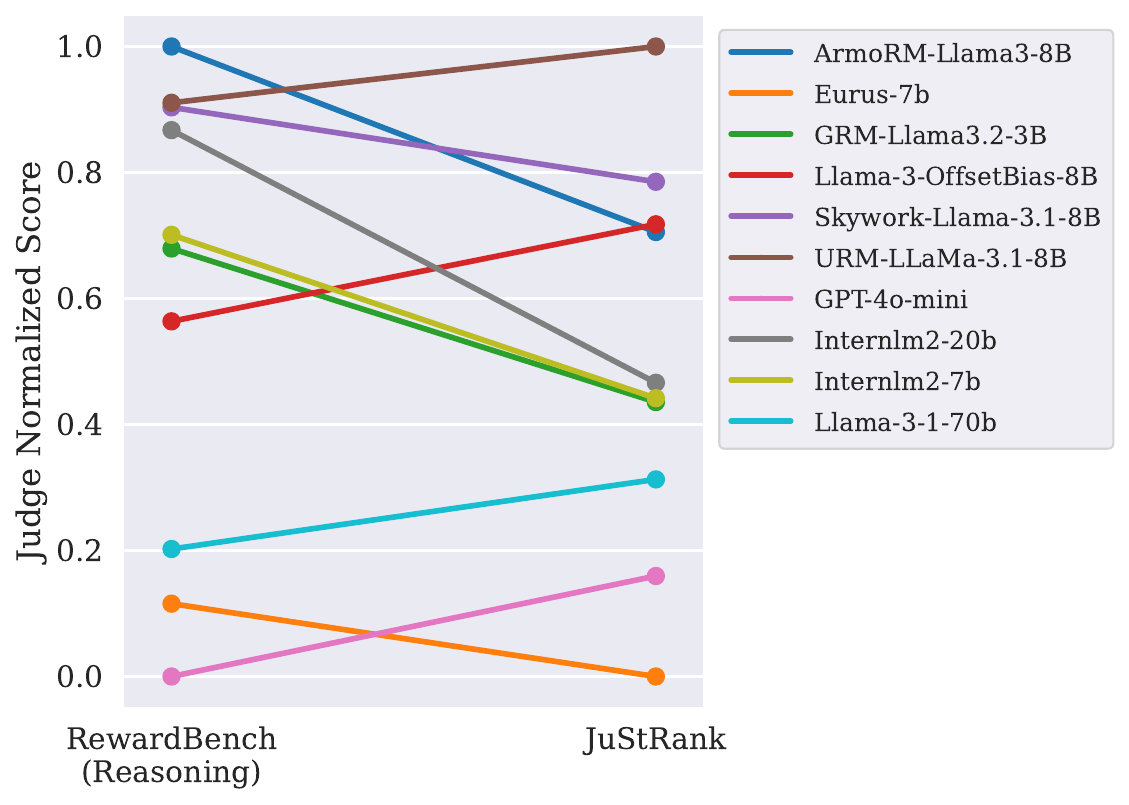}}

    \caption{\textbf{Comparison to RewardBench}. The plot depicts the relative performance of judges present in both \benchmark{} and RewardBench~\cite{lambert2024rewardbench}. For comparison, we perform Min-Max normalization over the judge performance scores (\textit{accuracy} for RewardBench, \textit{Kendall's Tau} for our results). The results shown are for the BT aggregation method; the LLM judges use the \textit{Anchor} realization, which is closest to the setting in RewardBench. Each panel portrays a different subset of RewardBench.}
    \label{fig:rewardbench_subsets}
\end{figure*}

\begin{figure*}[t]
\includegraphics[width=1\textwidth]{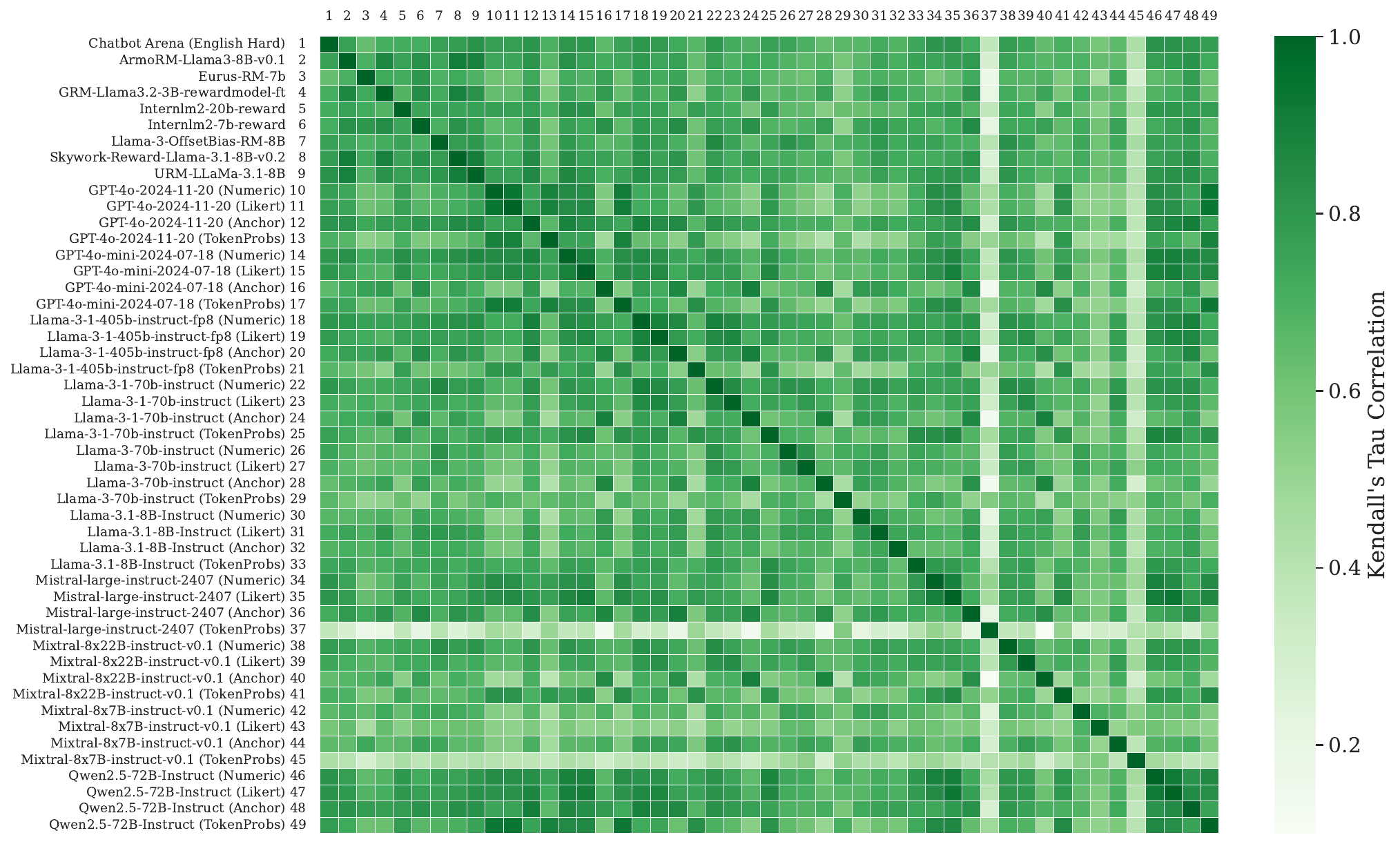}

    \caption{\textbf{Judge Correlations}. Kendall's Tau correlations between the system rankings produced by the different judge realizations, using the BT aggregation method. The first row/column denotes correlations with the reference ranking from Chatbot Arena.}
    \label{fig:judge_corrs}
\end{figure*}

\begin{figure*}[t]
\centering
\subfloat[]{\includegraphics[width=.73\textwidth]{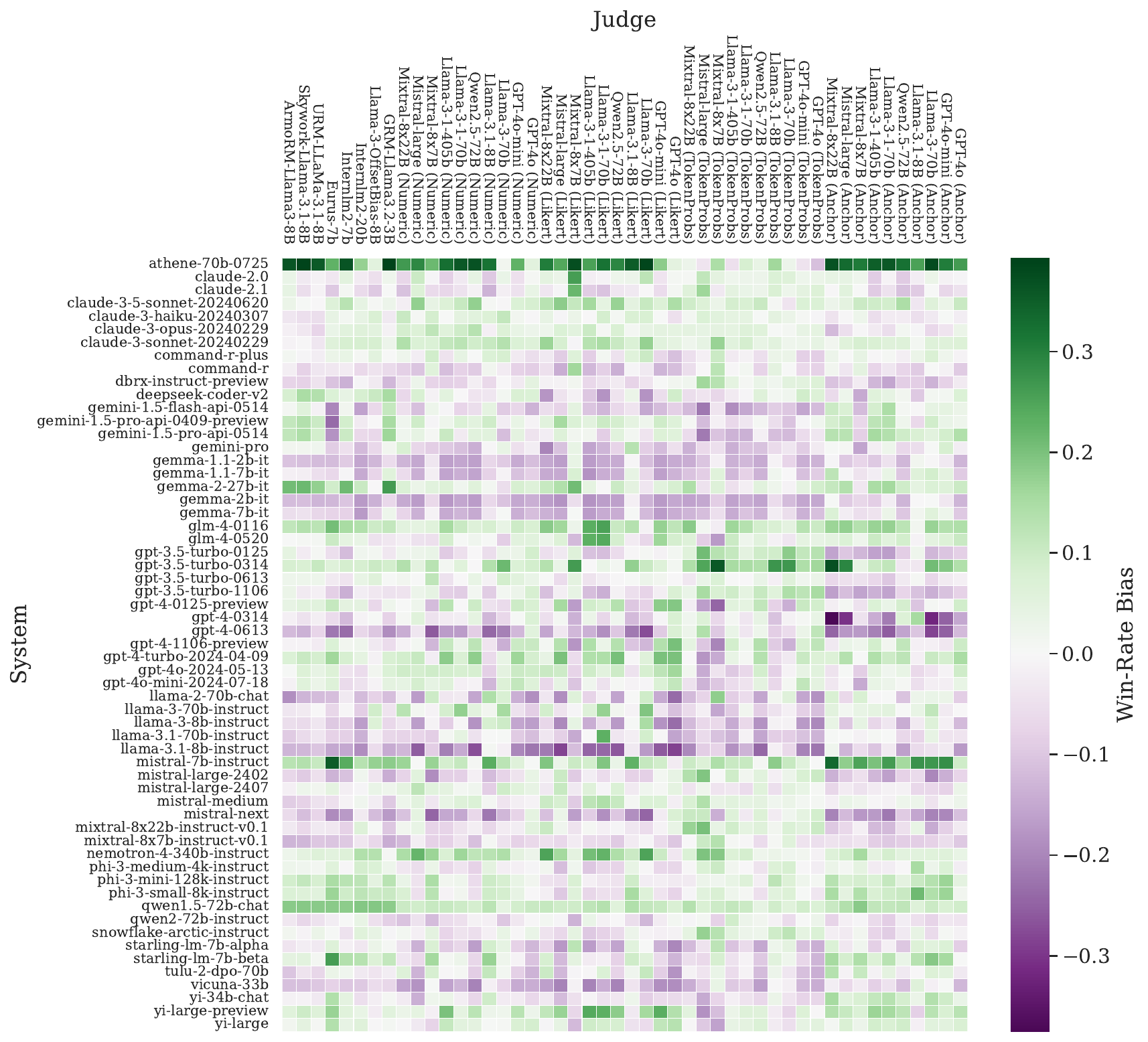} \label{fig:all_bias}}
\\
\subfloat[]{\includegraphics[width=.73\textwidth]{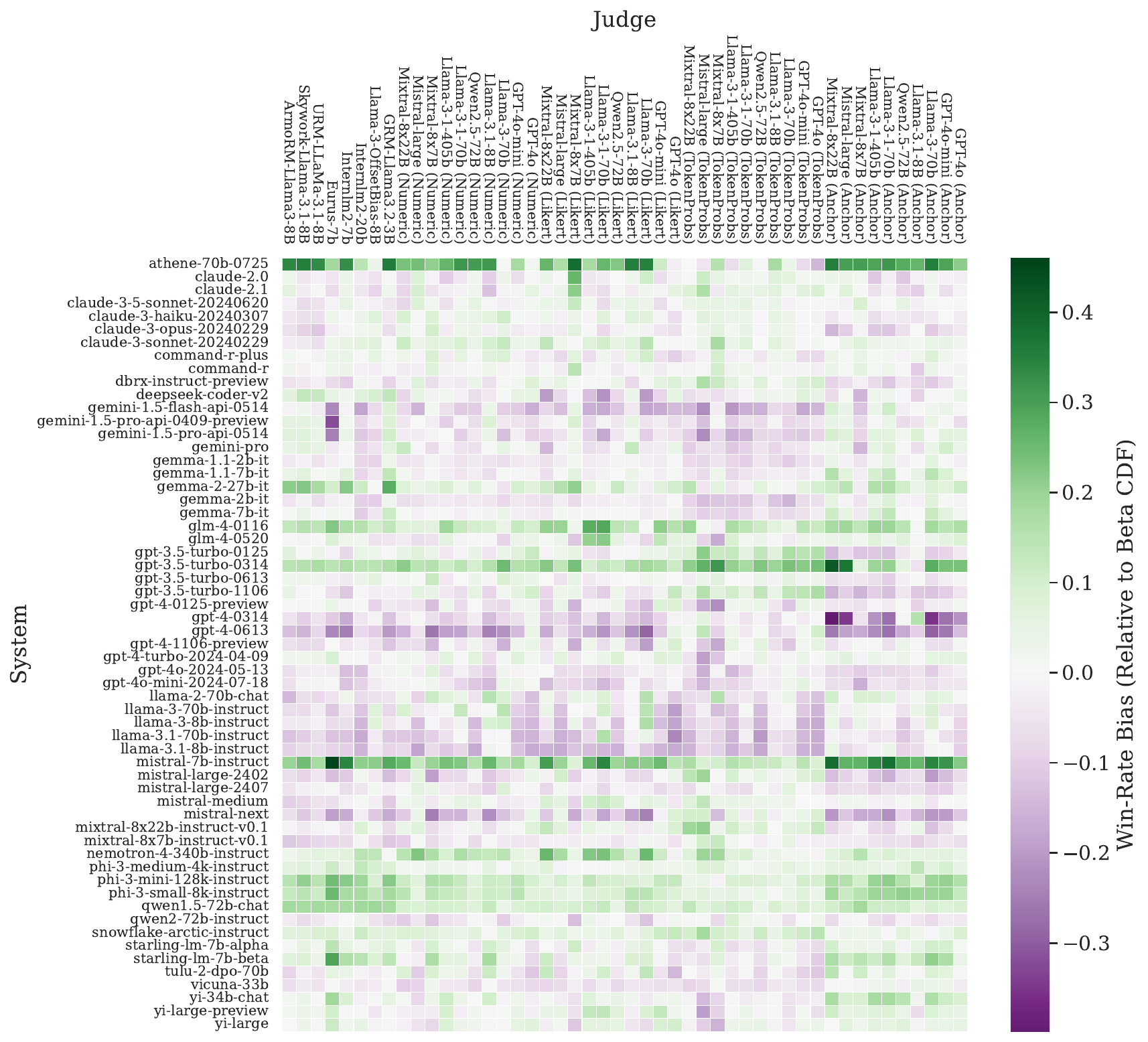} \label{fig:all_bias_corrected}}

\caption{\textbf{System-specific judge biases}. The heat maps depict the win-rate biases of various judges towards specific systems (\S\ref{ssec:bias}), with respect to the ground-truth win-rates from Chatbot Arena. (a): Bias w.r.t. the raw ground-truth win-rates $WR^g$; (b): Bias w.r.t. the fit value for the gold win-rate $WR^{g'}$ on the beta distribution fit (App.~\ref{app:beta}) for each judge.}
\end{figure*}

\input{full_betas}

\input{self_bias_table}

\input{distributions_fig}

\input{detailed_leaderboard}

%% file: full_leaderboard.tex
\onecolumn
\begin{longtable}{| p{.35\textwidth} | p{.15\textwidth} |
p{.15\textwidth} |
c |}
\toprule
Judge Model & Realization & Aggregation & Agreement ($\tau$) \\
& & & w/ Gold Ranking \\
\midrule
Qwen2.5-72B-Instruct & Likert & Win-Rate & .827 \\
URM-LLaMa-3.1-8B & Reward & Mean & .823 \\
GPT-4o-2024-11-20 & Anchor & Mean & .822 \\
URM-LLaMa-3.1-8B & Reward & BT & .819 \\
Qwen2.5-72B-Instruct & Likert & BT & .817 \\
URM-LLaMa-3.1-8B & Reward & Win-Rate & .816 \\
Qwen2.5-72B-Instruct & Numeric & BT & .814 \\
GPT-4o-2024-11-20 & Anchor & Win-Rate & .814 \\
Qwen2.5-72B-Instruct & Numeric & Win-Rate & .813 \\
Llama-3-1-405b-instruct-fp8 & Numeric & Mean & .812 \\
Llama-3-1-405b-instruct-fp8 & Numeric & Win-Rate & .812 \\
Mistral-large-instruct-2407 & Likert & BT & .811 \\
GPT-4o-2024-11-20 & Anchor & BT & .809 \\
Mistral-large-instruct-2407 & Numeric & BT & .809 \\
URM-LLaMa-3.1-8B & Reward & Median & .809 \\
GPT-4o-mini-2024-07-18 & Numeric & Win-Rate & .807 \\
Llama-3-1-405b-instruct-fp8 & Numeric & BT & .805 \\
GPT-4o-mini-2024-07-18 & Numeric & BT & .804 \\
Mistral-large-instruct-2407 & Numeric & Win-Rate & .802 \\
Qwen2.5-72B-Instruct & Likert & Mean & .801 \\
ArmoRM-Llama3-8B-v0.1 & Reward & Mean & .800 \\
Qwen2.5-72B-Instruct & Anchor & Mean & .799 \\
GPT-4o-mini-2024-07-18 & Likert & BT & .798 \\
Llama-3-1-70b-instruct & Numeric & Win-Rate & .798 \\
Llama-3-1-70b-instruct & Numeric & BT & .798 \\
Mistral-large-instruct-2407 & Likert & Win-Rate & .798 \\
Qwen2.5-72B-Instruct & Anchor & BT & .794 \\
Llama-3-1-405b-instruct-fp8 & Likert & Win-Rate & .793 \\
Llama-3-1-70b-instruct & TokenProbs & Win-Rate & .793 \\
GPT-4o-mini-2024-07-18 & Likert & Win-Rate & .793 \\
ArmoRM-Llama3-8B-v0.1 & Reward & Median & .793 \\
Llama-3-1-405b-instruct-fp8 & Likert & BT & .787 \\
Mistral-large-instruct-2407 & Anchor & Win-Rate & .786 \\
Skywork-Llama-3.1-8B-v0.2 & Reward & Mean & .786 \\
Qwen2.5-72B-Instruct & Anchor & Win-Rate & .786 \\
Mistral-large-instruct-2407 & Likert & Mean & .782 \\
GPT-4o-mini-2024-07-18 & Numeric & Mean & .781 \\
Skywork-Llama-3.1-8B-v0.2 & Reward & Win-Rate & .780 \\
Llama-3-1-405b-instruct-fp8 & Likert & Mean & .780 \\
Skywork-Llama-3.1-8B-v0.2 & Reward & BT & .778 \\
Llama-3.1-8B-Instruct & TokenProbs & Mean & .778 \\
Qwen2.5-72B-Instruct & TokenProbs & BT & .777 \\
Llama-3.1-8B-Instruct & TokenProbs & Median & .776 \\
Mixtral-8x22B-instruct-v0.1 & Numeric & BT & .776 \\
Llama-3-1-70b-instruct & TokenProbs & Median & .776 \\
GPT-4o-2024-11-20 & Numeric & BT & .774 \\
GPT-4o-mini-2024-07-18 & Likert & Mean & .773 \\
Qwen2.5-72B-Instruct & Numeric & Mean & .773 \\
GPT-4o-2024-11-20 & Likert & BT & .773 \\
GPT-4o-2024-11-20 & Numeric & Win-Rate & .771 \\
Llama-3-OffsetBias-RM-8B & Reward & Win-Rate & .765 \\
Llama-3-1-70b-instruct & TokenProbs & BT & .765 \\
Llama-3-OffsetBias-RM-8B & Reward & BT & .765 \\
Skywork-Llama-3.1-8B-v0.2 & Reward & Median & .764 \\
Llama-3-1-70b-instruct & TokenProbs & Mean & .764 \\
Mistral-large-instruct-2407 & Anchor & Mean & .764 \\
Llama-3-1-70b-instruct & Numeric & Mean & .764 \\
ArmoRM-Llama3-8B-v0.1 & Reward & BT & .763 \\
ArmoRM-Llama3-8B-v0.1 & Reward & Win-Rate & .762 \\
Llama-3-OffsetBias-RM-8B & Reward & Median & .759 \\
GPT-4o-mini-2024-07-18 & TokenProbs & Win-Rate & .759 \\
GPT-4o-2024-11-20 & Likert & Win-Rate & .758 \\
Llama-3-OffsetBias-RM-8B & Reward & Mean & .757 \\
Mixtral-8x22B-instruct-v0.1 & Numeric & Win-Rate & .756 \\
GPT-4o-mini-2024-07-18 & TokenProbs & BT & .752 \\
Qwen2.5-72B-Instruct & TokenProbs & Median & .752 \\
Mistral-large-instruct-2407 & Numeric & Mean & .750 \\
Llama-3-70b-instruct & Numeric & BT & .749 \\
Qwen2.5-72B-Instruct & TokenProbs & Win-Rate & .748 \\
Llama-3-1-405b-instruct-fp8 & Anchor & Win-Rate & .748 \\
Llama-3-1-70b-instruct & Likert & Mean & .746 \\
GPT-4o-2024-11-20 & Likert & Mean & .744 \\
Llama-3.1-8B-Instruct & TokenProbs & Win-Rate & .744 \\
Llama-3-1-405b-instruct-fp8 & Anchor & Mean & .744 \\
Llama-3.1-8B-Instruct & TokenProbs & BT & .741 \\
Llama-3-1-405b-instruct-fp8 & TokenProbs & Win-Rate & .741 \\
GPT-4o-mini-2024-07-18 & TokenProbs & Mean & .741 \\
Mixtral-8x22B-instruct-v0.1 & Likert & BT & .738 \\
GPT-4o-2024-11-20 & Numeric & Mean & .738 \\
Llama-3-1-405b-instruct-fp8 & TokenProbs & Median & .737 \\
Llama-3.1-8B-Instruct & Likert & Mean & .736 \\
Llama-3-70b-instruct & Numeric & Win-Rate & .733 \\
Llama-3-1-405b-instruct-fp8 & TokenProbs & Mean & .733 \\
Llama-3-1-70b-instruct & Likert & Win-Rate & .732 \\
Mixtral-8x22B-instruct-v0.1 & Likert & Win-Rate & .732 \\
Qwen2.5-72B-Instruct & TokenProbs & Mean & .732 \\
Internlm2-7b-reward & Reward & Mean & .731 \\
Llama-3-1-405b-instruct-fp8 & Anchor & BT & .730 \\
Mistral-large-instruct-2407 & TokenProbs & Mean & .730 \\
Internlm2-20b-reward & Reward & Mean & .728 \\
Mistral-large-instruct-2407 & Anchor & BT & .725 \\
Internlm2-20b-reward & Reward & Median & .724 \\
GPT-4o-mini-2024-07-18 & TokenProbs & Median & .723 \\
Llama-3.1-8B-Instruct & Likert & BT & .723 \\
Llama-3-1-70b-instruct & Likert & BT & .722 \\
Internlm2-7b-reward & Reward & Median & .721 \\
Mixtral-8x22B-instruct-v0.1 & Likert & Mean & .719 \\
Internlm2-7b-reward & Reward & Win-Rate & .717 \\
Internlm2-20b-reward & Reward & BT & .717 \\
Mixtral-8x22B-instruct-v0.1 & TokenProbs & Win-Rate & .717 \\
Llama-3-1-70b-instruct & Anchor & Win-Rate & .716 \\
GRM-Llama3.2-3B & Reward & Mean & .716 \\
Internlm2-20b-reward & Reward & Win-Rate & .716 \\
Mixtral-8x22B-instruct-v0.1 & Numeric & Mean & .715 \\
Llama-3-1-70b-instruct & Anchor & Mean & .714 \\
GRM-Llama3.2-3B & Reward & Win-Rate & .712 \\
Internlm2-7b-reward & Reward & BT & .712 \\
GRM-Llama3.2-3B & Reward & BT & .711 \\
GRM-Llama3.2-3B & Reward & Median & .706 \\
GPT-4o-2024-11-20 & TokenProbs & Median & .704 \\
Llama-3-70b-instruct & Numeric & Mean & .704 \\
Mixtral-8x22B-instruct-v0.1 & TokenProbs & BT & .702 \\
GPT-4o-2024-11-20 & TokenProbs & Mean & .701 \\
GPT-4o-2024-11-20 & TokenProbs & BT & .700 \\
Llama-3-70b-instruct & Likert & BT & .698 \\
Llama-3-70b-instruct & TokenProbs & Win-Rate & .696 \\
GPT-4o-2024-11-20 & TokenProbs & Win-Rate & .696 \\
Llama-3.1-8B-Instruct & Anchor & Mean & .695 \\
Llama-3.1-8B-Instruct & Likert & Win-Rate & .694 \\
Llama-3-1-70b-instruct & Anchor & BT & .688 \\
Llama-3-70b-instruct & Likert & Win-Rate & .681 \\
Llama-3.1-8B-Instruct & Numeric & Mean & .680 \\
Llama-3-70b-instruct & Likert & Mean & .678 \\
Llama-3.1-8B-Instruct & Anchor & BT & .677 \\
GPT-4o-mini-2024-07-18 & Anchor & Mean & .675 \\
Llama-3-1-405b-instruct-fp8 & TokenProbs & BT & .672 \\
Llama-3.1-8B-Instruct & Numeric & BT & .668 \\
GPT-4o-mini-2024-07-18 & Anchor & Win-Rate & .668 \\
Llama-3-70b-instruct & Anchor & Mean & .667 \\
Llama-3-70b-instruct & TokenProbs & Mean & .666 \\
Mixtral-8x22B-instruct-v0.1 & Anchor & Mean & .665 \\
Llama-3-70b-instruct & TokenProbs & BT & .663 \\
GPT-4o-mini-2024-07-18 & Anchor & BT & .659 \\
Mixtral-8x7B-instruct-v0.1 & Numeric & BT & .656 \\
Mixtral-8x7B-instruct-v0.1 & Anchor & BT & .655 \\
Mixtral-8x22B-instruct-v0.1 & TokenProbs & Mean & .650 \\
Eurus-RM-7b & Reward & Median & .643 \\
Eurus-RM-7b & Reward & Mean & .641 \\
Mixtral-8x22B-instruct-v0.1 & Anchor & BT & .641 \\
Llama-3.1-8B-Instruct & Anchor & Win-Rate & .639 \\
Llama-3-70b-instruct & Anchor & Win-Rate & .638 \\
Llama-3-70b-instruct & Anchor & BT & .633 \\
Llama-3.1-8B-Instruct & Numeric & Win-Rate & .632 \\
Eurus-RM-7b & Reward & Win-Rate & .629 \\
Eurus-RM-7b & Reward & BT & .628 \\
Mixtral-8x7B-instruct-v0.1 & Numeric & Win-Rate & .626 \\
Mixtral-8x7B-instruct-v0.1 & Numeric & Mean & .626 \\
Mixtral-8x7B-instruct-v0.1 & Anchor & Win-Rate & .622 \\
Mixtral-8x22B-instruct-v0.1 & Anchor & Win-Rate & .612 \\
Mixtral-8x7B-instruct-v0.1 & Anchor & Mean & .610 \\
Mixtral-8x7B-instruct-v0.1 & Likert & BT & .590 \\
Mixtral-8x7B-instruct-v0.1 & Likert & Mean & .585 \\
Mixtral-8x7B-instruct-v0.1 & Likert & Win-Rate & .543 \\
Mixtral-8x7B-instruct-v0.1 & TokenProbs & BT & .427 \\
Mistral-large-instruct-2407 & TokenProbs & Win-Rate & .417 \\
Mixtral-8x7B-instruct-v0.1 & TokenProbs & Mean & .411 \\
Mixtral-8x7B-instruct-v0.1 & TokenProbs & Win-Rate & .371 \\
Mistral-large-instruct-2407 & TokenProbs & BT & .369 \\
Mistral-large-instruct-2407 & TokenProbs & Median & .363 \\
\bottomrule
\caption{\textbf{Judges by ranking performance}. The judges are sorted by the Kendall's Tau correlation between their overall system ranking and the gold ranking from Chatbot Arena (\S\ref{ssec:chatbot}).}  \label{tab:leaderboard_full}
\end{longtable}

%% file: full_betas.tex
\begin{figure*}[t]
\centering
\subfloat{\includegraphics[width=.32\columnwidth]{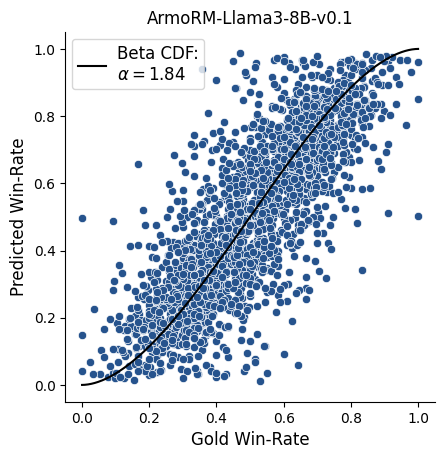}}
\subfloat{\includegraphics[width=.32\columnwidth]{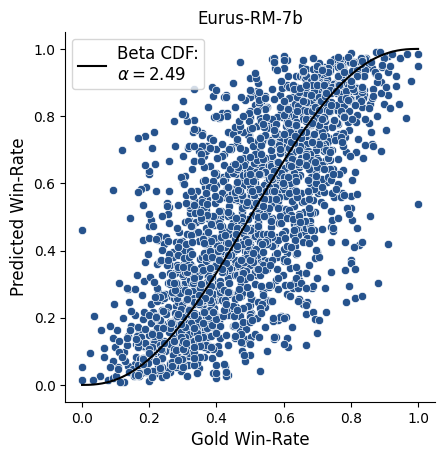}}
\subfloat{\includegraphics[width=.32\columnwidth]{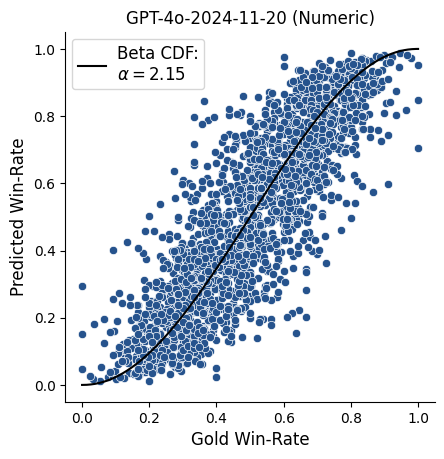}}

\subfloat{\includegraphics[width=.32\columnwidth]{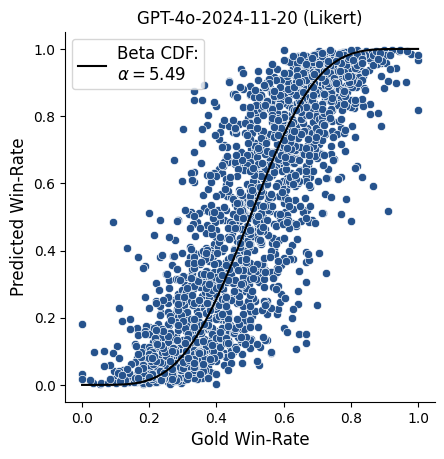}}
\subfloat{\includegraphics[width=.32\columnwidth]{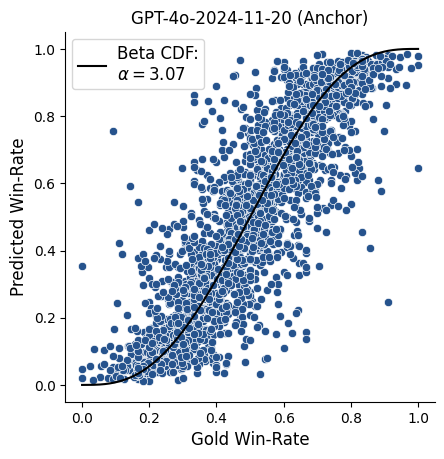}}
\subfloat{\includegraphics[width=.32\columnwidth]{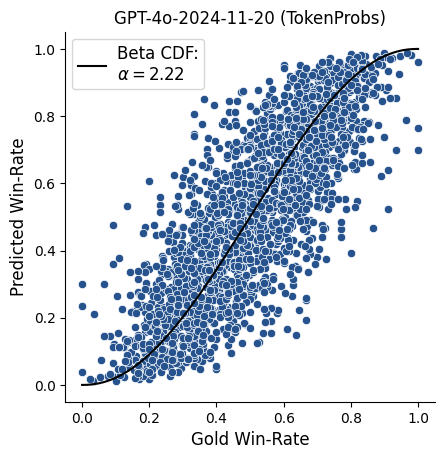}}

\subfloat{\includegraphics[width=.32\columnwidth]{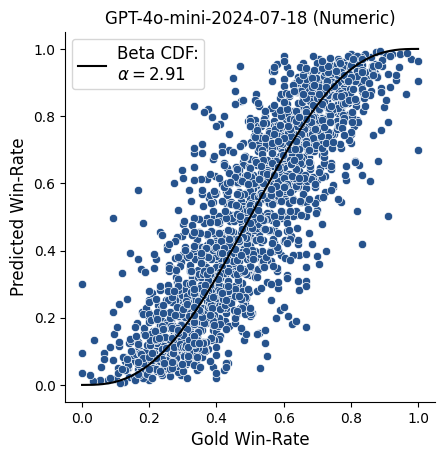}}
\subfloat{\includegraphics[width=.32\columnwidth]{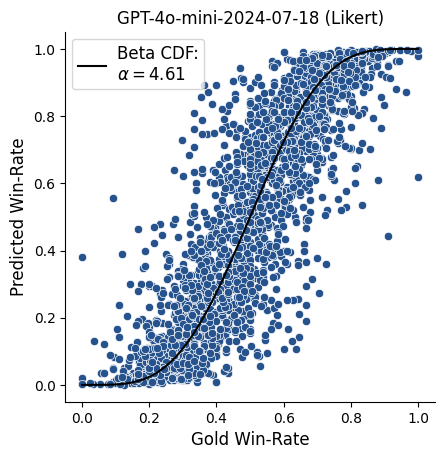}}
\subfloat{\includegraphics[width=.32\columnwidth]{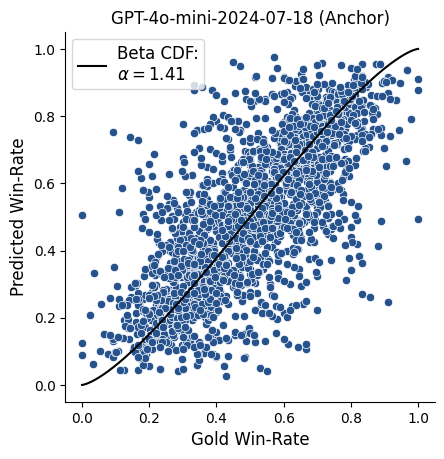}}

\subfloat{\includegraphics[width=.32\columnwidth]{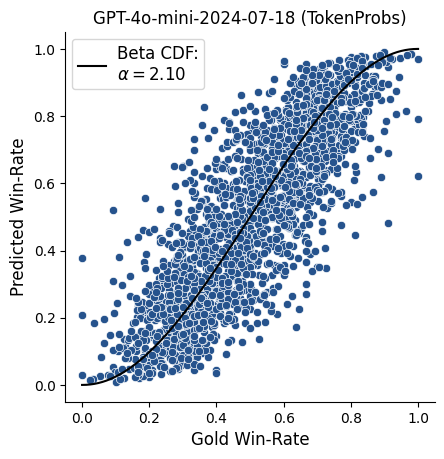}}
\subfloat{\includegraphics[width=.32\columnwidth]{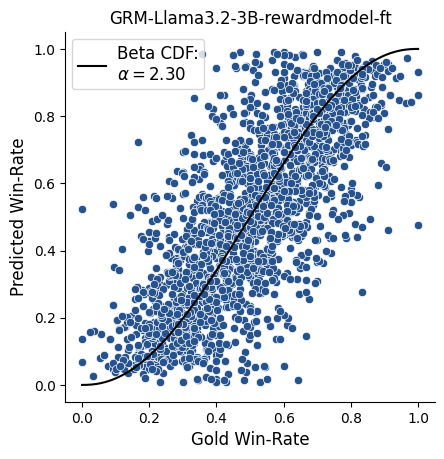}}
\subfloat{\includegraphics[width=.32\columnwidth]{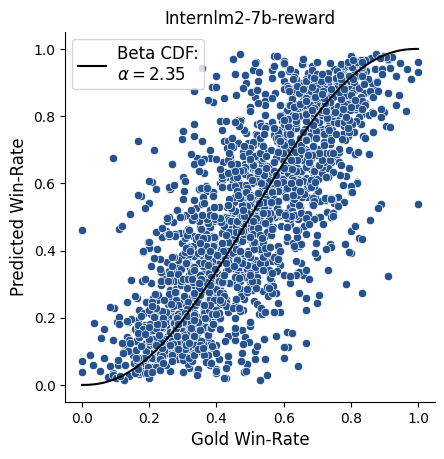}}

\caption{\textbf{Beta distribution fit of pairwise win-rates} \textit{(Part 1/4)}}

\end{figure*}
\begin{figure*}\ContinuedFloat

\subfloat{\includegraphics[width=.32\columnwidth]{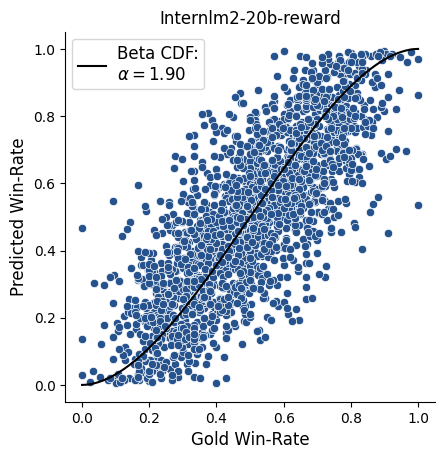}}
\subfloat{\includegraphics[width=.32\columnwidth]{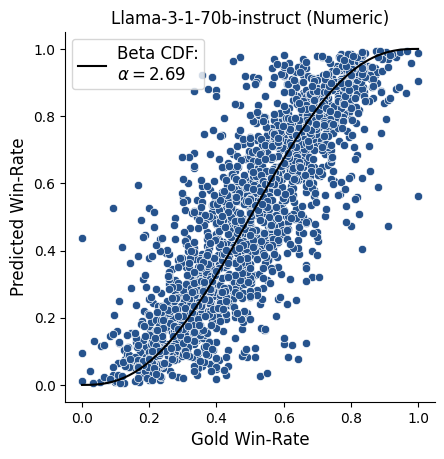}}
\subfloat{\includegraphics[width=.32\columnwidth]{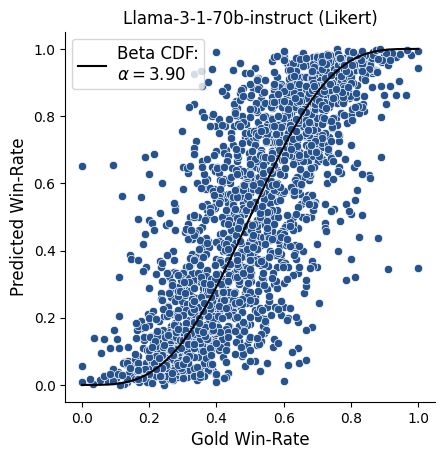}}

\subfloat{\includegraphics[width=.32\columnwidth]{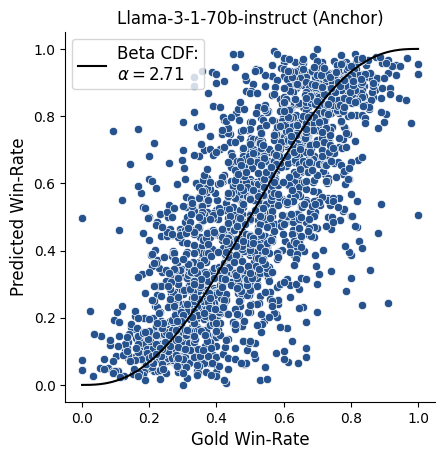}}
\subfloat{\includegraphics[width=.32\columnwidth]{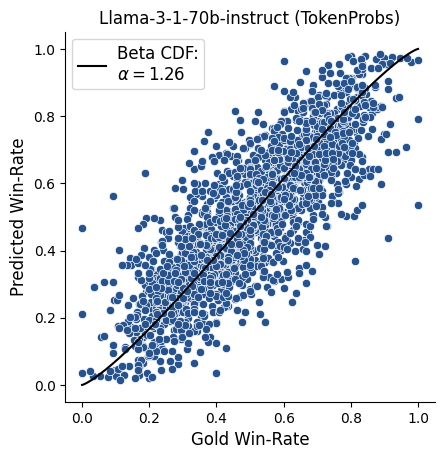}}
\subfloat{\includegraphics[width=.32\columnwidth]{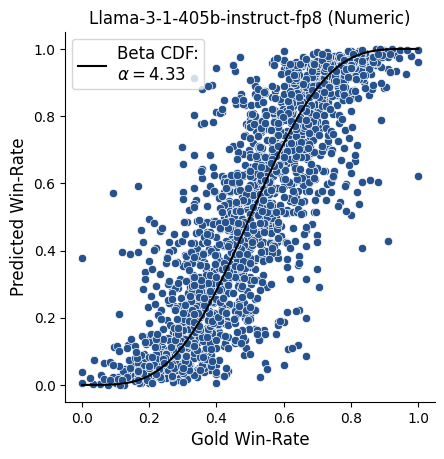}}

\subfloat{\includegraphics[width=.32\columnwidth]{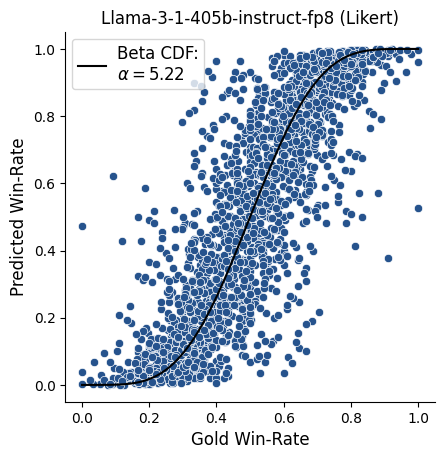}}
\subfloat{\includegraphics[width=.32\columnwidth]{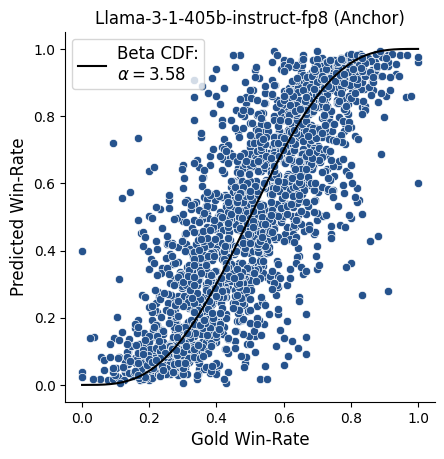}}
\subfloat{\includegraphics[width=.32\columnwidth]{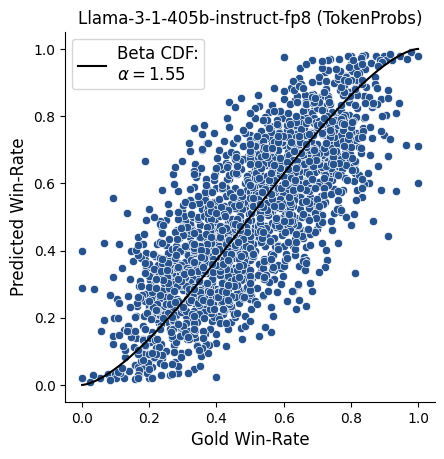}}

\subfloat{\includegraphics[width=.32\columnwidth]{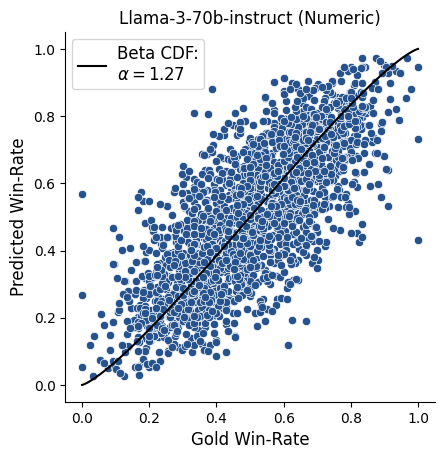}}
\subfloat{\includegraphics[width=.32\columnwidth]{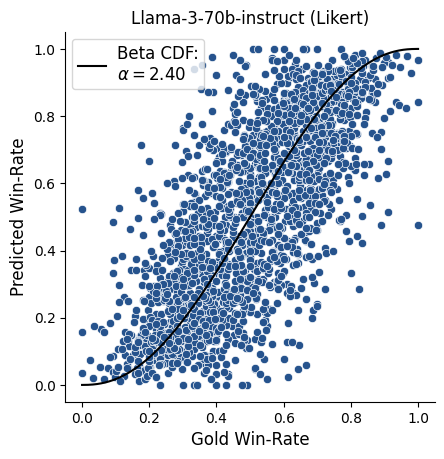}}
\subfloat{\includegraphics[width=.32\columnwidth]{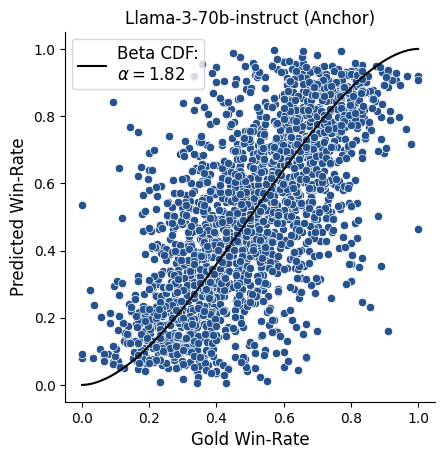}}

\caption{\textbf{Beta distribution fit of pairwise win-rates} \textit{(Part 2/4)}}

\end{figure*}
\begin{figure*}\ContinuedFloat

\subfloat{\includegraphics[width=.32\columnwidth]{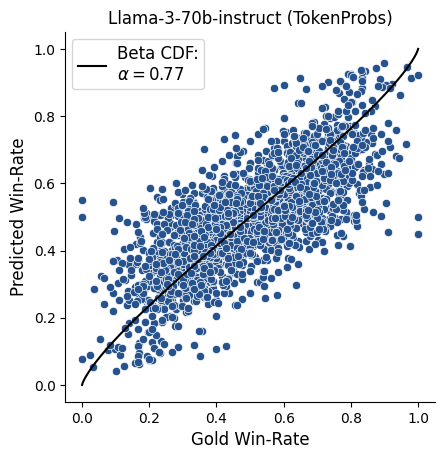}}
\subfloat{\includegraphics[width=.32\columnwidth]{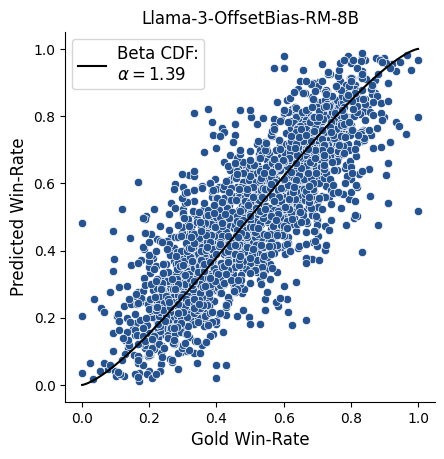}}
\subfloat{\includegraphics[width=.32\columnwidth]{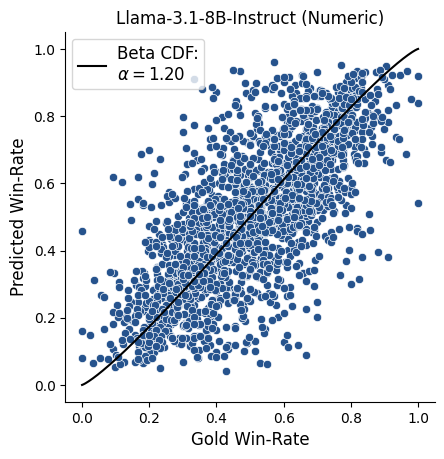}}

\subfloat{\includegraphics[width=.32\columnwidth]{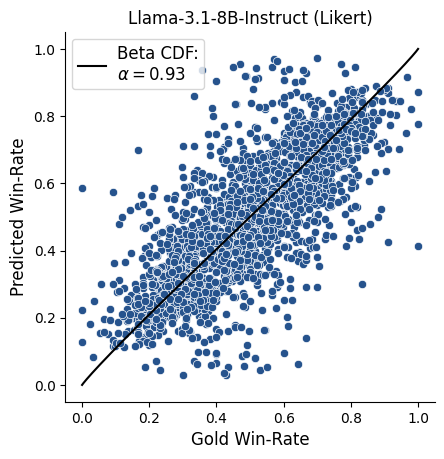}}
\subfloat{\includegraphics[width=.32\columnwidth]{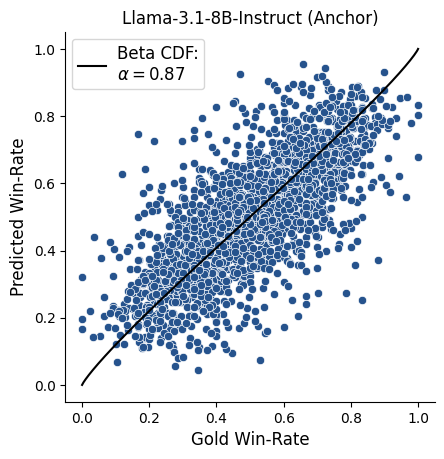}}
\subfloat{\includegraphics[width=.32\columnwidth]{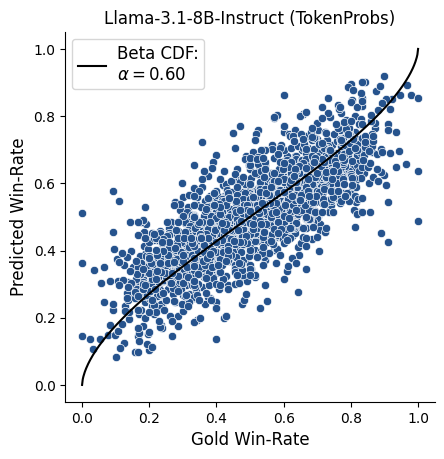}}

\subfloat{\includegraphics[width=.32\columnwidth]{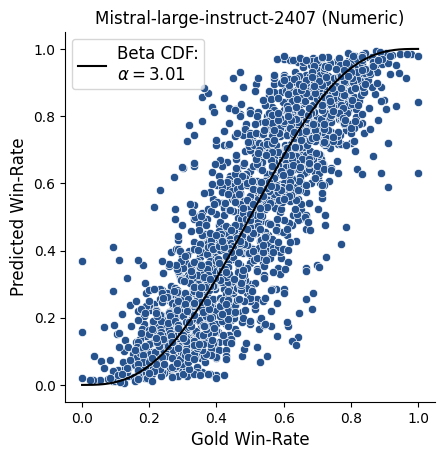}}
\subfloat{\includegraphics[width=.32\columnwidth]{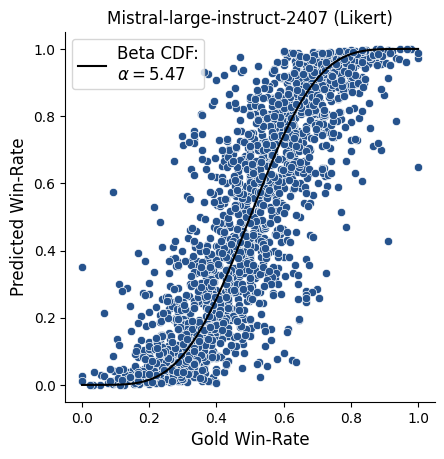}}
\subfloat{\includegraphics[width=.32\columnwidth]{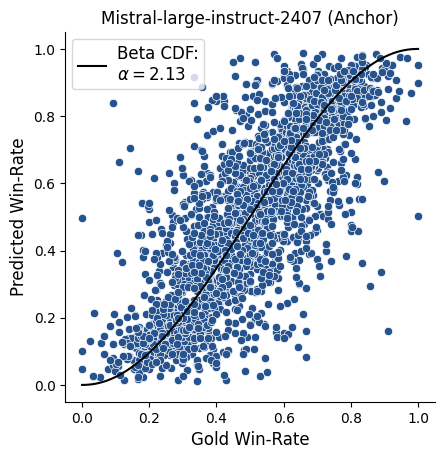}}

\subfloat{\includegraphics[width=.32\columnwidth]{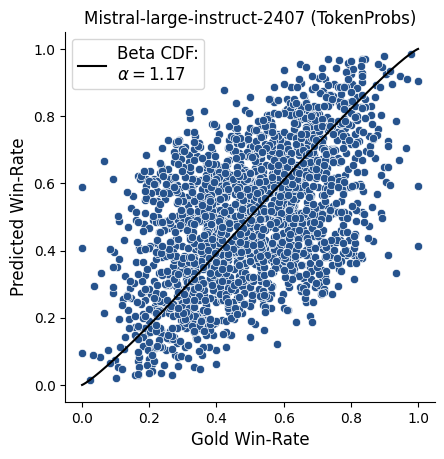}}
\subfloat{\includegraphics[width=.32\columnwidth]{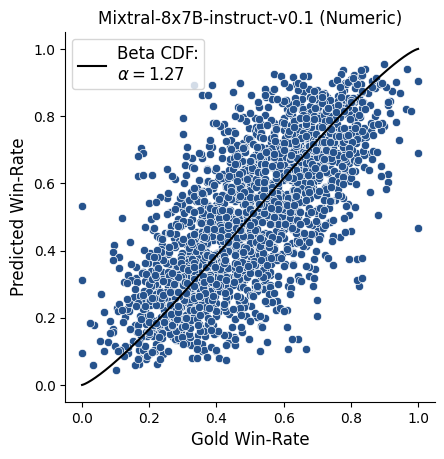}}
\subfloat{\includegraphics[width=.32\columnwidth]{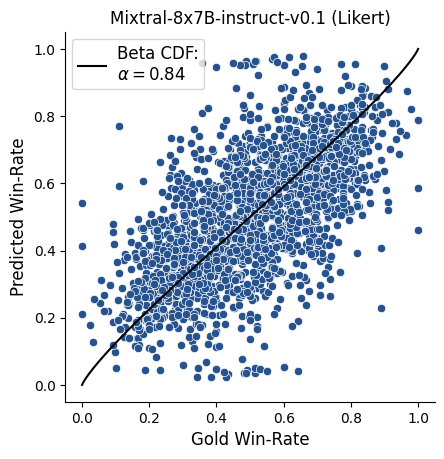}}

\caption{\textbf{Beta distribution fit of pairwise win-rates} \textit{(Part 3/4)}}

\end{figure*}
\begin{figure*}\ContinuedFloat

\subfloat{\includegraphics[width=.32\columnwidth]{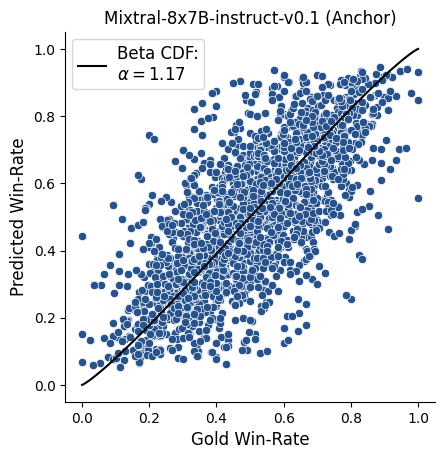}}
\subfloat{\includegraphics[width=.32\columnwidth]{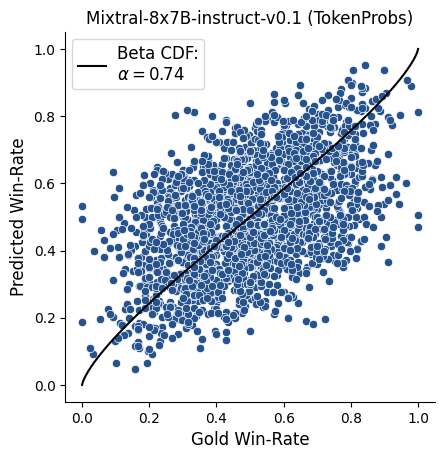}}
\subfloat{\includegraphics[width=.32\columnwidth]{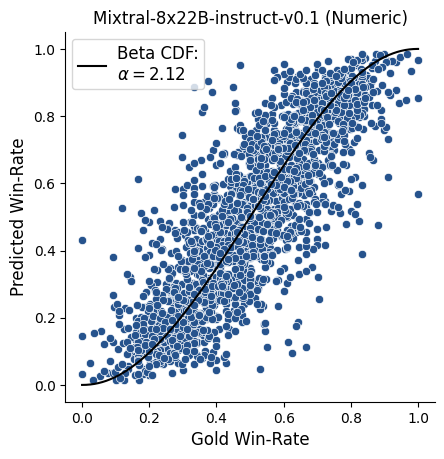}}

\subfloat{\includegraphics[width=.32\columnwidth]{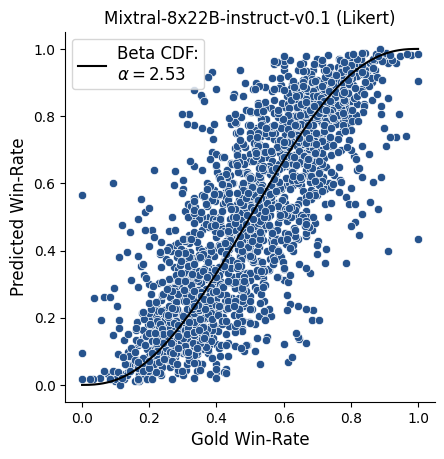}}
\subfloat{\includegraphics[width=.32\columnwidth]{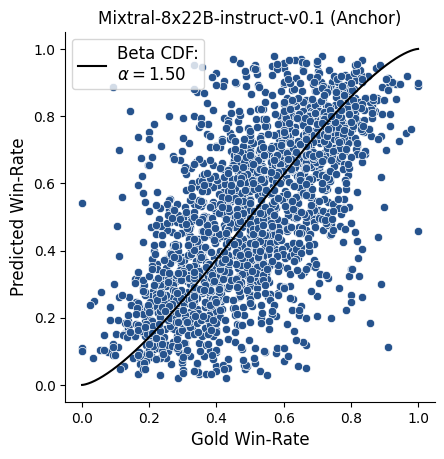}}
\subfloat{\includegraphics[width=.32\columnwidth]{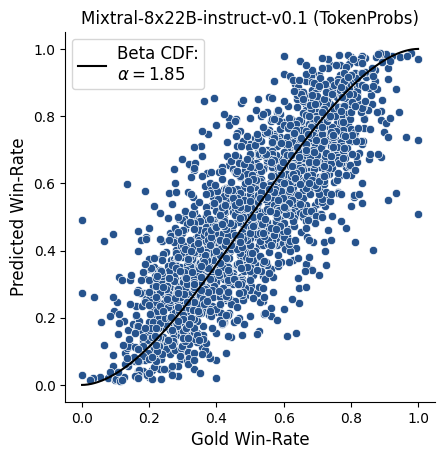}}

\subfloat{\includegraphics[width=.32\columnwidth]{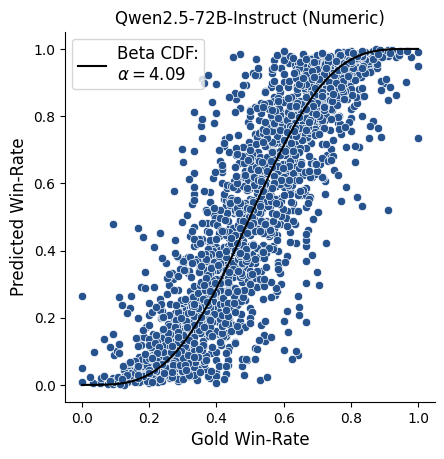}}
\subfloat{\includegraphics[width=.32\columnwidth]{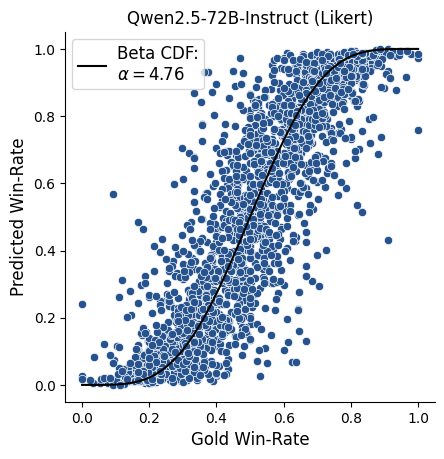}}
\subfloat{\includegraphics[width=.32\columnwidth]{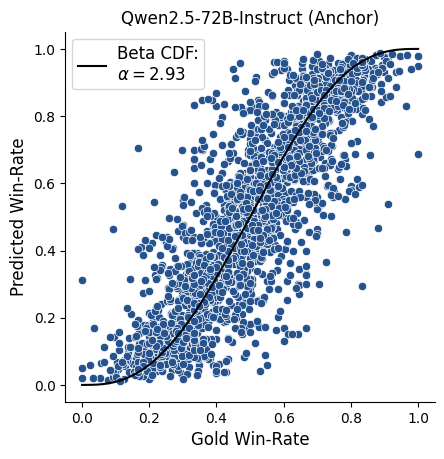}}

\subfloat{\includegraphics[width=.32\columnwidth]{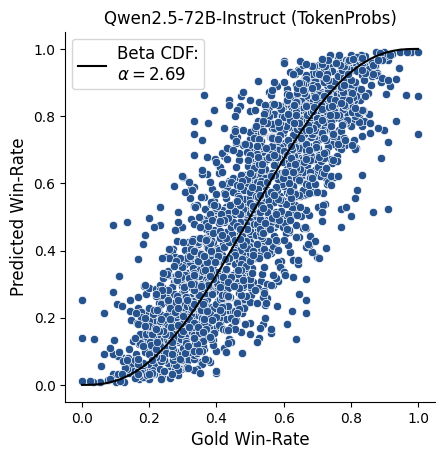}}
\subfloat{\includegraphics[width=.32\columnwidth]{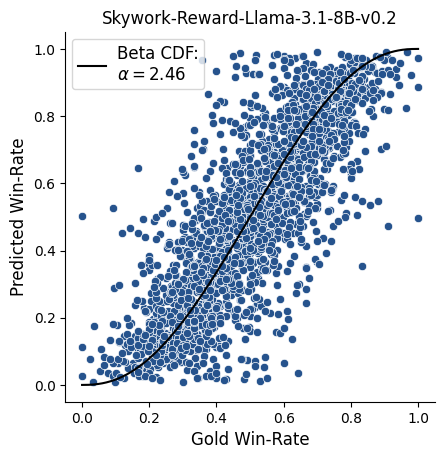}}
\subfloat{\includegraphics[width=.32\columnwidth]{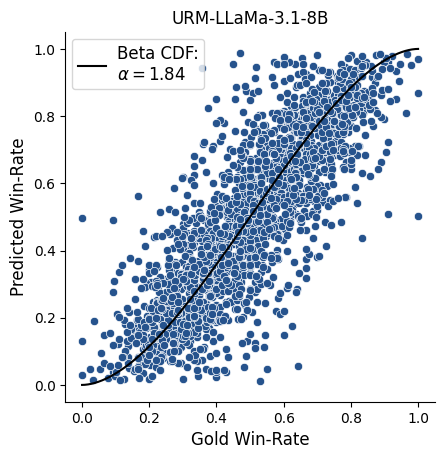}}

\caption{\textbf{Beta distribution fit of pairwise win-rates} \textit{(Part 4/4)}. Each point represents the win-rate between a pair of systems, $WR(s_a, s_b)$; the curve and $\alpha$ value describe a fit to the beta probability distribution. Refer to Appendix~\ref{app:beta} for details.}
\label{fig:all_betas}

\end{figure*}

%% file: self_bias_table.tex
\begin{table*}[ht]
\centering
\begin{tabular}{lll}
\toprule
Judge & Self-bias & Significance $p$-value \\
\midrule
GPT-4o-mini-2024-07-18 (Anchor) & $-0.05$ & -- \\
GPT-4o-mini-2024-07-18 (Likert) & $-0.04$ & -- \\
GPT-4o-mini-2024-07-18 (Numeric) & $+0.03$ & $>0.5$ (N.S.) \\
GPT-4o-mini-2024-07-18 (TokenProbs) & $+0.06$ & $0.13$ (N.S.) \\
Llama-3-1-70b-instruct (Anchor) & $-0.05$ & -- \\
Llama-3-1-70b-instruct (Likert) & $+0.16$ & $7.1e-03$ \\
Llama-3-1-70b-instruct (Numeric) & $-0.00$ & -- \\
Llama-3-1-70b-instruct (TokenProbs) & $-0.03$ & -- \\
Llama-3-70b-instruct (Anchor) & $+0.09$ & $4.7e-04$ \\
Llama-3-70b-instruct (Likert) & $+0.15$ & $8.4e-08$ \\
Llama-3-70b-instruct (Numeric) & $+0.14$ & $1.8e-13$ \\
Llama-3-70b-instruct (TokenProbs) & $-0.01$ & -- \\
Llama-3.1-8B-Instruct (Anchor) & $-0.07$ & -- \\
Llama-3.1-8B-Instruct (Likert) & $-0.04$ & -- \\
Llama-3.1-8B-Instruct (Numeric) & $+0.02$ & $>0.5$ (N.S.) \\
Llama-3.1-8B-Instruct (TokenProbs) & $-0.04$ & -- \\
Mistral-large-instruct-2407 (Anchor) & $-0.07$ & -- \\
Mistral-large-instruct-2407 (Likert) & $+0.02$ & $>0.5$ (N.S.) \\
Mistral-large-instruct-2407 (Numeric) & $+0.06$ & $0.33$ (N.S.) \\
Mistral-large-instruct-2407 (TokenProbs) & $+0.01$ & $>0.5$ (N.S.) \\
\bottomrule
\end{tabular}
\caption{\textbf{Judge self-bias.} The table shows the self-bias values for LLM judge realizations, i.e., the value of the corrected bias ${B'_{s_a}}^p$ (\S\ref{ssec:bias}) where the LLM judge $p$ and system $s_a$ correspond to the same underlying LLM. For positive self-bias values we test the statistical significance using paired t-tests (one-sided, with Bonferroni correction). N.S.: non-significant ($p>0.05$).}
\label{tab:self_bias}
\end{table*}

%% file: distributions_fig.tex
\begin{figure*}[t]
\subfloat{\includegraphics[width=.315\columnwidth]{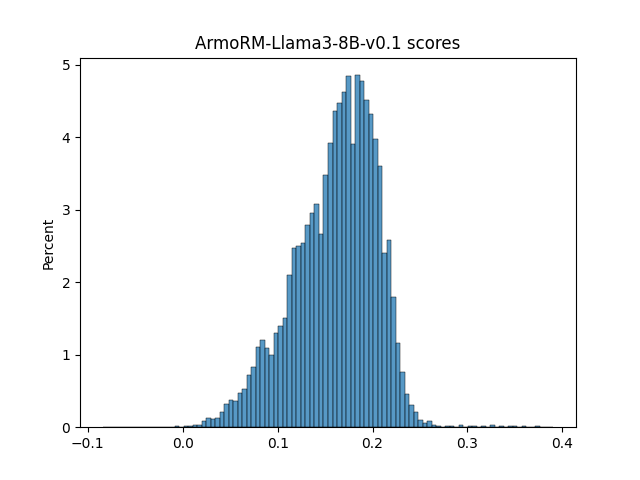}}
\subfloat{\includegraphics[width=.315\columnwidth]{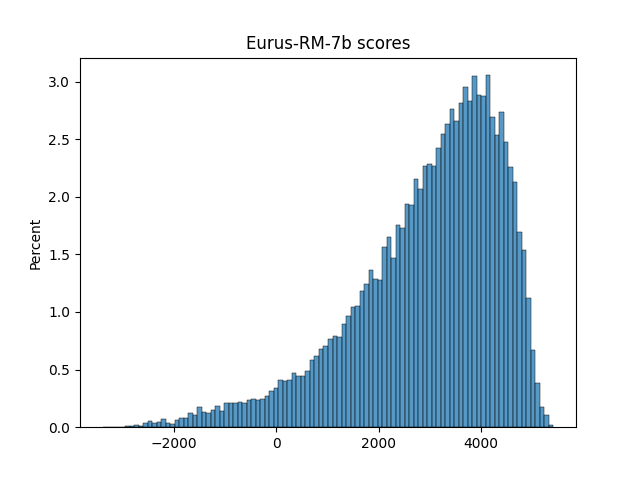}}
\subfloat{\includegraphics[width=.315\columnwidth]{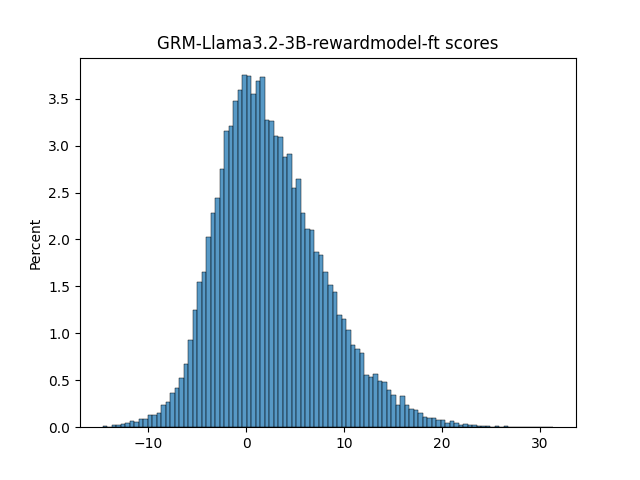}}

\subfloat{\includegraphics[width=.315\columnwidth]{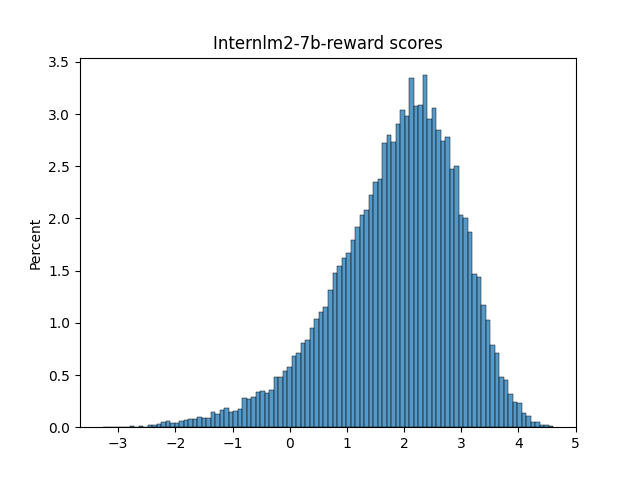}}
\subfloat{\includegraphics[width=.315\columnwidth]{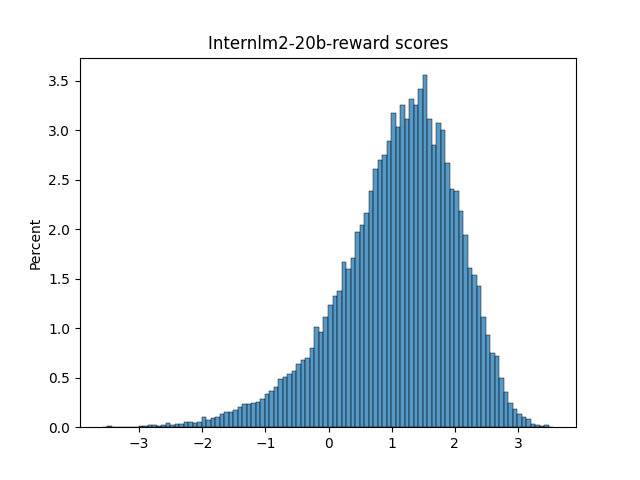}}
\subfloat{\includegraphics[width=.315\columnwidth]{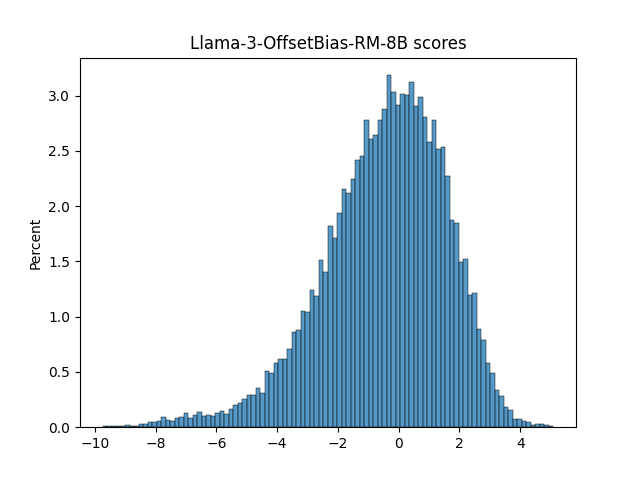}}

\subfloat{\includegraphics[width=.315\columnwidth]{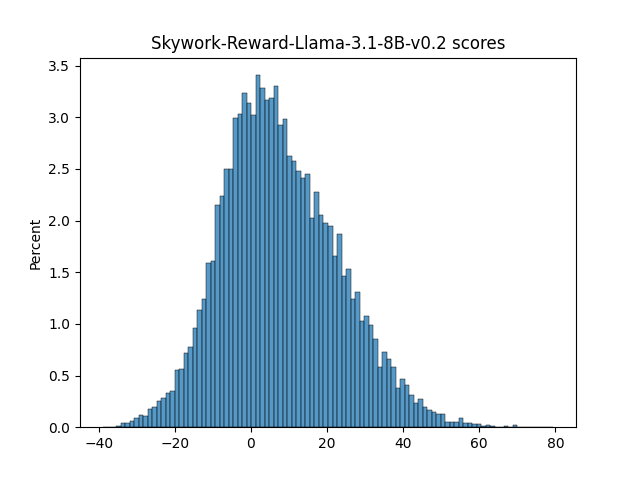}}
\subfloat{\includegraphics[width=.315\columnwidth]{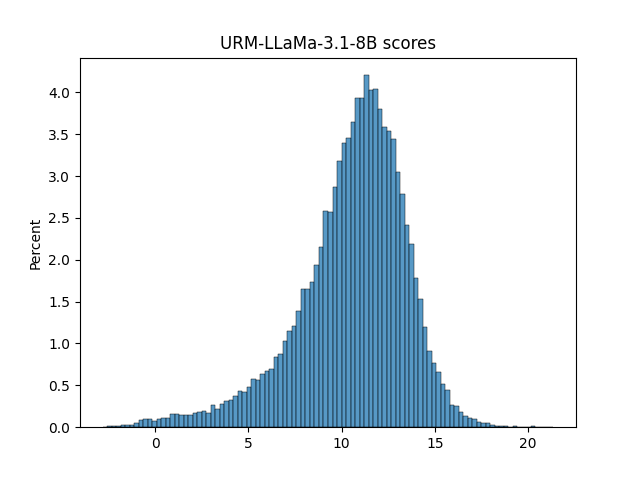}}
\subfloat{\includegraphics[width=.315\columnwidth]{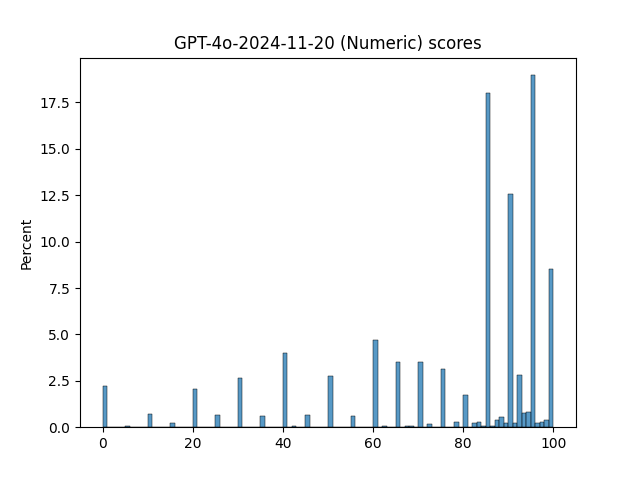}}

\subfloat{\includegraphics[width=.315\columnwidth]{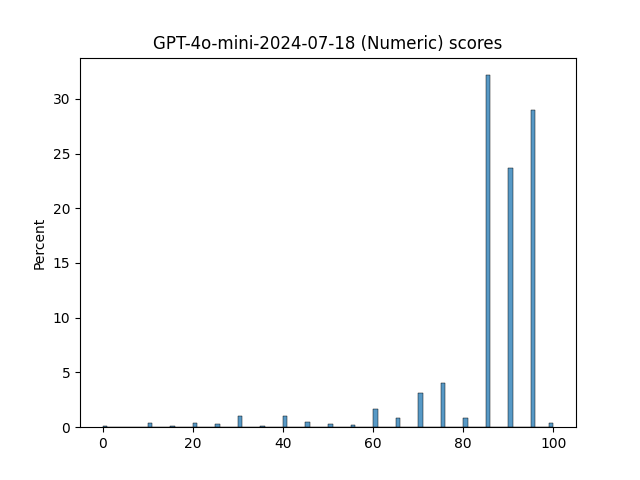}}
\subfloat{\includegraphics[width=.315\columnwidth]{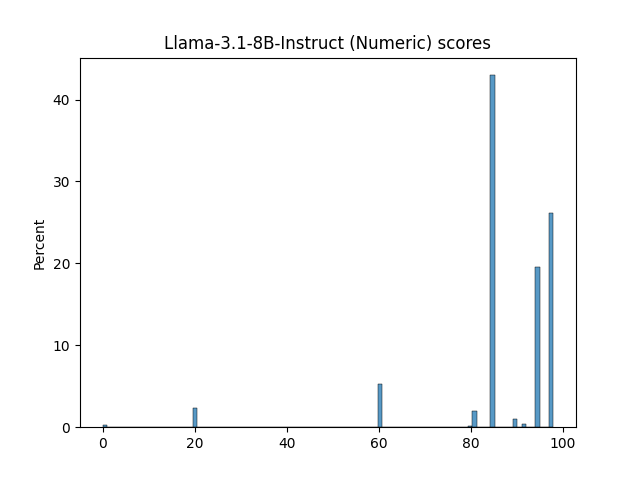}}
\subfloat{\includegraphics[width=.315\columnwidth]{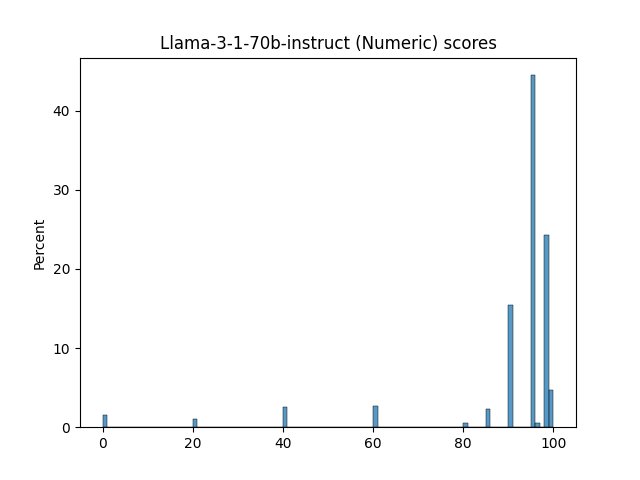}}

\subfloat{\includegraphics[width=.315\columnwidth]{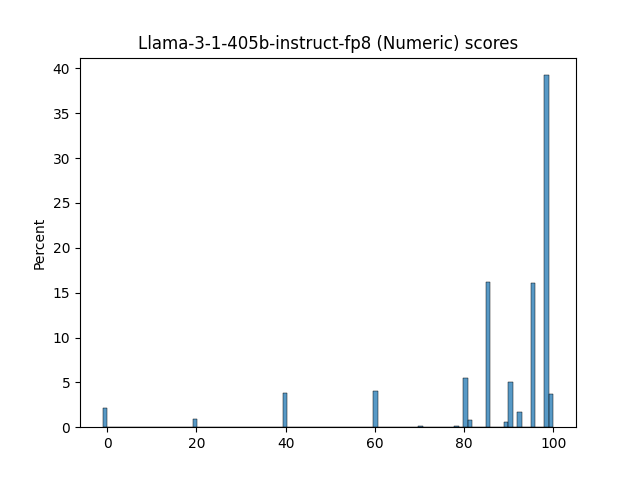}}
\subfloat{\includegraphics[width=.315\columnwidth]{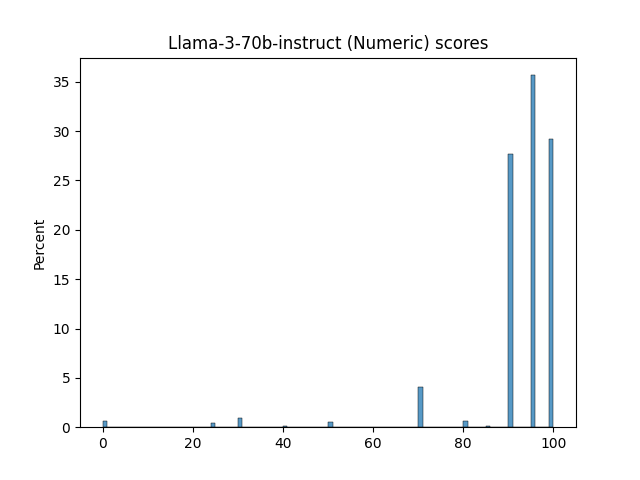}}
\subfloat{\includegraphics[width=.315\columnwidth]{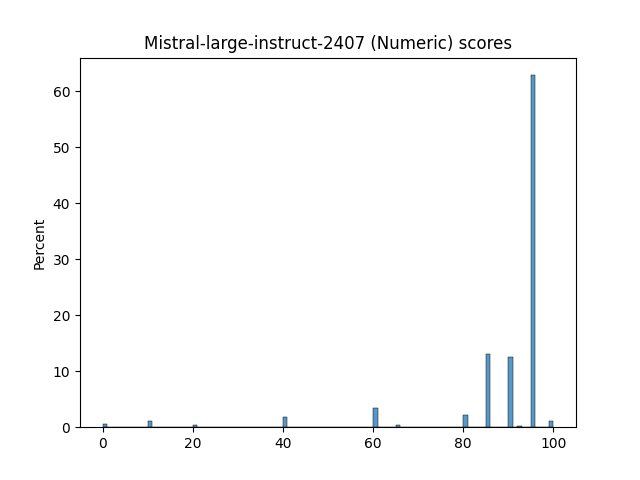}}

\caption{\textbf{Judge score distributions} \textit{(Part 1/3)}}
\end{figure*}
\begin{figure*}\ContinuedFloat

\subfloat{\includegraphics[width=.315\columnwidth]{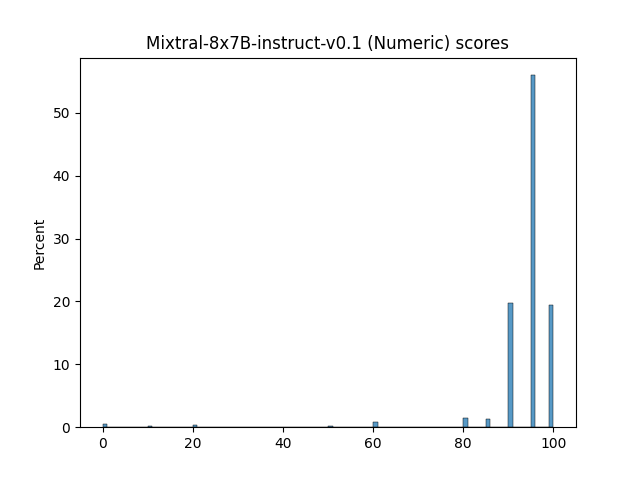}}
\subfloat{\includegraphics[width=.315\columnwidth]{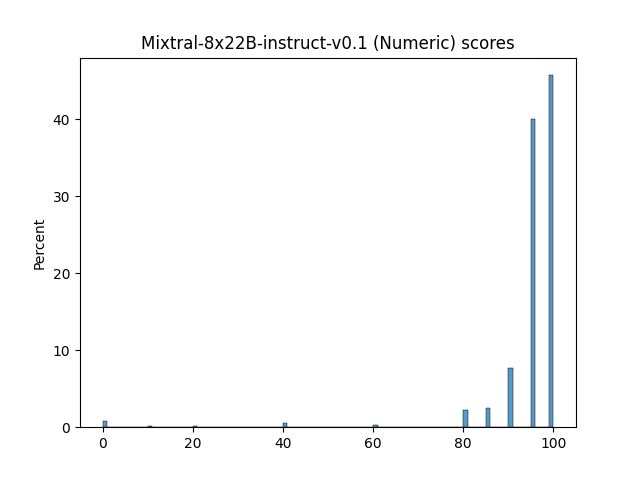}}
\subfloat{\includegraphics[width=.315\columnwidth]{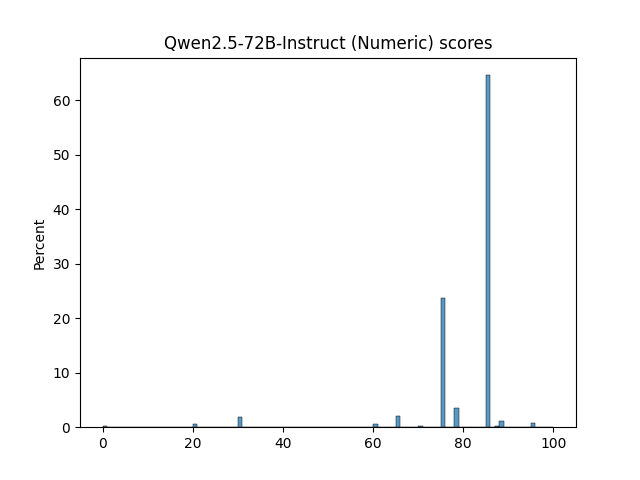}}

\subfloat{\includegraphics[width=.315\columnwidth]{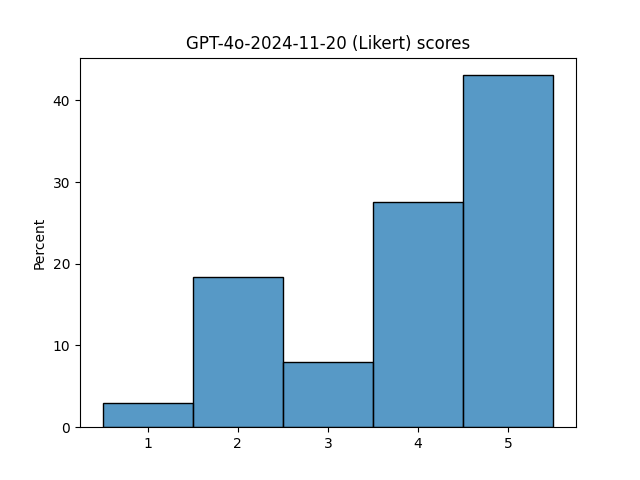}}
\subfloat{\includegraphics[width=.315\columnwidth]{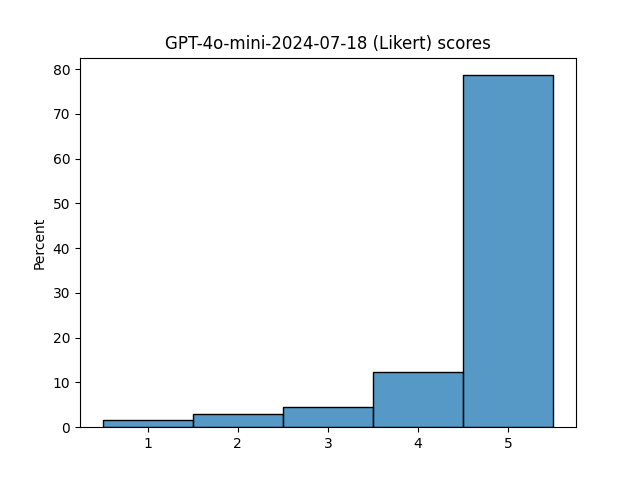}}
\subfloat{\includegraphics[width=.315\columnwidth]{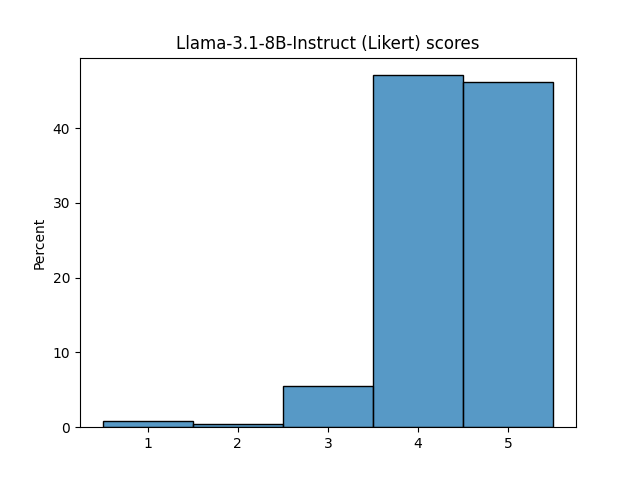}}

\subfloat{\includegraphics[width=.315\columnwidth]{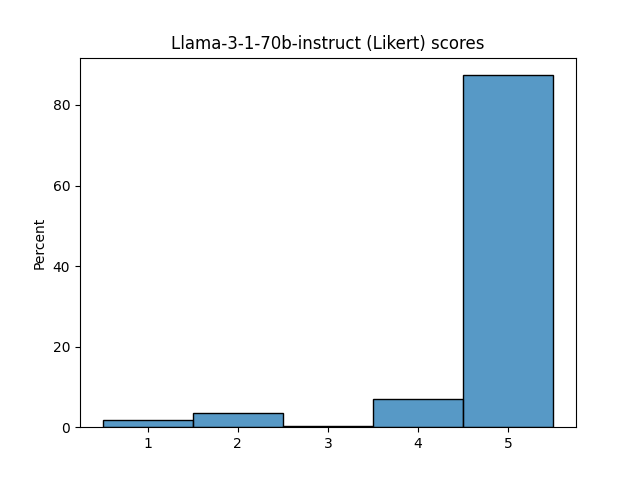}}
\subfloat{\includegraphics[width=.315\columnwidth]{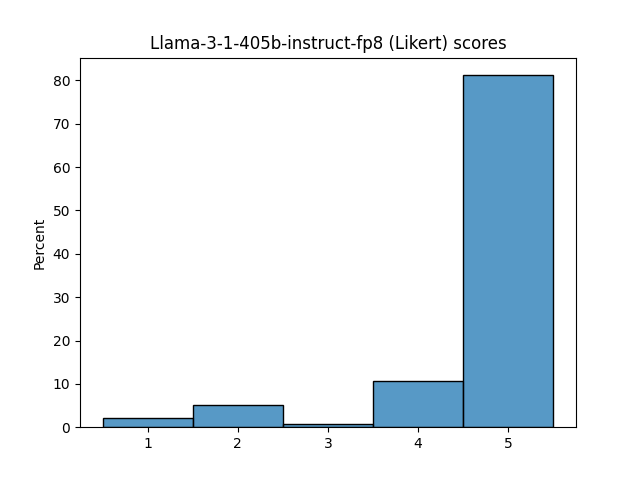}}
\subfloat{\includegraphics[width=.315\columnwidth]{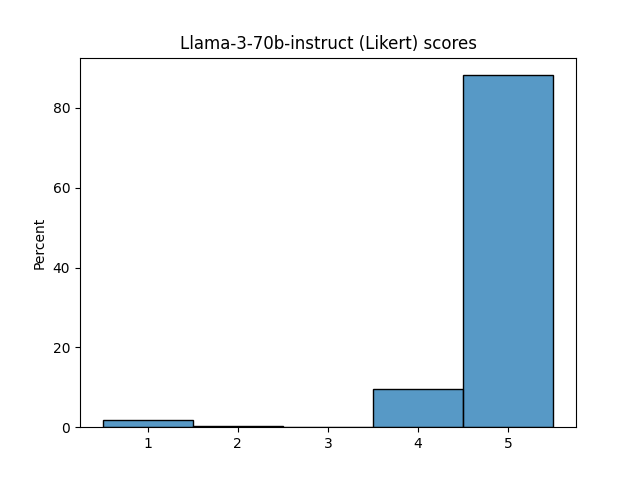}}

\subfloat{\includegraphics[width=.315\columnwidth]{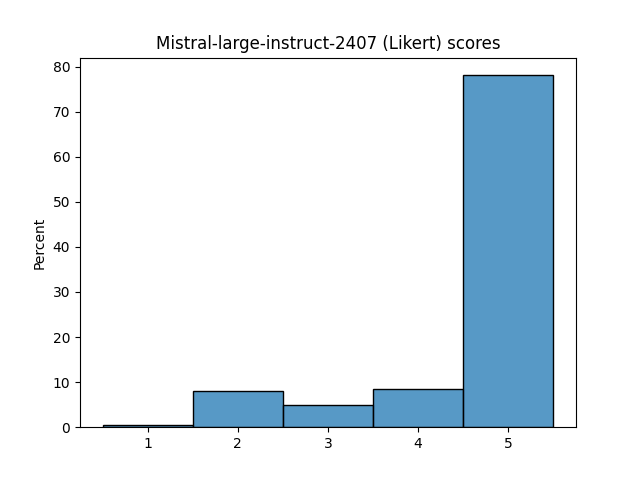}}
\subfloat{\includegraphics[width=.315\columnwidth]{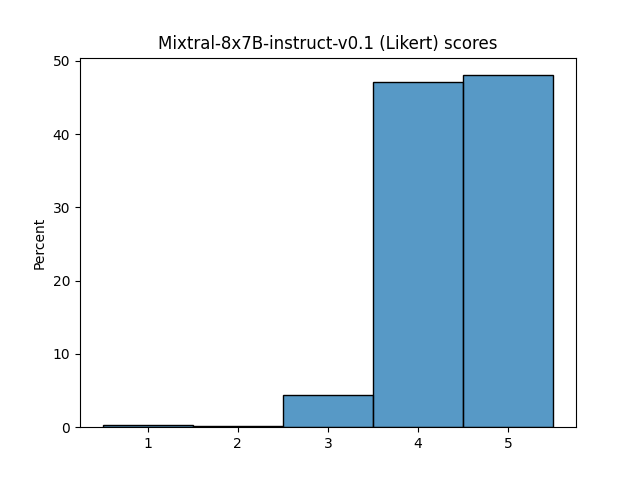}}
\subfloat{\includegraphics[width=.315\columnwidth]{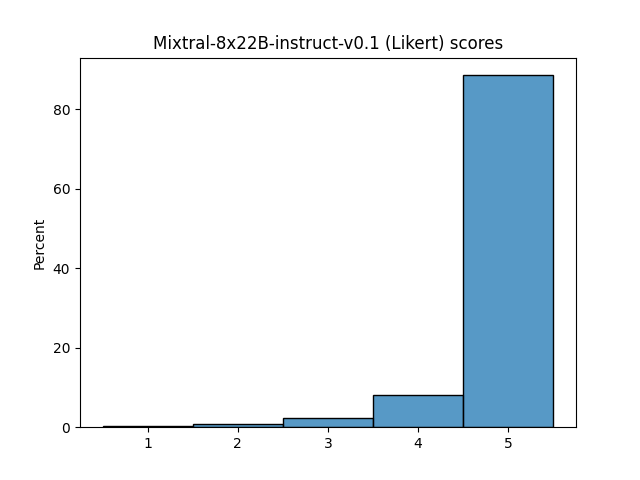}}

\subfloat{\includegraphics[width=.315\columnwidth]{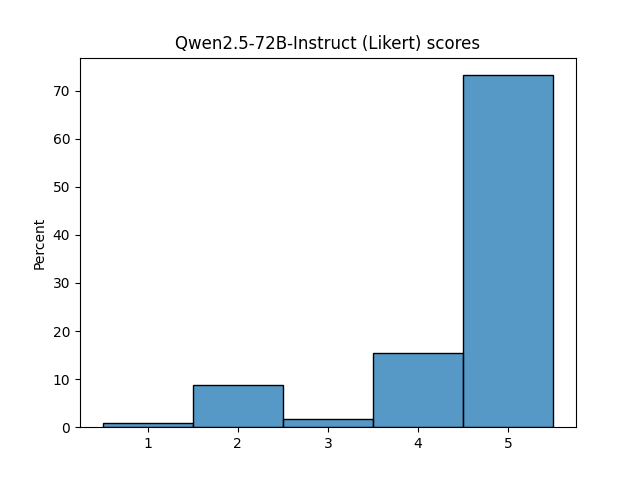}}
\subfloat{\includegraphics[width=.315\columnwidth]{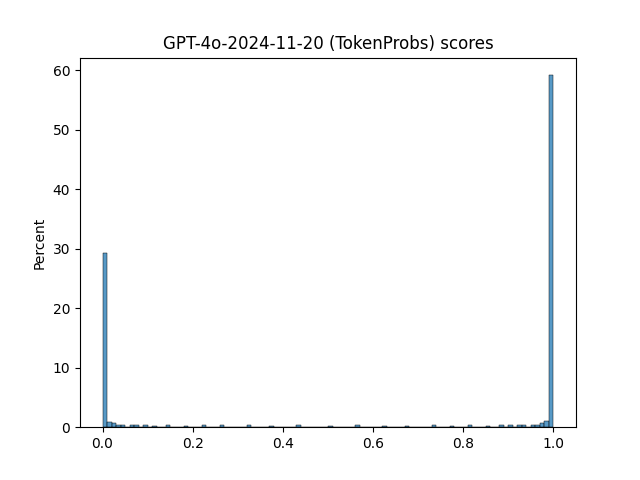}}
\subfloat{\includegraphics[width=.315\columnwidth]{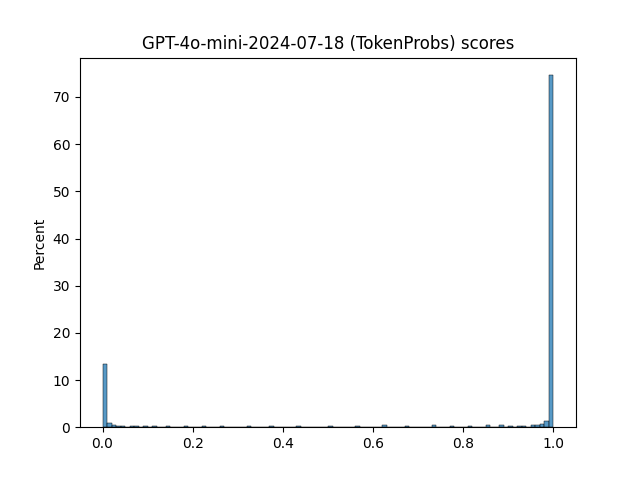}}

\caption{\textbf{Judge score distributions} \textit{(Part 2/3)}}

\end{figure*}
\begin{figure*}\ContinuedFloat
\vspace{-3em}

\subfloat{\includegraphics[width=.315\columnwidth]{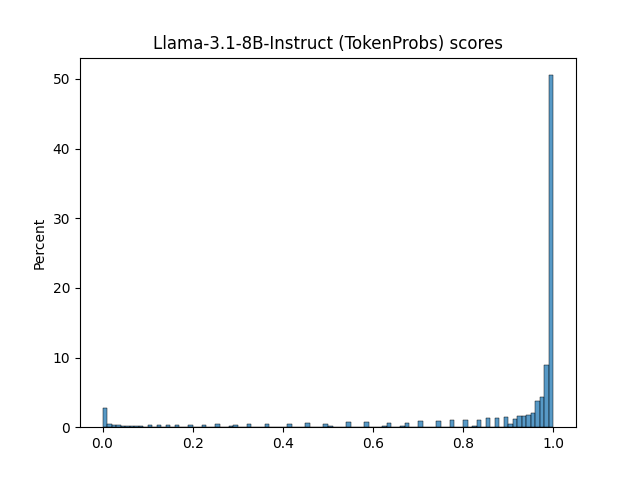}}
\subfloat{\includegraphics[width=.315\columnwidth]{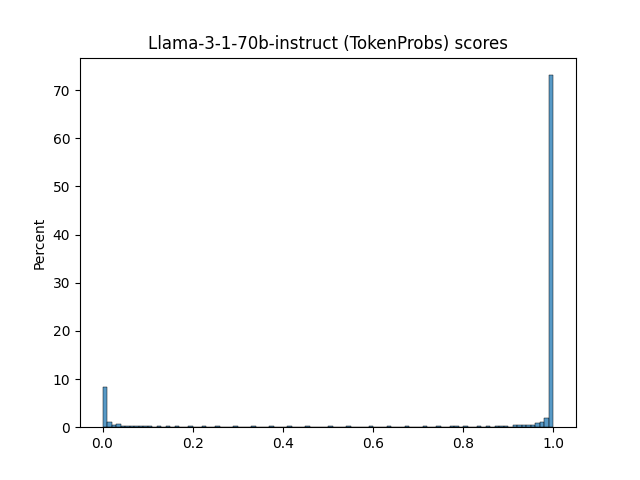}}
\subfloat{\includegraphics[width=.315\columnwidth]{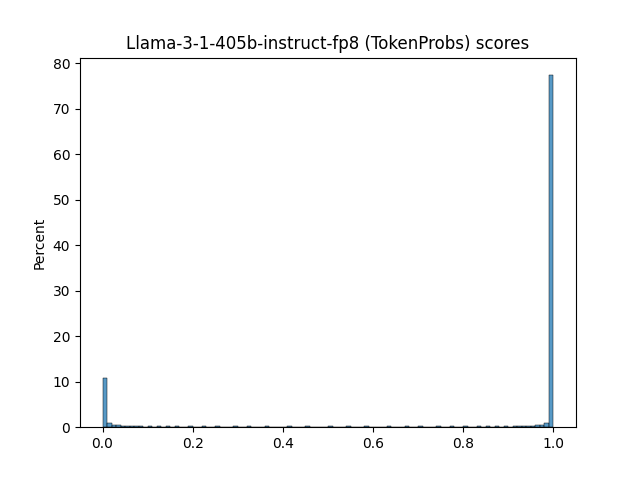}}

\subfloat{\includegraphics[width=.315\columnwidth]{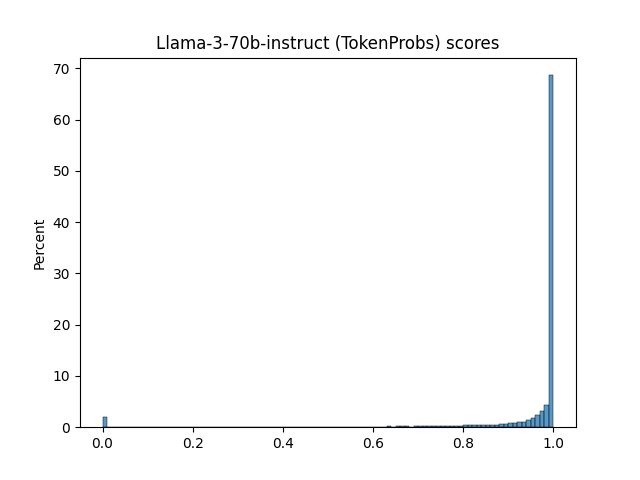}}
\subfloat{\includegraphics[width=.315\columnwidth]{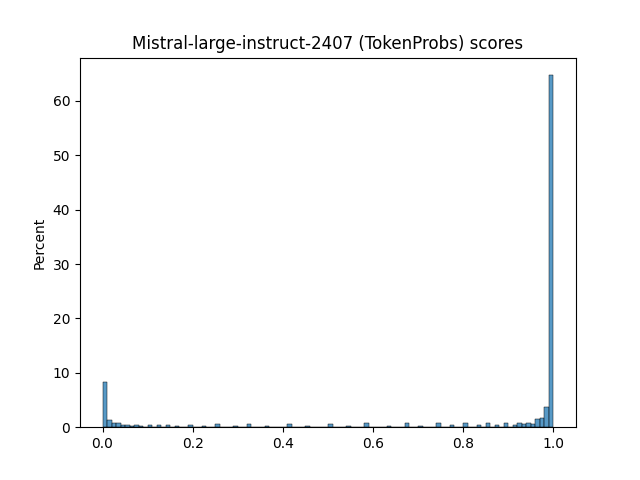}}
\subfloat{\includegraphics[width=.315\columnwidth]{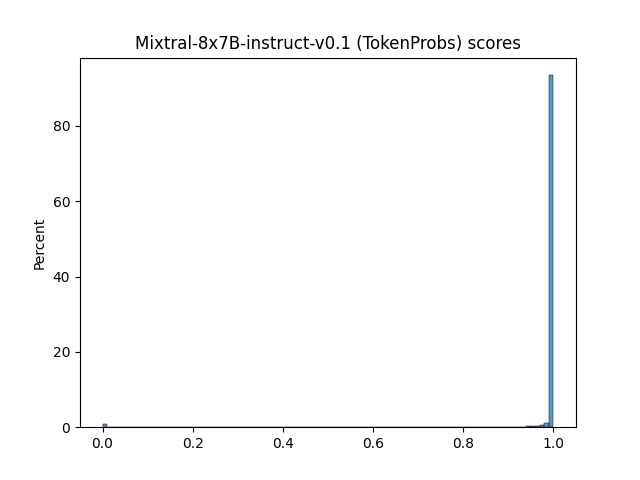}}

\subfloat{\includegraphics[width=.315\columnwidth]{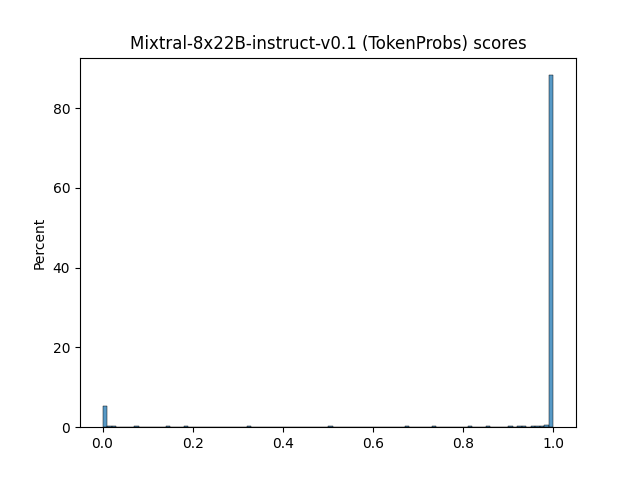}}
\subfloat{\includegraphics[width=.315\columnwidth]{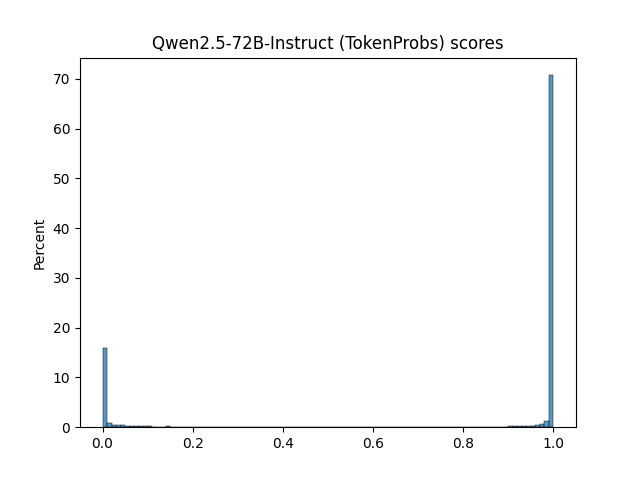}}
\subfloat{\includegraphics[width=.315\columnwidth]{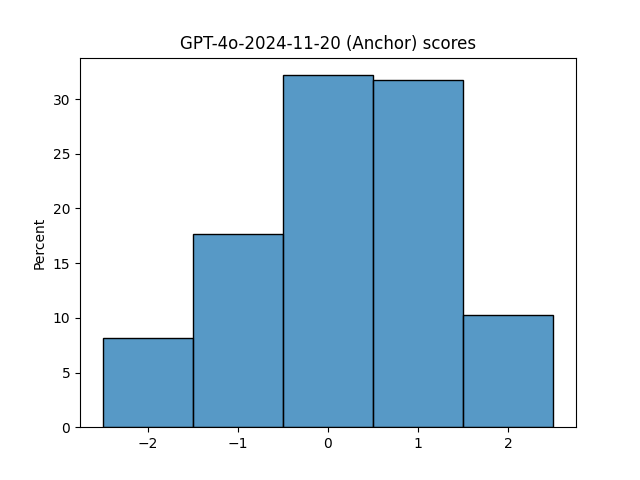}}

\subfloat{\includegraphics[width=.315\columnwidth]{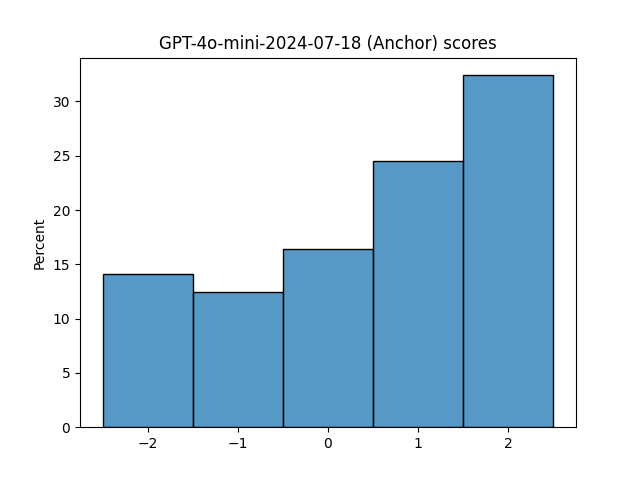}}
\subfloat{\includegraphics[width=.315\columnwidth]{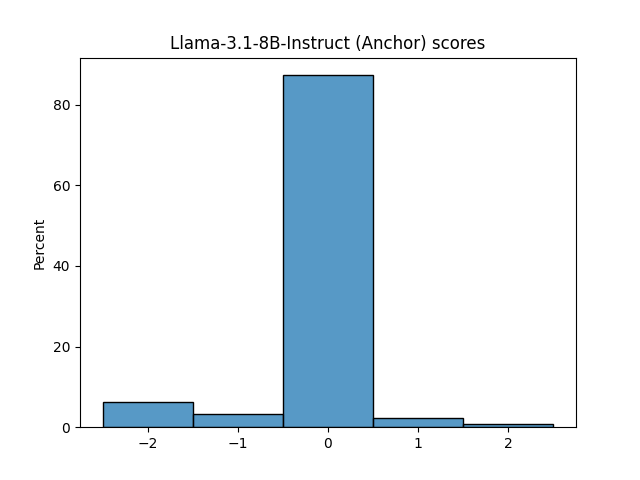}}
\subfloat{\includegraphics[width=.315\columnwidth]{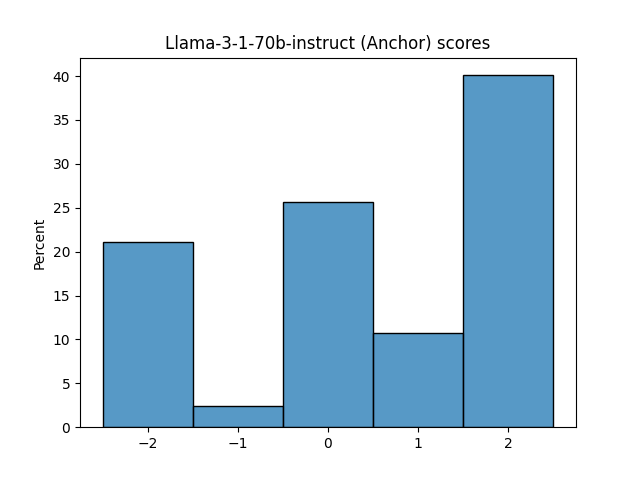}}

\subfloat{\includegraphics[width=.315\columnwidth]{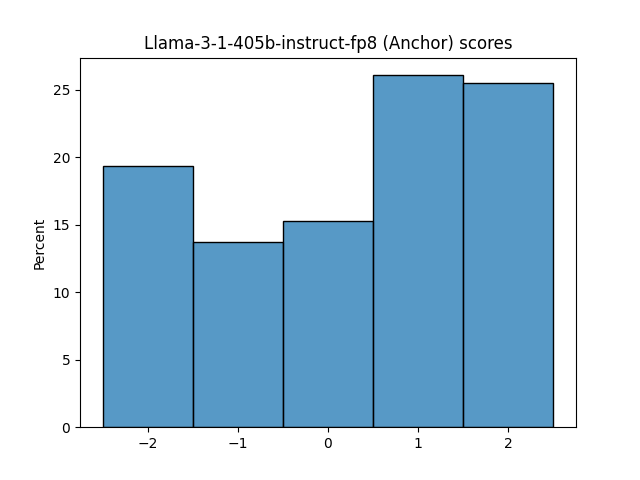}}
\subfloat{\includegraphics[width=.315\columnwidth]{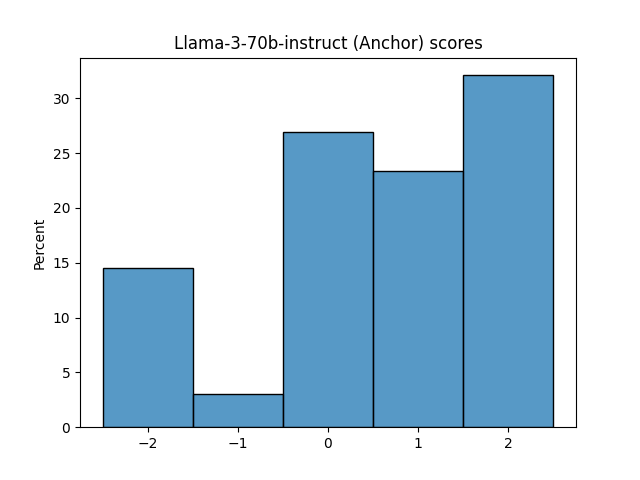}}
\subfloat{\includegraphics[width=.315\columnwidth]{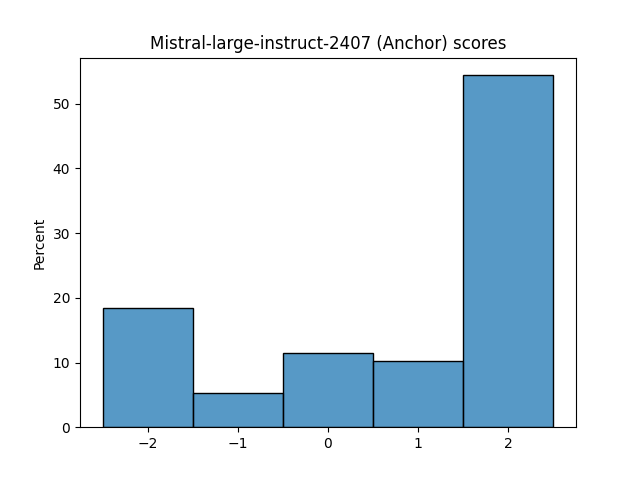}}

\subfloat{\includegraphics[width=.315\columnwidth]{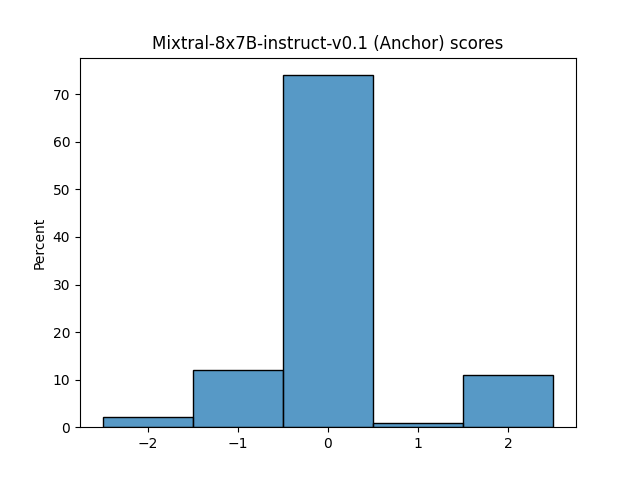}}
\subfloat{\includegraphics[width=.315\columnwidth]{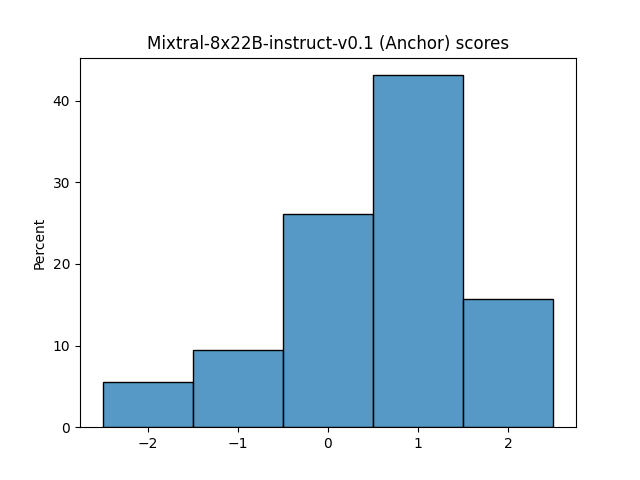}}
\subfloat{\includegraphics[width=.315\columnwidth]{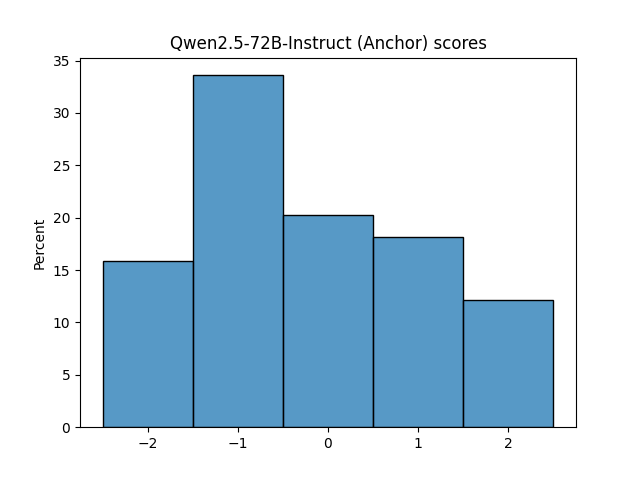}}

\caption{\textbf{Judge score distributions} \textit{(Part 3/3)}.}
\label{fig:score_distributions}

\end{figure*}

%% file: detailed_leaderboard.tex
\begin{table*}[ht]
\small
\centering
\begin{tabular}{llccc}
\toprule
Judge Model & Realization & Agreement & Decisiveness & Bias \\
& & with Gold $\tau \uparrow$ & $\alpha \uparrow$ & \biasstd $\downarrow$
\\\midrule
URM-LLaMa-3.1-8B & Reward & .819 & 1.84 & .085 \\
Qwen2.5-72B-Instruct & Likert & .817 & 4.76 & .079 \\
Qwen2.5-72B-Instruct & Numeric & .814 & 4.09 & .079 \\
Mistral-large-instruct-2407 & Likert & .811 & 5.47 & .086 \\
GPT-4o-2024-11-20 & Anchor & .809 & 3.07 & .085 \\
Mistral-large-instruct-2407 & Numeric & .809 & 3.01 & .082 \\
Llama-3-1-405b-instruct-fp8 & Numeric & .805 & 4.33 & .087 \\
GPT-4o-mini-2024-07-18 & Numeric & .804 & 2.91 & .077 \\
GPT-4o-mini-2024-07-18 & Likert & .798 & 4.61 & .087 \\
Llama-3-1-70b-instruct & Numeric & .798 & 2.69 & .087 \\
Qwen2.5-72B-Instruct & Anchor & .794 & 2.93 & .090 \\
Llama-3-1-405b-instruct-fp8 & Likert & .787 & 5.22 & .097 \\
Skywork-Llama-3.1-8B-v0.2 & Reward & .778 & 2.46 & .100 \\
Qwen2.5-72B-Instruct & TokenProbs & .777 & 2.69 & .082 \\
Mixtral-8x22B-instruct-v0.1 & Numeric & .776 & 2.12 & .089 \\
GPT-4o-2024-11-20 & Numeric & .774 & 2.15 & .077 \\
GPT-4o-2024-11-20 & Likert & .773 & 5.49 & .089 \\
Llama-3-1-70b-instruct & TokenProbs & .765 & 1.26 & .070 \\
Llama-3-OffsetBias-RM-8B & Reward & .765 & 1.39 & .076 \\
ArmoRM-Llama3-8B-v0.1 & Reward & .763 & 1.84 & .092 \\
GPT-4o-mini-2024-07-18 & TokenProbs & .752 & 2.10 & .084 \\
Llama-3-70b-instruct & Numeric & .749 & 1.27 & .084 \\
Llama-3.1-8B-Instruct & TokenProbs & .741 & .598 & .061 \\
Mixtral-8x22B-instruct-v0.1 & Likert & .738 & 2.53 & .108 \\
Llama-3-1-405b-instruct-fp8 & Anchor & .730 & 3.58 & .112 \\
Mistral-large-instruct-2407 & Anchor & .725 & 2.13 & .111 \\
Llama-3.1-8B-Instruct & Likert & .723 & .935 & .090 \\
Llama-3-1-70b-instruct & Likert & .722 & 3.90 & .120 \\
Internlm2-20b-reward & Reward & .717 & 1.90 & .098 \\
Internlm2-7b-reward & Reward & .712 & 2.35 & .113 \\
GRM-Llama3.2-3B & Reward & .711 & 2.30 & .114 \\
Mixtral-8x22B-instruct-v0.1 & TokenProbs & .702 & 1.85 & .088 \\
GPT-4o-2024-11-20 & TokenProbs & .700 & 2.22 & .093 \\
Llama-3-70b-instruct & Likert & .698 & 2.40 & .122 \\
Llama-3-1-70b-instruct & Anchor & .688 & 2.71 & .126 \\
Llama-3.1-8B-Instruct & Anchor & .677 & .868 & .085 \\
Llama-3-1-405b-instruct-fp8 & TokenProbs & .672 & 1.55 & .092 \\
Llama-3.1-8B-Instruct & Numeric & .668 & 1.20 & .104 \\
Llama-3-70b-instruct & TokenProbs & .663 & .775 & .071 \\
GPT-4o-mini-2024-07-18 & Anchor & .659 & 1.41 & .111 \\
Mixtral-8x7B-instruct-v0.1 & Numeric & .656 & 1.27 & .102 \\
Mixtral-8x7B-instruct-v0.1 & Anchor & .655 & 1.17 & .102 \\
Mixtral-8x22B-instruct-v0.1 & Anchor & .641 & 1.50 & .140 \\
Llama-3-70b-instruct & Anchor & .633 & 1.82 & .132 \\
Eurus-RM-7b & Reward & .628 & 2.49 & .138 \\
Mixtral-8x7B-instruct-v0.1 & Likert & .590 & .838 & .110 \\
Mixtral-8x7B-instruct-v0.1 & TokenProbs & .427 & .739 & .107 \\
Mistral-large-instruct-2407 & TokenProbs & .369 & 1.17 & .123 \\
\bottomrule
\end{tabular}
\caption{\textbf{Judge characteristics}. The table presents three measures for each judge realization: an overall ranking quality $\tau$ \textit{(\S\ref{sec:benchmark}, Kendall's Tau correlation with the Chatbot Arena gold ranking)}, a decisiveness score $\alpha$ \textit{(\S\ref{ssec:exaggeration}, App.~\ref{app:beta})}, and its propensity for system-specific biases \biasstd{} \textit{(\S\ref{ssec:bias})}. Correlations $\tau$ shown are for the BT aggregation method; $\alpha$ and \biasstd{} are calculated on the judge scores before aggregation. $\downarrow$: Lower is better.}
\label{tab:leaderboard_detailed}
\end{table*}

%% file: acl_latex.bbl
\begin{thebibliography}{49}
\providecommand{\natexlab}[1]{#1}

\bibitem[{Ashury~Tahan et~al.(2024)Ashury~Tahan, Gera, Sznajder, Choshen, Ein-Dor, and Shnarch}]{ashury-tahan-etal-2024-label}
Shir Ashury~Tahan, Ariel Gera, Benjamin Sznajder, Leshem Choshen, Liat Ein-Dor, and Eyal Shnarch. 2024.
\newblock \href {https://doi.org/10.18653/v1/2024.acl-long.456} {Label-efficient model selection for text generation}.
\newblock In \emph{Proceedings of the 62nd Annual Meeting of the Association for Computational Linguistics (Volume 1: Long Papers)}, pages 8384--8402, Bangkok, Thailand. Association for Computational Linguistics.

\bibitem[{Bavaresco et~al.(2024)Bavaresco, Bernardi, Bertolazzi, Elliott, Fern{\'a}ndez, Gatt, Ghaleb, Giulianelli, Hanna, Koller et~al.}]{bavaresco2024llms}
Anna Bavaresco, Raffaella Bernardi, Leonardo Bertolazzi, Desmond Elliott, Raquel Fern{\'a}ndez, Albert Gatt, Esam Ghaleb, Mario Giulianelli, Michael Hanna, Alexander Koller, et~al. 2024.
\newblock \href {https://arxiv.org/abs/2406.18403} {{LLM}s instead of human judges? a large scale empirical study across 20 {NLP} evaluation tasks}.
\newblock \emph{arXiv:2406.18403}.

\bibitem[{Bradley and Terry(1952)}]{bradley1952rank}
Ralph~Allan Bradley and Milton~E Terry. 1952.
\newblock Rank analysis of incomplete block designs: I. the method of paired comparisons.
\newblock \emph{Biometrika}, 39(3/4):324--345.

\bibitem[{Cai et~al.(2024)Cai, Cao, Chen, Chen, Chen, Chen, Chen, Chen, Chen, Chu et~al.}]{cai2024internlm2}
Zheng Cai, Maosong Cao, Haojiong Chen, Kai Chen, Keyu Chen, Xin Chen, Xun Chen, Zehui Chen, Zhi Chen, Pei Chu, et~al. 2024.
\newblock \href {https://arxiv.org/abs/2403.17297} {Intern{LM}2 technical report}.
\newblock \emph{arXiv:2403.17297}.

\bibitem[{Chen et~al.(2024)Chen, Zhu, Chen, Soselia, Zhou, Goldstein, Huang, Shoeybi, and Catanzaro}]{chen2024odin}
Lichang Chen, Chen Zhu, Jiuhai Chen, Davit Soselia, Tianyi Zhou, Tom Goldstein, Heng Huang, Mohammad Shoeybi, and Bryan Catanzaro. 2024.
\newblock \href {https://openreview.net/forum?id=zcIV8OQFVF} {{ODIN}: Disentangled reward mitigates hacking in {RLHF}}.
\newblock In \emph{Forty-first International Conference on Machine Learning}.

\bibitem[{Chiang et~al.(2024)Chiang, Zheng, Sheng, Angelopoulos, Li, Li, Zhu, Zhang, Jordan, Gonzalez et~al.}]{chiang2024chatbot}
Wei-Lin Chiang, Lianmin Zheng, Ying Sheng, Anastasios~Nikolas Angelopoulos, Tianle Li, Dacheng Li, Banghua Zhu, Hao Zhang, Michael Jordan, Joseph~E Gonzalez, et~al. 2024.
\newblock \href {https://openreview.net/forum?id=3MW8GKNyzI} {Chatbot {A}rena: An open platform for evaluating {LLM}s by human preference}.
\newblock In \emph{Forty-first International Conference on Machine Learning}.

\bibitem[{Conitzer et~al.(2024)Conitzer, Freedman, Heitzig, Holliday, Jacobs, Lambert, Moss{\'e}, Pacuit, Russell, Schoelkopf et~al.}]{conitzer2024social}
Vincent Conitzer, Rachel Freedman, Jobst Heitzig, Wesley~H Holliday, Bob~M Jacobs, Nathan Lambert, Milan Moss{\'e}, Eric Pacuit, Stuart Russell, Hailey Schoelkopf, et~al. 2024.
\newblock \href {https://arxiv.org/abs/2404.10271} {Social choice should guide ai alignment in dealing with diverse human feedback}.
\newblock \emph{arXiv:2404.10271}.

\bibitem[{Deutsch et~al.(2022)Deutsch, Dror, and Roth}]{deutsch-etal-2022-examining}
Daniel Deutsch, Rotem Dror, and Dan Roth. 2022.
\newblock \href {https://doi.org/10.18653/v1/2022.naacl-main.442} {Re-examining system-level correlations of automatic summarization evaluation metrics}.
\newblock In \emph{Proceedings of the 2022 Conference of the North American Chapter of the Association for Computational Linguistics: Human Language Technologies}, pages 6038--6052, Seattle, United States. Association for Computational Linguistics.

\bibitem[{Dorner et~al.(2024)Dorner, Nastl, and Hardt}]{dorner2024limits}
Florian~E Dorner, Vivian~Y Nastl, and Moritz Hardt. 2024.
\newblock \href {https://arxiv.org/abs/2410.13341} {Limits to scalable evaluation at the frontier: {LLM} as judge won't beat twice the data}.
\newblock \emph{arXiv:2410.13341}.

\bibitem[{Dubey et~al.(2024)Dubey, Jauhri, Pandey, Kadian, Al-Dahle, Letman, Mathur, Schelten, Yang, Fan et~al.}]{dubey2024llama}
Abhimanyu Dubey, Abhinav Jauhri, Abhinav Pandey, Abhishek Kadian, Ahmad Al-Dahle, Aiesha Letman, Akhil Mathur, Alan Schelten, Amy Yang, Angela Fan, et~al. 2024.
\newblock \href {https://arxiv.org/abs/2407.21783} {The {L}lama 3 herd of models}.
\newblock \emph{arXiv:2407.21783}.

\bibitem[{Dubois et~al.(2024)Dubois, Galambosi, Liang, and Hashimoto}]{dubois2024length}
Yann Dubois, Bal{\'a}zs Galambosi, Percy Liang, and Tatsunori~B Hashimoto. 2024.
\newblock \href {https://arxiv.org/abs/2404.04475} {Length-controlled {A}lpaca{E}val: A simple way to debias automatic evaluators}.
\newblock \emph{arXiv:2404.04475}.

\bibitem[{Feuer et~al.(2024)Feuer, Goldblum, Datta, Nambiar, Besaleli, Dooley, Cembalest, and Dickerson}]{feuer2024style}
Benjamin Feuer, Micah Goldblum, Teresa Datta, Sanjana Nambiar, Raz Besaleli, Samuel Dooley, Max Cembalest, and John~P Dickerson. 2024.
\newblock \href {https://arxiv.org/abs/2409.15268} {Style outweighs substance: Failure modes of {LLM} judges in alignment benchmarking}.
\newblock \emph{arXiv:2409.15268}.

\bibitem[{Gureja et~al.(2024)Gureja, Miranda, Islam, Maheshwary, Sharma, Winata, Lambert, Ruder, Hooker, and Fadaee}]{Gureja2024MRewardBenchER}
Srishti Gureja, Lester James~Validad Miranda, Shayekh~Bin Islam, Rishabh Maheshwary, Drishti Sharma, Gusti Winata, Nathan Lambert, Sebastian Ruder, Sara Hooker, and Marzieh Fadaee. 2024.
\newblock \href {https://arxiv.org/abs/2410.15522} {M-{R}eward{B}ench: Evaluating reward models in multilingual settings}.
\newblock \emph{arXiv:2410.15522}.

\bibitem[{Jiang et~al.(2024)Jiang, Sablayrolles, Roux, Mensch, Savary, Bamford, Chaplot, Casas, Hanna, Bressand et~al.}]{jiang2024mixtral}
Albert~Q Jiang, Alexandre Sablayrolles, Antoine Roux, Arthur Mensch, Blanche Savary, Chris Bamford, Devendra~Singh Chaplot, Diego de~las Casas, Emma~Bou Hanna, Florian Bressand, et~al. 2024.
\newblock \href {https://arxiv.org/abs/2401.04088} {Mixtral of experts}.
\newblock \emph{arXiv:2401.04088}.

\bibitem[{Joshi et~al.(2017)Joshi, Choi, Weld, and Zettlemoyer}]{joshi-etal-2017-triviaqa}
Mandar Joshi, Eunsol Choi, Daniel Weld, and Luke Zettlemoyer. 2017.
\newblock \href {https://doi.org/10.18653/v1/P17-1147} {{T}rivia{QA}: A large scale distantly supervised challenge dataset for reading comprehension}.
\newblock In \emph{Proceedings of the 55th Annual Meeting of the Association for Computational Linguistics (Volume 1: Long Papers)}, pages 1601--1611, Vancouver, Canada. Association for Computational Linguistics.

\bibitem[{Kirk et~al.(2024)Kirk, Whitefield, R{\"o}ttger, Bean, Margatina, Ciro, Mosquera, Bartolo, Williams, He, Vidgen, and Hale}]{kirk2024prism}
Hannah~Rose Kirk, Alexander Whitefield, Paul R{\"o}ttger, Andrew Bean, Katerina Margatina, Juan Ciro, Rafael Mosquera, Max Bartolo, Adina Williams, He~He, Bertie Vidgen, and Scott~A Hale. 2024.
\newblock \href {https://arxiv.org/abs/2404.16019} {The {PRISM} alignment project: What participatory, representative and individualised human feedback reveals about the subjective and multicultural alignment of large language models}.
\newblock \emph{arXiv:2404.16019}.

\bibitem[{Kull et~al.(2017)Kull, Filho, and Flach}]{kull2017beta}
Meelis Kull, Telmo~Silva Filho, and Peter Flach. 2017.
\newblock \href {https://proceedings.mlr.press/v54/kull17a.html} {{Beta calibration: a well-founded and easily implemented improvement on logistic calibration for binary classifiers}}.
\newblock In \emph{Proceedings of the 20th International Conference on Artificial Intelligence and Statistics}, volume~54 of \emph{Proceedings of Machine Learning Research}, pages 623--631. PMLR.

\bibitem[{Lambert et~al.(2024)Lambert, Pyatkin, Morrison, Miranda, Lin, Chandu, Dziri, Kumar, Zick, Choi et~al.}]{lambert2024rewardbench}
Nathan Lambert, Valentina Pyatkin, Jacob Morrison, LJ~Miranda, Bill~Yuchen Lin, Khyathi Chandu, Nouha Dziri, Sachin Kumar, Tom Zick, Yejin Choi, et~al. 2024.
\newblock \href {https://arxiv.org/abs/2403.13787} {Reward{B}ench: Evaluating reward models for language modeling}.
\newblock \emph{arXiv:2403.13787}.

\bibitem[{Lee et~al.(2024{\natexlab{a}})Lee, Hwang, Kim, Park, and Jung}]{lee2024llm}
Dongryeol Lee, Yerin Hwang, Yongil Kim, Joonsuk Park, and Kyomin Jung. 2024{\natexlab{a}}.
\newblock \href {https://arxiv.org/abs/2410.20774} {Are {LLM}-judges robust to expressions of uncertainty? investigating the effect of epistemic markers on {LLM}-based evaluation}.
\newblock \emph{arXiv:2410.20774}.

\bibitem[{Lee et~al.(2024{\natexlab{b}})Lee, Phatale, Mansoor, Mesnard, Ferret, Lu, Bishop, Hall, Carbune, Rastogi, and Prakash}]{lee2024rlaifvsrlhfscaling}
Harrison Lee, Samrat Phatale, Hassan Mansoor, Thomas Mesnard, Johan Ferret, Kellie Lu, Colton Bishop, Ethan Hall, Victor Carbune, Abhinav Rastogi, and Sushant Prakash. 2024{\natexlab{b}}.
\newblock \href {https://arxiv.org/abs/2309.00267} {{RLAIF} vs. {RLHF}: Scaling reinforcement learning from human feedback with ai feedback}.
\newblock \emph{arXiv:2309.00267}.

\bibitem[{Li et~al.(2024)Li, Chiang, Frick, Dunlap, Wu, Zhu, Gonzalez, and Stoica}]{li2024crowdsourced}
Tianle Li, Wei-Lin Chiang, Evan Frick, Lisa Dunlap, Tianhao Wu, Banghua Zhu, Joseph~E Gonzalez, and Ion Stoica. 2024.
\newblock \href {https://arxiv.org/abs/2406.11939} {From crowdsourced data to high-quality benchmarks: Arena-hard and benchbuilder pipeline}.
\newblock \emph{arXiv:2406.11939}.

\bibitem[{Likert(1932)}]{likert1932technique}
Rensis Likert. 1932.
\newblock A technique for the measurement of attitudes.
\newblock \emph{Archives of Psychology}.

\bibitem[{Liu et~al.(2024{\natexlab{a}})Liu, Zeng, Liu, Yan, He, Wang, Yan, Liu, and Zhou}]{liu2024skywork}
Chris~Yuhao Liu, Liang Zeng, Jiacai Liu, Rui Yan, Jujie He, Chaojie Wang, Shuicheng Yan, Yang Liu, and Yahui Zhou. 2024{\natexlab{a}}.
\newblock \href {https://arxiv.org/abs/2410.18451} {Skywork-reward: Bag of tricks for reward modeling in {LLM}s}.
\newblock \emph{arXiv:2410.18451}.

\bibitem[{Liu et~al.(2024{\natexlab{b}})Liu, Yao, Min, Cao, Hou, and Li}]{liu2024rm}
Yantao Liu, Zijun Yao, Rui Min, Yixin Cao, Lei Hou, and Juanzi Li. 2024{\natexlab{b}}.
\newblock \href {https://arxiv.org/abs/2410.16184} {{RM}-bench: Benchmarking reward models of language models with subtlety and style}.
\newblock \emph{arXiv:2410.16184}.

\bibitem[{Lou et~al.(2024)Lou, Yan, Shen, Yan, Xie, and Zhang}]{lou2024uncertainty}
Xingzhou Lou, Dong Yan, Wei Shen, Yuzi Yan, Jian Xie, and Junge Zhang. 2024.
\newblock \href {https://arxiv.org/abs/2410.00847} {Uncertainty-aware reward model: Teaching reward models to know what is unknown}.
\newblock \emph{arXiv:2410.00847}.

\bibitem[{Louis and Nenkova(2013)}]{louis2013automatically}
Annie Louis and Ani Nenkova. 2013.
\newblock \href {https://doi.org/10.1162/COLI_a_00123} {Automatically assessing machine summary content without a gold standard}.
\newblock \emph{Computational Linguistics}, 39(2):267--300.

\bibitem[{Mizrahi et~al.(2024)Mizrahi, Kaplan, Malkin, Dror, Shahaf, and Stanovsky}]{mizrahi2024state}
Moran Mizrahi, Guy Kaplan, Dan Malkin, Rotem Dror, Dafna Shahaf, and Gabriel Stanovsky. 2024.
\newblock \href {https://arxiv.org/abs/2401.00595} {State of what art? a call for multi-prompt {LLM} evaluation}.
\newblock \emph{arXiv:2401.00595}.

\bibitem[{Panickssery et~al.(2024)Panickssery, Bowman, and Feng}]{panickssery2024llm}
Arjun Panickssery, Samuel~R. Bowman, and Shi Feng. 2024.
\newblock \href {https://proceedings.neurips.cc/paper_files/paper/2024/file/7f1f0218e45f5414c79c0679633e47bc-Paper-Conference.pdf} {Llm evaluators recognize and favor their own generations}.
\newblock In \emph{Advances in Neural Information Processing Systems}, volume~37, pages 68772--68802. Curran Associates, Inc.

\bibitem[{Park et~al.(2024)Park, Jwa, Meiying, Kim, and Choi}]{park-etal-2024-offsetbias}
Junsoo Park, Seungyeon Jwa, Ren Meiying, Daeyoung Kim, and Sanghyuk Choi. 2024.
\newblock \href {https://aclanthology.org/2024.findings-emnlp.57} {{O}ffset{B}ias: Leveraging debiased data for tuning evaluators}.
\newblock In \emph{Findings of the Association for Computational Linguistics: EMNLP 2024}, pages 1043--1067, Miami, Florida, USA. Association for Computational Linguistics.

\bibitem[{Perlitz et~al.(2024)Perlitz, Gera, Arviv, Yehudai, Bandel, Shnarch, Shmueli-Scheuer, and Choshen}]{perlitz2024llmbenchmarksagreefixing}
Yotam Perlitz, Ariel Gera, Ofir Arviv, Asaf Yehudai, Elron Bandel, Eyal Shnarch, Michal Shmueli-Scheuer, and Leshem Choshen. 2024.
\newblock \href {https://arxiv.org/abs/2407.13696} {Do these {LLM} benchmarks agree? {F}ixing benchmark evaluation with {B}ench{B}ench}.
\newblock \emph{arXiv:2407.13696}.

\bibitem[{Peyrard et~al.(2021)Peyrard, Zhao, Eger, and West}]{peyrard-etal-2021-better}
Maxime Peyrard, Wei Zhao, Steffen Eger, and Robert West. 2021.
\newblock \href {https://doi.org/10.18653/v1/2021.acl-long.179} {Better than average: Paired evaluation of {NLP} systems}.
\newblock In \emph{Proceedings of the 59th Annual Meeting of the Association for Computational Linguistics and the 11th International Joint Conference on Natural Language Processing (Volume 1: Long Papers)}, pages 2301--2315, Online. Association for Computational Linguistics.

\bibitem[{Reiter and Belz(2009)}]{reiter-belz-2009-investigation}
Ehud Reiter and Anja Belz. 2009.
\newblock \href {https://doi.org/10.1162/coli.2009.35.4.35405} {An investigation into the validity of some metrics for automatically evaluating natural language generation systems}.
\newblock \emph{Computational Linguistics}, 35(4):529--558.

\bibitem[{Saito et~al.(2023)Saito, Wachi, Wataoka, and Akimoto}]{saito2023verbosity}
Keita Saito, Akifumi Wachi, Koki Wataoka, and Youhei Akimoto. 2023.
\newblock \href {https://arxiv.org/abs/2310.10076} {Verbosity bias in preference labeling by large language models}.
\newblock \emph{arXiv:2310.10076}.

\bibitem[{Silva~Filho et~al.(2023)Silva~Filho, Song, Perello-Nieto, Santos-Rodriguez, Kull, and Flach}]{silva2023classifier}
Telmo Silva~Filho, Hao Song, Miquel Perello-Nieto, Raul Santos-Rodriguez, Meelis Kull, and Peter Flach. 2023.
\newblock \href {https://doi.org/10.1007/s10994-023-06336-7} {Classifier calibration: a survey on how to assess and improve predicted class probabilities}.
\newblock \emph{Machine Learning}, 112(9):3211--3260.

\bibitem[{Tan et~al.(2024)Tan, Zhuang, Montgomery, Tang, Cuadron, Wang, Popa, and Stoica}]{tan2024judgebench}
Sijun Tan, Siyuan Zhuang, Kyle Montgomery, William~Y Tang, Alejandro Cuadron, Chenguang Wang, Raluca~Ada Popa, and Ion Stoica. 2024.
\newblock \href {https://arxiv.org/abs/2410.12784} {Judge{B}ench: A benchmark for evaluating {LLM}-based judges}.
\newblock \emph{arXiv:2410.12784}.

\bibitem[{Thakur et~al.(2024)Thakur, Choudhary, Ramayapally, Vaidyanathan, and Hupkes}]{thakur2024judging}
Aman~Singh Thakur, Kartik Choudhary, Venkat~Srinik Ramayapally, Sankaran Vaidyanathan, and Dieuwke Hupkes. 2024.
\newblock \href {https://arxiv.org/abs/2406.12624} {Judging the judges: Evaluating alignment and vulnerabilities in {LLM}s-as-judges}.
\newblock \emph{arXiv:2406.12624}.

\bibitem[{Tian et~al.(2023)Tian, Mitchell, Zhou, Sharma, Rafailov, Yao, Finn, and Manning}]{tian-etal-2023-just}
Katherine Tian, Eric Mitchell, Allan Zhou, Archit Sharma, Rafael Rafailov, Huaxiu Yao, Chelsea Finn, and Christopher Manning. 2023.
\newblock \href {https://doi.org/10.18653/v1/2023.emnlp-main.330} {Just ask for calibration: Strategies for eliciting calibrated confidence scores from language models fine-tuned with human feedback}.
\newblock In \emph{Proceedings of the 2023 Conference on Empirical Methods in Natural Language Processing}, pages 5433--5442, Singapore. Association for Computational Linguistics.

\bibitem[{Tukey(1949)}]{tukey1949comparing}
John~W Tukey. 1949.
\newblock Comparing individual means in the analysis of variance.
\newblock \emph{Biometrics}, pages 99--114.

\bibitem[{von D{\"a}niken et~al.(2024)von D{\"a}niken, Deriu, and Cieliebak}]{von2024measure}
Pius von D{\"a}niken, Jan Deriu, and Mark Cieliebak. 2024.
\newblock \href {https://arxiv.org/abs/2412.03152} {A measure of the system dependence of automated metrics}.
\newblock \emph{arXiv:2412.03152}.

\bibitem[{Von~D{\"a}niken et~al.(2024)Von~D{\"a}niken, Deriu, Tuggener, and Cieliebak}]{von-daniken-etal-2024-favi}
Pius Von~D{\"a}niken, Jan Deriu, Don Tuggener, and Mark Cieliebak. 2024.
\newblock \href {https://doi.org/10.18653/v1/2024.acl-long.243} {Favi-{S}core: A measure for favoritism in automated preference ratings for generative {AI} evaluation}.
\newblock In \emph{Proceedings of the 62nd Annual Meeting of the Association for Computational Linguistics (Volume 1: Long Papers)}, pages 4437--4454, Bangkok, Thailand. Association for Computational Linguistics.

\bibitem[{Wang et~al.(2024)Wang, Xiong, Xie, Zhao, and Zhang}]{wang-etal-2024-interpretable}
Haoxiang Wang, Wei Xiong, Tengyang Xie, Han Zhao, and Tong Zhang. 2024.
\newblock \href {https://aclanthology.org/2024.findings-emnlp.620} {Interpretable preferences via multi-objective reward modeling and mixture-of-experts}.
\newblock In \emph{Findings of the Association for Computational Linguistics: EMNLP 2024}, pages 10582--10592, Miami, Florida, USA. Association for Computational Linguistics.

\bibitem[{Wang et~al.(2023)Wang, Li, Chen, Cai, Zhu, Lin, Cao, Liu, Liu, and Sui}]{wang2023large}
Peiyi Wang, Lei Li, Liang Chen, Zefan Cai, Dawei Zhu, Binghuai Lin, Yunbo Cao, Qi~Liu, Tianyu Liu, and Zhifang Sui. 2023.
\newblock \href {https://arxiv.org/abs/2305.17926} {Large language models are not fair evaluators}.
\newblock \emph{arXiv:2305.17926}.

\bibitem[{Wei et~al.(2024)Wei, He, Xia, Wong, Lin, and Han}]{wei2024systematic}
Hui Wei, Shenghua He, Tian Xia, Andy Wong, Jingyang Lin, and Mei Han. 2024.
\newblock \href {https://arxiv.org/abs/2408.13006} {Systematic evaluation of {LLM}-as-a-judge in {LLM} alignment tasks: Explainable metrics and diverse prompt templates}.
\newblock \emph{arXiv:2408.13006}.

\bibitem[{Xu et~al.(2024)Xu, Zhu, Zhao, Pan, Li, and Wang}]{xu2024pride}
Wenda Xu, Guanglei Zhu, Xuandong Zhao, Liangming Pan, Lei Li, and William Wang. 2024.
\newblock \href {https://doi.org/10.18653/v1/2024.acl-long.826} {Pride and prejudice: {LLM} amplifies self-bias in self-refinement}.
\newblock In \emph{Proceedings of the 62nd Annual Meeting of the Association for Computational Linguistics (Volume 1: Long Papers)}, pages 15474--15492, Bangkok, Thailand. Association for Computational Linguistics.

\bibitem[{Yang et~al.(2024)Yang, Ding, Lin, Zhang, and Zhang}]{yang2024regularizing}
Rui Yang, Ruomeng Ding, Yong Lin, Huan Zhang, and Tong Zhang. 2024.
\newblock \href {https://openreview.net/forum?id=jwh9MHEfmY} {Regularizing hidden states enables learning generalizable reward model for {LLM}s}.
\newblock In \emph{Advances in Neural Information Processing Systems}.

\bibitem[{Ye et~al.(2024)Ye, Wang, Huang, Chen, Zhang, Moniz, Gao, Geyer, Huang, Chen et~al.}]{ye2024justice}
Jiayi Ye, Yanbo Wang, Yue Huang, Dongping Chen, Qihui Zhang, Nuno Moniz, Tian Gao, Werner Geyer, Chao Huang, Pin-Yu Chen, et~al. 2024.
\newblock \href {https://arxiv.org/abs/2410.02736} {Justice or prejudice? quantifying biases in {LLM}-as-a-judge}.
\newblock \emph{arXiv:2410.02736}.

\bibitem[{Yehudai et~al.(2024)Yehudai, Carmeli, Mass, Arviv, Mills, Shnarch, and Choshen}]{yehudai2024achieving}
Asaf Yehudai, Boaz Carmeli, Yosi Mass, Ofir Arviv, Nathaniel Mills, Eyal Shnarch, and Leshem Choshen. 2024.
\newblock \href {https://openreview.net/forum?id=RjYKTQ0L0W} {Achieving human parity in content-grounded datasets generation}.
\newblock In \emph{The Twelfth International Conference on Learning Representations}.

\bibitem[{Yuan et~al.(2024)Yuan, Cui, Wang, Ding, Wang, Deng, Shan, Chen, Xie, Lin, Liu, Zhou, Peng, Liu, and Sun}]{yuan2024advancing}
Lifan Yuan, Ganqu Cui, Hanbin Wang, Ning Ding, Xingyao Wang, Jia Deng, Boji Shan, Huimin Chen, Ruobing Xie, Yankai Lin, Zhenghao Liu, Bowen Zhou, Hao Peng, Zhiyuan Liu, and Maosong Sun. 2024.
\newblock \href {https://arxiv.org/abs/2404.02078} {Advancing {LLM} reasoning generalists with preference trees}.
\newblock \emph{arXiv:2404.02078}.

\bibitem[{Zheng et~al.(2023)Zheng, Chiang, Sheng, Zhuang, Wu, Zhuang, Lin, Li, Li, Xing, Zhang, Gonzalez, and Stoica}]{zheng2023llmaaj}
Lianmin Zheng, Wei-Lin Chiang, Ying Sheng, Siyuan Zhuang, Zhanghao Wu, Yonghao Zhuang, Zi~Lin, Zhuohan Li, Dacheng Li, Eric Xing, Hao Zhang, Joseph~E Gonzalez, and Ion Stoica. 2023.
\newblock \href {https://proceedings.neurips.cc/paper_files/paper/2023/file/91f18a1287b398d378ef22505bf41832-Paper-Datasets_and_Benchmarks.pdf} {Judging {LLM}-as-a-judge with {MT}-bench and chatbot arena}.
\newblock In \emph{Advances in Neural Information Processing Systems}, volume~36, pages 46595--46623. Curran Associates, Inc.

\end{thebibliography}
